%% file: main.tex
\documentclass[10pt,twocolumn,letterpaper]{article}

\usepackage[pagenumbers]{wacv} %

\usepackage[nolist]{acronym}

\usepackage{graphicx}
\usepackage{amsmath}
\usepackage{amssymb}
\usepackage{booktabs}
\usepackage{placeins}
\usepackage[english]{babel}
\usepackage{blindtext}
\usepackage[Export]{adjustbox}
\usepackage{tabularx}
\usepackage{arydshln}
\usepackage{cellspace}
\usepackage{multirow}
\usepackage{diagbox}
\usepackage{makecell}
\usepackage[pagebackref,breaklinks,colorlinks]{hyperref}

\usepackage[capitalize]{cleveref}
\crefname{section}{Sec.}{Secs.}
\Crefname{section}{Section}{Sections}
\Crefname{table}{Table}{Tables}
\crefname{table}{Tab.}{Tabs.}

\input{acronym}

\begin{document}

\title{Controlling Human Shape and Pose in Text-to-Image Diffusion Models via Domain Adaptation}

\author{
Benito Buchheim \hspace{2em} Max Reimann \hspace{2em} Jürgen Döllner \\[0.5ex]  
University of Potsdam, Digital Engineering Faculty, Potsdam, Germany
 }
\maketitle

\begin{abstract}
We present a methodology for conditional control of human shape and pose in pretrained text-to-image diffusion models using a 3D human parametric model (SMPL). 
Fine-tuning these diffusion models to adhere to new conditions requires large datasets and high-quality annotations, which can be more cost-effectively acquired through synthetic data generation rather than real-world data. However, the domain gap and low scene diversity of synthetic data can compromise the pretrained model's visual fidelity. We propose a domain-adaptation technique that maintains image quality by isolating synthetically trained conditional information in the classifier-free guidance vector and composing it with another control network to adapt the generated images to the input domain. 
To achieve SMPL control, we fine-tune a ControlNet-based architecture on the synthetic SURREAL dataset of rendered humans and apply our domain adaptation at generation time. 
Experiments demonstrate that our model achieves greater shape and pose diversity than the 2d pose-based ControlNet, while maintaining the visual fidelity and improving stability, proving its usefulness for downstream tasks such as human animation.
\url{https://ivpg.github.io/humanLDM}
\end{abstract}

\begin{figure}[ht]
    \centering
    \begin{subfigure}[b]{\linewidth}
        \includegraphics[width=\textwidth, clip, trim={0 0 0 130}]{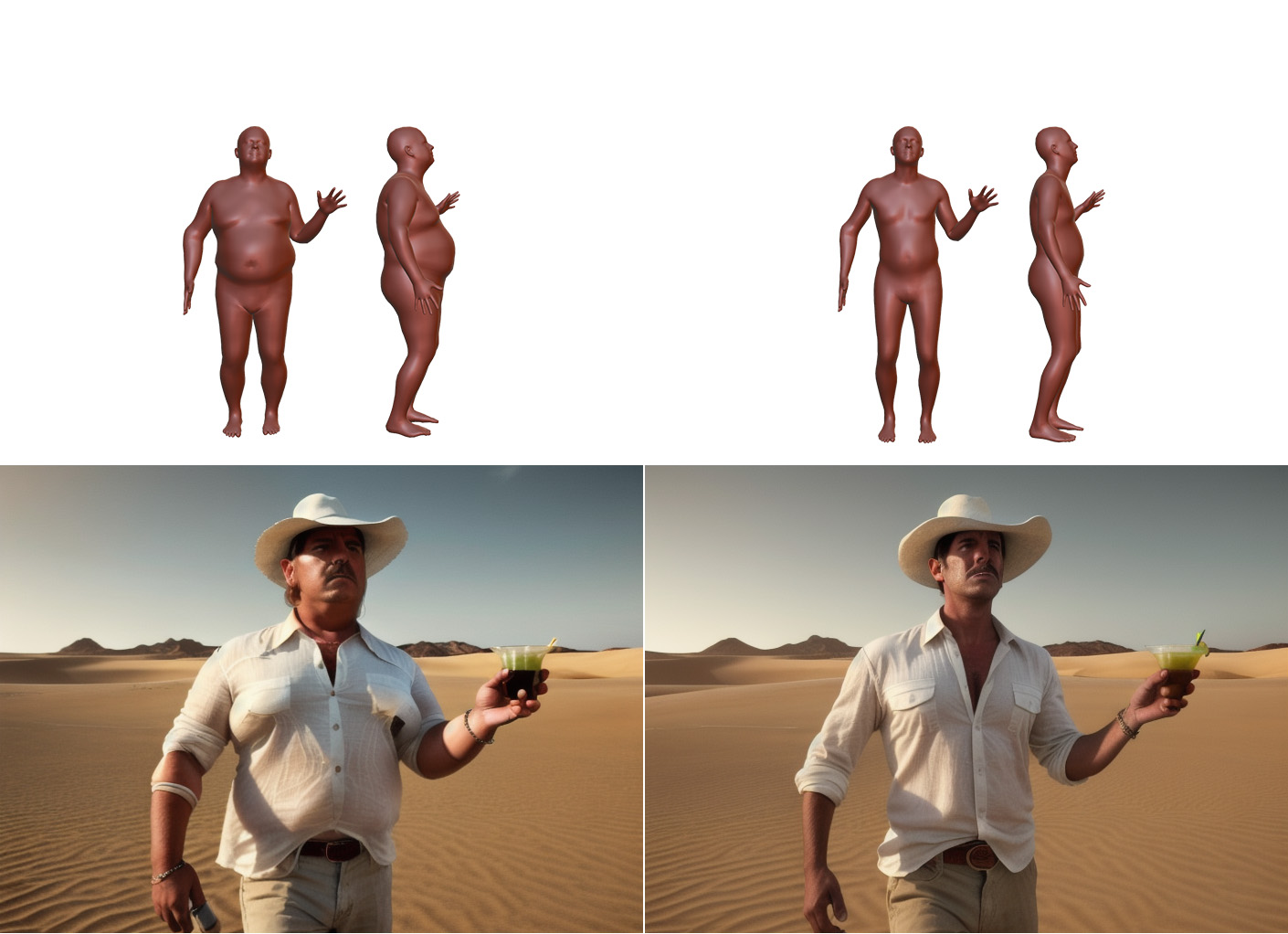}
        \caption{Input SMPL control mesh and our SD~\cite{rombach2022high} outputs}
    \end{subfigure}
    \begin{subfigure}[b]{\linewidth}
        \begin{subfigure}[b]{0.49\linewidth}
            \includegraphics[width=\textwidth]{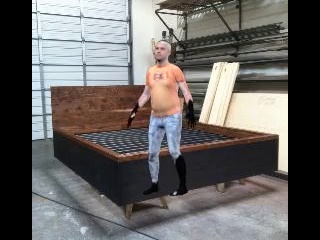}
        \end{subfigure}
        \begin{subfigure}[b]{0.49\linewidth}
            \includegraphics[width=\textwidth]{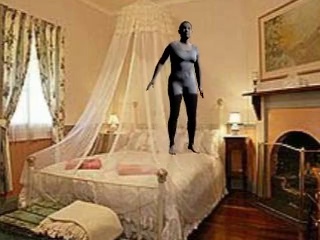}
        \end{subfigure}
        \caption{Synthetic training data \cite{SURREAL}}
    \end{subfigure}
    \caption{Our approach allows 3d parametric control over human pose and shape (a) in LDMs using SMPL~\cite{SMPL:2015} meshes. We train on synthetic data (b) and propose a domain adaptation technique to adapt model outputs into the original visual domain. }
    \label{fig:teaser}
\end{figure}

\input{introduction}

\input{relatedwork}

\input{method}

\input{results}

\input{guidance_ablations}

\input{conclusion}

{\small
\bibliographystyle{ieee_fullname}
\bibliography{main}
}

\clearpage

\renewcommand*{\thesection}{\Alph{section}}
\setcounter{section}{0}
\renewcommand*{\theHsection}{chX.\the\value{section}}
\section*{Appendix}
\input{background}

\input{supplemental_content}

\end{document}

%% file: acronym.tex
\begin{acronym}
\acro{CG}{Computer Graphics}
\acro{CNN}{Convolutional Neural Network}
\acro{CPU}{Central Processing Unit}
\acro{CV}{Computer Vision}
\acro{DL}{Deep Learning}
\acro{DNN}{Deep Neural Network}
\acro{EPE}{Endpoint Error}
\acro{FPS}{Frames per second}
\acro{GPU}{Graphics Processing Unit}
\acro{GAN}{Generative Adversarial Network}
\acro{K}{Kilo, thousand}
\acro{M}{Million}
\acro{MB}{Megabyte}
\acro{ReLU}{Rectified Linear Unit}
\acro{SSIM}{Structural Similarity Index Measure}
\acro{PSNR}{Peak Signal-to-Noise Ratio}
\acro{FID}{Fréchet Inception Distance}
\acro{LPIPS}{Learned Perceptual Image Patch Similarity}
\acro{MPJPE}{mean per-joint point error}
\acro{MPJPE-PA}{Mean Per Joint Position Error after Procrustes Analysis}
\acro{PVET}{Per-Vertex-Error in T-Pose}
\acro{KID}{Kernel Inception Distance} 
\acro{IS}{Inception Score}
\acro{LPIPS}{Learned Perceptual Image Patch Similarity}
\acro{LDM}{Latent Diffusion Model}
\acro{SD}{Stable Diffusion}
\acro{HPH}{Hierarchical Probabilistic Human Estimation}
\end{acronym}

%% file: introduction.tex
\section{Introduction}
\label{sec:intro}

The advent of advanced generative diffusion models, particularly \acp{LDM} \cite{rombach2022high}, has revolutionized the field of image generation. Models for text-to-image synthesis such as Stable Diffusion \cite{rombach2022high}, DALLE-3 \cite{betker2023improving} or Gemini \cite{team2023gemini} have made high quality image synthesis from complex prompts accessible to a broad audience of users worldwide. Text alone, however, is not sufficient in many cases to describe exact scene layouts, subjects or style. Recent techniques for added spatial control have enhanced the capabilities of these models by incorporating various forms of spatial guidance. ControlNet and  T2I-Adapter~\cite{zhangAddingConditionalControl2023,mouT2IAdapterLearningAdapters2023}, for instance, augment \acp{LDM} with localized, task-specific conditions like edges and human poses, allowing for precise control over the generated images.  
 
Our work focuses on precise control over humans in generated scenes (\cref{fig:teaser}). While 2D pose conditions~\cite{zhangAddingConditionalControl2023, mouT2IAdapterLearningAdapters2023} allow specifying constraints on human poses, body shapes are not controllable, and specified poses might suffer from 3D ambiguities, complicating precise human-centric illustration and animation. To address this, we aim to control body pose and shape using the commonly used 3D human parametric model, SMPL~\cite{SMPL:2015}. Our approach modifies the ControlNet \cite{zhangAddingConditionalControl2023} architecture to use SMPL parameter embeddings in the cross-attention blocks that typically attend to text condition embeddings. This modification enables the model to control global image content based on SMPL guidance, seamlessly integrating human geometry information into the generation process (\cref{fig:teaser}a).

However, training such control networks for large \acp{LDM} like \ac{SD} \cite{rombach2022high} typically requires an annotated training dataset with a large and diverse set of high-quality images to retain the output fidelity of the original network. Most existing real-world datasets for 3D annotations, including SMPL parameters, do not meet these criteria, particularly regarding scale and diversity in scenes and body shapes. Synthetic datasets such as SURREAL \cite{SURREAL} provide images with diverse body shapes and scenes at scale, and offer further advantages such as high-quality annotations and lower regulatory barriers.
However, the  visual fidelity of synthetic datasets is typically degraded. For instance, SURREAL \cite{SURREAL} suffers from non-photorealistic shading (\cref{fig:teaser}b), inadequate blending between subjects and background, as well as low image resolution, leading to a visual domain shift in SD networks fine-tuned on it.
Our goal is to extract the human shape and pose information while discarding the visual aesthetics from such datasets.
While traditional domain adaptation techniques \cite{kouw2019review,atapour2018real} aim to translate synthetic data to real-world appearance 
, adapting synthetic datasets to the full visual range of outputs of large text-to-image \acp{LDM} representing diverse and complex scenes like \ac{SD} is challenging.

We introduce a technique for domain adaptation in the latent space of \acp{LDM} to extract the control condition of a synthentically-trained \ac{LDM} control network and apply it in the visual domain of the original model using classifier-free guidance. For human body shape and pose control, we obtain a SMPL-conditioned vector from our synthetically trained ControlNet, which is then composed with outputs of another control network for visual domain guidance, effectively adapting the isolated SMPL-condition to the visual appearance and fidelity of the original \ac{SD} model.
Our results demonstrate significantly improved adherence to shape and pose compared to current control approaches~\cite{zhangAddingConditionalControl2023, mouT2IAdapterLearningAdapters2023}, while simultaneously retaining the visual fidelity (in terms of KID and Inception Score) of \ac{SD}.

In summary our contributions are as follows:
\begin{itemize}
    \item We introduce a classifier-free guidance-based technique for domain adaptation in \acp{LDM} to adapt the visual domain of control models trained on synthetic data to the original high fidelity domain .
    \item We propose a SMPL-based control model for shape and pose control trained on the synthetic SURREAL \cite{SURREAL} dataset. 
    \item We demonstrate the efficacy of our approach in terms of visual fidelity, SMPL accuracy, and comparisons against state-of-the-art methods and ablated configurations.
\end{itemize}

%% file: relatedwork.tex
\section{Related Work}

\textbf{Image Generator Domain Adaptation.}
Domain adaptation \cite{kouw2019review} aims to shift the data distribution of a model to a new domain different to the one it was trained on. Various methods of domain adaptation for image generation models have been explored, both to adapt model weights \cite{gal2022stylegan,sun2020test}, as well as outputs during inference \cite{patashnik2021styleclip,mokadyNULLTextInversionEditing,zhaoNulltextGuidanceDiffusion2023}. The latent space of GANs such as StyleGAN~\cite{karras2020analyzing} has been shown to be highly adaptable to new conditional domains such as CLIP\cite{radford2021learning}-embeddings \cite{patashnik2021styleclip,gal2022stylegan} %
or \ac{LDM}-based classifier-free guidance \cite{song2024stylegan}. 
Adding conditional control to LDMs without retraining on labeled data can be seen as a form of test-time domain adaptation (TTA) \cite{liang2023comprehensive}, adapting the model feature-space \cite{tumanyanPlugandPlayDiffusionFeatures2022} or latent noise vector \cite{mokadyNULLTextInversionEditing,zhaoNulltextGuidanceDiffusion2023} to the new conditional domain during generation. However, these methods do not address visual domain shifts. Our proposed generator domain adapation technique, similar to traditional TTA \cite{liang2023comprehensive}, transfers the visual domain from a synthetically trained source to a desired visual target domain at generation time. Instead of adapting training data \cite{atapour2018real} or model weights \cite{sun2020test}, we isolate the conditioning signal from its visual appearance using classifier-free guidance. 
Following Liu et al. \cite{liu2022compositional}, who propose composition of multiple conditional domains, we use guidance composition to combine the isolated SMPL guidance vector with a visual domain guidance vector from a control network trained in the original \ac{SD} domain.

\begin{figure*}[ht]
    \centering
    \begin{subfigure}[b]{0.30\textwidth}
        \centering
        \includegraphics[page=1, width=\textwidth]{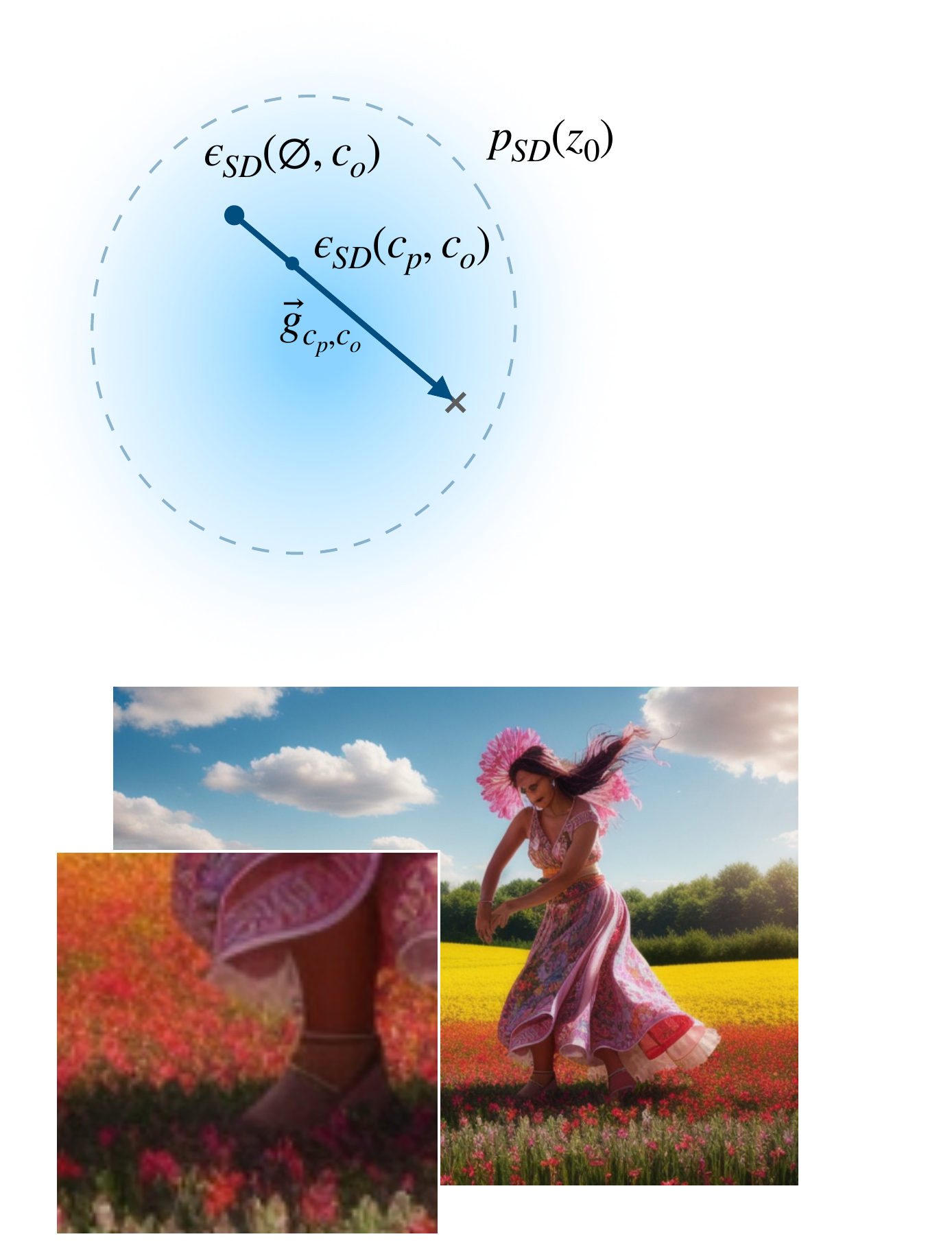}
        \caption{ControlNet}
    \end{subfigure}
    \begin{subfigure}[b]{0.30\textwidth}
        \centering
        \includegraphics[page=2, width=\textwidth]{graphics/Approach_v004_cropped.pdf}
        \caption{Naive Fine-Tuning}
    \end{subfigure}
    \begin{subfigure}[b]{0.30\textwidth}
        \centering
        \includegraphics[page=3, width=\textwidth]{graphics/Approach_v004_cropped.pdf}
        \caption{Guidance Composition}
    \end{subfigure}
    \caption{A pretrained ControlNet $\epsilon_{SD}$ conditioned on 2d poses (a) can generate pose-guided images in the data domain $p_{SD}$ (blue) of the Stable Diffusion model. To enable SMPL-based human shape and 3d-pose control, the model is fine-tuned on a synthetic dataset (b), shifting the model outputs into the synthetic data domain $p_{Syn}$ (orange). Our approach proposes classifier-free guidance composition (c) to adapt the visual output domain to the original data domain while retaining shape and pose control. }
    \label{fig:overview_guidance}
\end{figure*}

\textbf{Controlling Text-to-Image Models.}
Several methods manipulate and control diffusion model outputs. ControlNet \cite{zhangAddingConditionalControl2023} and T2I-Adapter \cite{mouT2IAdapterLearningAdapters2023} enhance pretrained \acp{LDM} with spatially localized, task-specific conditions like edges, depth maps, and human poses by integrating retrained model blocks or trainable adapters \cite{zhangAddingConditionalControl2023,mouT2IAdapterLearningAdapters2023}. DreamBooth fine-tunes models for subject-specific generation and style control using diffusion guidance \cite{natanielruizDreamBoothFineTuning2023}. InstructPix2Pix enables text-based semantic editing and stylization of input images \cite{brooksInstructPix2PixLearningFollow2023}.
Training-free methods, manipulate guidance vectors \cite{subrtovaDiffusionImageAnalogies2023,michelleshuDreamWalkStyleSpace2024}, null-text embeddings \cite{zhaoNulltextGuidanceDiffusion2023,mokadyNULLTextInversionEditing}, or features \cite{tumanyanPlugandPlayDiffusionFeatures2022}. 
DreamWalk \cite{michelleshuDreamWalkStyleSpace2024} decomposes prompts into components and, similarly to ours, uses compositional guidance \cite{liu2022compositional} to emphasize style or spatial concepts during classifier-free guidance. In contrast to the preceding works, that either assume availability of in-domain training data at scale or rely on less precise prompt- or feature-injection at runtime, our approach addresses adding fine-grained spatial control to text-to-image \acp{LDM} in the presence of only synthetic data.

\textbf{SMPL-based Control.}
Few methods have explored 3D body control in LDMs. Champ \cite{shenhaozhuChampControllableConsistent2024} utilizes SMPL-conditioning, and MagicAnimate  \cite{zhongcongxuMagicAnimateTemporallyConsistent2023} densepose-conditioning of LDMs to enhance shape alignment and motion guidance in human image animation. By extracting pose and shape information from a reference video and conditioning a LDM on the rendered semantic maps, they can generate consistent and controllable human animations that replicate the actions observed in the reference video. Compared to these approaches, our method focuses on free-form text to image generation with SMPL-guidance without requiring a reference image while retaining the full representation capability of the original \ac{LDM}.

%% file: method.tex
\section{Method}
\label{sec:method}

\subsection{Problem Setting}
Traditional fine-tuning of a generative model on conditional data typically shifts the output distribution to match the conditional training data. Mathematically, an unconditional generative model with parameters $\theta$ learns the probability distribution $P_\theta(\boldsymbol{x})$ of a training dataset $\mathcal{D} = \{\boldsymbol{x}_1, \boldsymbol{x}_2, ...\}$.
Fine-tuning with a second \emph{annotated} dataset $\mathcal{D'} = \{ (\boldsymbol{y}_1, \boldsymbol{c}_1), (\boldsymbol{y}_2, \boldsymbol{c}_2), ... \}$, optimizes $\theta$ to approximate the conditional distribution $P_{\theta'}(\boldsymbol{y} \mid \boldsymbol{c})$.
This step shifts the marginal distribution of the model to the domain of the dataset $\mathcal{D'}$, effectively approximating $P_{\theta'}(\boldsymbol{y})$.
This process retains the visual output fidelity if distributions $P(\boldsymbol{x})$ and $P(\boldsymbol{y})$ are similar. For instance, ControlNet \cite{zhangAddingConditionalControl2023} matches the visual characteristics of Stable Diffusion by training on similar data with added control annotations (see  \cref{fig:overview_guidance}a). Here, the pretrained 2d-pose conditioned ControlNet remains within the original training domain $P_{\text{SD}}$.
In contrast, fine-tuning on synthetic data, often with lower visual quality, can lead to a domain shift in $P_{\theta'}(\boldsymbol{y})$ and degrade output fidelity. As shown in  \cref{fig:overview_guidance}b, a ControlNet fine-tuned on a 3d pose and shape dataset adheres to the target conditioning, but also shifts to the aesthetics of the synthetic data $P_{\text{Syn}}$.
Faced with the challenge of fine-tuning LDMs with out-of-domain synthetic data to enable control over conditional information, we instead propose a domain adaptation technique that makes use of inherent properties of the LDM image generation process. The key idea is to isolate the conditional information from its domain and apply it within the original model's domain using classifier-free guidance to achieve $P_{\theta'} \sim P(\boldsymbol{x} \mid \boldsymbol{c})$. As depicted in \cref{fig:overview_guidance}c, our guidance composition approach applies shape and pose conditioning in the target domain, maintaining high visual fidelity.

\begin{figure}[htb]
    \centering
    \includegraphics[width=\linewidth, trim={1cm 0.25cm 1.2cm 0.25cm}, clip]{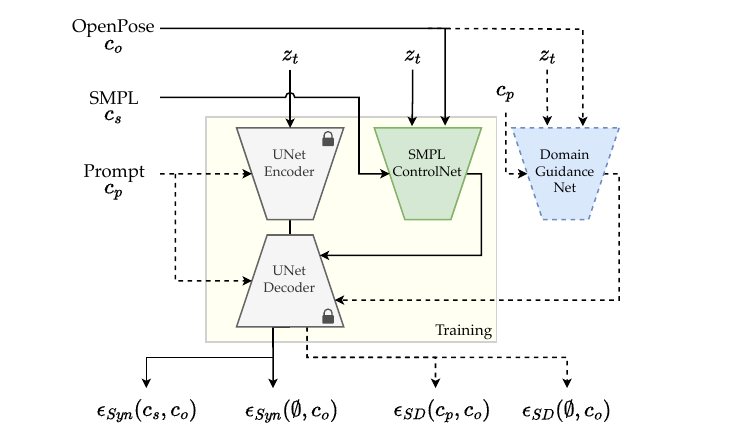}
    \caption{Overview of the networks involved in our SD-control approach during one denoising timestep $t$. 
    During finetuning (solid lines), the overall model output $\epsilon_{\text{Syn}}(c_s,c_o) = \epsilon_\theta(z_t,t,\emptyset,C_\text{SMPL}(\boldsymbol{c}_{\text{s}}, \boldsymbol{c}_{\text{o}}))$ is adapted to a synthetic SMPL image dataset. During inference (solid and dotted lines), a pose-conditioned guidance network ($C_\text{SD}$) is executed alongside $C_\text{SMPL}$ and a composite guidance vector is constructed from the outputs.}
    \label{fig:architecture_overview}
\end{figure}

\subsection{Guidance Domain Adaptation}

Next, we detail our general approach to guidance-based domain adaptation. We utilize two separate SD-control networks, with one responsible for visual domain guidance, i.e. targeting the desired appearance distribution $P_{\text{SD}}$, and the other for attribute guidance, i.e., targeting the conditional attribute distribution $P_{\text{Syn}}(\boldsymbol{y}|\boldsymbol{c})$. By composing these with classifier-free guidance, we isolate the attribute and shift its visual domain to the desired appearance.
 
 \textbf{Domain Guidance and Pose Guidance Networks.} 
  \cref{fig:architecture_overview} illustrates the network components of our approach. 
 We employ a ControlNet $C_\text{SD}$ which serves as the domain appearance guidance net with output latents $\epsilon_\theta$ defined as $\epsilon_{\text{SD}}(\boldsymbol{c}_p,\boldsymbol{c}_o) = \epsilon_\theta(z_t,t,\boldsymbol{c}_p,C_\text{SD}(\boldsymbol{c}_{\text{p}}, \boldsymbol{c}_{\text{o}}))$, where the inputs are the LDM time step $t$, latents $z_t$, prompt $\boldsymbol{c}_p$, and spatial control condition $\boldsymbol{c}_o$. See supplemental A for background on LDMs, ControlNet, and notation. The appearance guidance net was trained on images within the original SD data domain, ensuring $\epsilon_{\text{SD}}(\boldsymbol{c}_p,\boldsymbol{c}_o)  \sim  P_\text{SD}$. 
The output latents in the \textit{synthetic} data domain are defined as $\epsilon_{\text{Syn}}(\boldsymbol{c}_s,\boldsymbol{c}_o) = \epsilon_\theta(z_t,t,\emptyset,C_\text{Syn}(\boldsymbol{c}_{\text{s}}, \boldsymbol{c}_{\text{o}}))$, where $C_\text{Syn}$ is an attribute guidance network finetuned in the synthetic data domain, thus  $\epsilon_{\text{Syn}}(\boldsymbol{c}_s,\boldsymbol{c}_o) \sim P_\text{Syn}$, conditioned on an attribute $\boldsymbol{c}_s$ we want to isolate. 

\textbf{Classifier-Free Guidance.}
Classifier-free guidance (CfG) \cite{hojonathanClassifierFreeDiffusionGuidance2022} steers the generative process of diffusion models by amplifying the conditional's gradient along a vector $\boldsymbol{\Vec{g}_c}$. During training, the model is simultaneously trained on both conditional ($\boldsymbol{c}$) and unconditional ($\emptyset$ or zeroed $\boldsymbol{c}$) objectives. These are combined during inference as:
\begin{align*}
\boldsymbol{\Vec{g}_c} &= \boldsymbol{\epsilon}_{\theta}(\boldsymbol{z}_t,t, \boldsymbol{c}) - \boldsymbol{\epsilon}_{\theta}(\boldsymbol{z}_t,t,\emptyset), \\
\boldsymbol{z}_{t-1} &= \boldsymbol{\epsilon}_{\theta}(\boldsymbol{z}_t,t,\emptyset) + w\boldsymbol{\Vec{g}_c},
\end{align*}
where $w$ controls the guidance strength (CfG scale).  This guidance vector helps in amplifying the desired characteristics specified by $\boldsymbol{c}$ while retaining the overall quality of the generated images. Applying CfG with output latents from the appearance guidance ControlNet $C_\text{SD}$ thus results in a guidance vector $\boldsymbol{\Vec{g}_{c_p,c_o}}$ conditioned on a spatial condition ($\boldsymbol{c}_o$) and pointing towards the prompt ($\boldsymbol{c}_p$):
\begin{equation}
    \boldsymbol{\Vec{g}_{c_p,c_o}} =\epsilon_{\text{SD}}(\boldsymbol{c}_{\text{p}}, \boldsymbol{c}_{\text{o}}) - \epsilon_{\text{SD}}(\emptyset,\boldsymbol{c}_{\text{o}}),  \\
\end{equation}
which we denote as the appearance guidance vector.

\textbf{Guidance Composition.}
Now consider the latents in the synthetic domain $\epsilon_{\text{Syn}}(\boldsymbol{c}_s,\boldsymbol{c}_o)$.  To shift their visual aesthetics into the original data domain ($P_\text{SD}$), we isolate the condition $\boldsymbol{c}_s$ using classifier-free guidance and compose it with the appearance guidance vector $\boldsymbol{\Vec{g}_{c_p,c_o}}$:
\begin{align}
\boldsymbol{\Vec{g}_{c_s,c_o}} &= \epsilon_{\text{Syn}}(\boldsymbol{c}_{\text{s}}, \boldsymbol{c}_{\text{o}}) -  \epsilon_{\text{Syn}}(\emptyset,\boldsymbol{c}_{\text{o}}) \label{eq:cfg_equations_syn}\\
\boldsymbol{\Vec{g}} &=  w_1\boldsymbol{\Vec{g}_{c_p,c_o}} + w_2\boldsymbol{\Vec{g}_{c_s,c_o}} \\
\boldsymbol{z}_{t-1} &= \epsilon_{SD}(\emptyset, \boldsymbol{c}_o) + \boldsymbol{\Vec{g}}
\label{eq:cfg_equations}
\end{align}

where $w_1$ and $w_2$ are weighting factors for the guidance vectors. 
Effectively, the guidance vector $\boldsymbol{\Vec{g}_{c_p,c_o}}$ thus points in the direction of the prompt and visual domain, while the guidance  vector $\boldsymbol{\Vec{g}_{c_s,c_o}}$ points towards  adherence to condition $c_s$. Since neither conditional nor unconditional $\epsilon_{\text{Syn}}$ receives prompt information, increasing the magnitude of $\boldsymbol{\Vec{g}_{c_s,c_o}}$ through $w_2$ does not impart more visual appearance information, thereby isolating the $\boldsymbol{c}_s$ from its visual appearance.
Linearly combining both vectors shifts $\boldsymbol{\Vec{g}_{c_s,c_o}}$ into the original data distribution. We ablate the contributions of the guidance vector components in \cref{subsec:guidanceablation} and provide additional ablations in Sec B.3 and a guidance scale analysis in Sec B.4 of the supplemental material.

\subsection{SMPL-Control Guidance}
\label{subsec:smplguidance}
We apply our proposed guidance domain-adaptation approach to condition Stable Diffusion on SMPL models.
As illustrated in \cref{fig:architecture_overview}, the shared spatial condition $\boldsymbol{c}_o$ to domain guidance network $C_\text{SD}$ and attribute guidance network $C_\text{Syn}$ is a 2D map of OpenPose \cite{openpose2019} pose joints. The attribute guidance network $C_\text{SMPL} = C_\text{Syn}$ is a ControlNet additionally conditioned on SMPL embeddings ($\boldsymbol{c}_s$), see Sec. A of the supplemental for a background on the SMPL model.
The visual domain guidance network $C_\text{SD}$ is a pretrained ControlNet or a T2I-Adapter \cite{mouT2IAdapterLearningAdapters2023} and receives a prompt ($\boldsymbol{c}_p$) in addition to $\boldsymbol{c}_o$. %
We apply guidance composition on the output latents as detailed in \cref{eq:cfg_equations}. Overall, the composite guidance vector $\boldsymbol{\Vec{g}}$ encodes prompt, SMPL-shape and pose control while retaining visual appearance (see \cref{fig:overview_guidance}c).

\textbf{Training.}
To obtain $C_\text{SMPL}$, we condition a ControlNet architecture on SMPL parameters by fine-tuning the cross-attention blocks, which originally attended to prompt embeddings ($\boldsymbol{c}_p$), using embeddings of SMPL parameters ($\boldsymbol{c}_s$) instead.
The shape ($\beta_\text{SMPL}$) and pose ($\theta_\text{SMPL}$) parameters of the SMPL model are concatenated and embedded by a single linear layer, i.e. $\boldsymbol{c}_s = \text{emb}(\theta_\text{SMPL},\beta_\text{SMPL})$, to match the expected input dimensions of the cross-attention blocks. Further, also the SD-UNet only receives empty prompts during training, effectively removing prompt conditioning from $C_\text{SMPL}$.
During training, the network is initialized with the weights of a 2d-pose ($\boldsymbol{c}_o$) conditioned ControlNet. %
After minimizing the training objective, the output latents $\epsilon_{\text{Syn}}(\boldsymbol{c}_s,\boldsymbol{c}_o)$ are adapted to the synthetic training dataset domain. 
Model outputs closely resemble the degraded aesthetics of its synthetic training dataset (e.g., SURREAL\cite{SURREAL}) manifesting in non-realistic skin, faces and clothing and free floating feet (see \cref{fig:overview_guidance}b). 
To enable classifier-free guidance, $C_\text{SMPL}$ is trained on conditional and unconditional inputs, i.e., $\epsilon_{\text{Syn}}(\boldsymbol{c}_{\text{o}}, \emptyset)$, in parallel by randomly zeroing $\boldsymbol{c}_s$, and in both conditional and unconditional cases uses the empty prompt embedding in the SD U-Net during training.

%% file: results.tex
\section{Results}
\label{sec:results}

\subsection{Implementation Details}
We train our models for a single epoch on 200K samples from the SURREAL \cite{SURREAL} dataset, containing SMPL-annotated images, additionally we generate 2d body-pose images (25 color-coded keypoints) using OpenPose \cite{openpose2019} for every image to obtain $\boldsymbol{c}_o$. Notably, no prompts are required to train $C_\text{SMPL}$. We did not see improvement in metrics when training on more samples from the dataset. The training was conducted on a single NVIDIA RTX 4090 with a learning rate of $10^{-5}$.
We adapt the pytorch-based huggingface diffusers library to implement our method, and initialize Stable Diffusion U-Net with SD1.5 weights\footnote{\tiny{\url{https://huggingface.co/runwayml/stable-diffusion-v1-5}}}\cite{rombach2022high} and initialize  $C_\text{SMPL}$ with weights of the ControlNet-OpenPose\footnote{\tiny{\url{https://huggingface.co/lllyasviel/sd-controlnet-openpose}}} \cite{zhangAddingConditionalControl2023}. During inference we also use these weights for the domain-guidance net ($C_\text{SD}$), or alternatively use the pose-conditioned T2I-Adapter-OpenPose\footnote{\tiny{\url{https://huggingface.co/TencentARC/t2iadapter_openpose_sd14v1}}}.

\subsection{Visual Fidelity}

\begin{table}[htbp]
    \centering
    \begin{tabular}{lcc}
        \hline
        \textbf{Generation Method} & \textbf{IS} $\uparrow$ & \textbf{KID} $\downarrow$ \\
        \hline
        ControlNet-OpenPose \cite{zhangAddingConditionalControl2023} & \underline{15.673} & 0.00614 \\
        T2I-Adapter-OpenPose \cite{mouT2IAdapterLearningAdapters2023} & \textbf{15.755} & \underline{0.00609} \\
        $C_\text{SMPL}$-ft-attn + CN & 15.521 & 0.00615 \\
        $C_\text{SMPL}$-ft-attn + T2I & 15.474 & \textbf{0.00597} \\[0.5ex]
        \hdashline
        \textit{Ablation Variants:} \\
        \quad $C_\text{SMPL}$-extra-attn & 15.157 & 0.01470 \\
        \quad $C_\text{SMPL}$-ft-all + CN & 15.222 & 0.00637 \\
        \quad $C_\text{SMPL}$-ft-all + T2I & 15.356 & 0.00607 \\
        \hline
        Reference: COCO Eval Split & 15.280 & - \\
        \hline
    \end{tabular}
    \caption{Inception Score (IS) and Kernel Inception Distance (KID) to the Coco Eval set on images generated from our evaluation set}
    \label{tab:merged_comparison}
\end{table}

We assess the visual fidelity by using poses and prompts from an evaluation set of annotated images to generate similar images, against which we benchmark fidelity metrics.

\textbf{Study Setup.} We curated a subset of the MSCOCO dataset~\cite{MSCOCO} with 5276 images, each containing one person with at least 90\% of keypoints visible and covering at least 10\% of the image's width and length. Using HierProbHuman~\cite{sengupta2021hierarchical} (which performs well for shape extraction), we predicted SMPL models, and use them together with the 2D poses and COCO captions as prompts to condition our benchmarked approaches. This setup generated human-centric photographic images with similar semantic content to our input dataset.
We compared our proposed methods, which involve fine-tuning only cross-attention blocks with ControlNet (ft-attn + CN) and T2I-Adapter (ft-attn + T2I) guidance, and as an ablated variant, fine-tuning the entire model (ft-all). To assess domain gap, we also fine-tune a ControlNet with an extra set of cross-attention blocks inserted after prompt-attention, finetuning only these blocks on the SMPL condition ($C_\text{SMPL}$-extra-attn) and not using guidance composition during inference. Baseline generators are the pretrained ControlNet-OpenPose and T2I-Adapter-OpenPose.

\textbf{Metrics.}
We computed the \ac{IS} \cite{salimans2016improved}, evaluating the quality and diversity of the generated images, and the  \ac{KID} \cite{binkowski2018demystifying} which is similar to the FID \cite{heusel2017gans}, but is more reliable for unbiased evaluation of distribution distances on smaller datasets.

\textbf{Results.} As shown in Table~\ref{tab:merged_comparison}, our attention-finetuning approach with CN or T2I guidance achieves visual fidelity comparable to the baselines. T2I guidance degrades slightly more in terms of IS from its baseline but achieves slightly higher visual similarity (lower KID) to COCO than ControlNet guidance. Fine-tuning all weights ($C_\text{SMPL}$-ft-all) slightly increases the visual domain shift (higher KID). However, considering the indirect mapping between prompt generation and COCO, the KID variations are small and likely within the margin of error. Notably, the SMPL-conditioned ControlNet without domain adaption ($C_\text{SMPL}$-extra-attn) shows a significant domain gap, indicated by a high KID, demonstrating the effectiveness of our domain adaptation approach.

\begin{figure}[t]
    \centering
    \includegraphics[width=\linewidth]{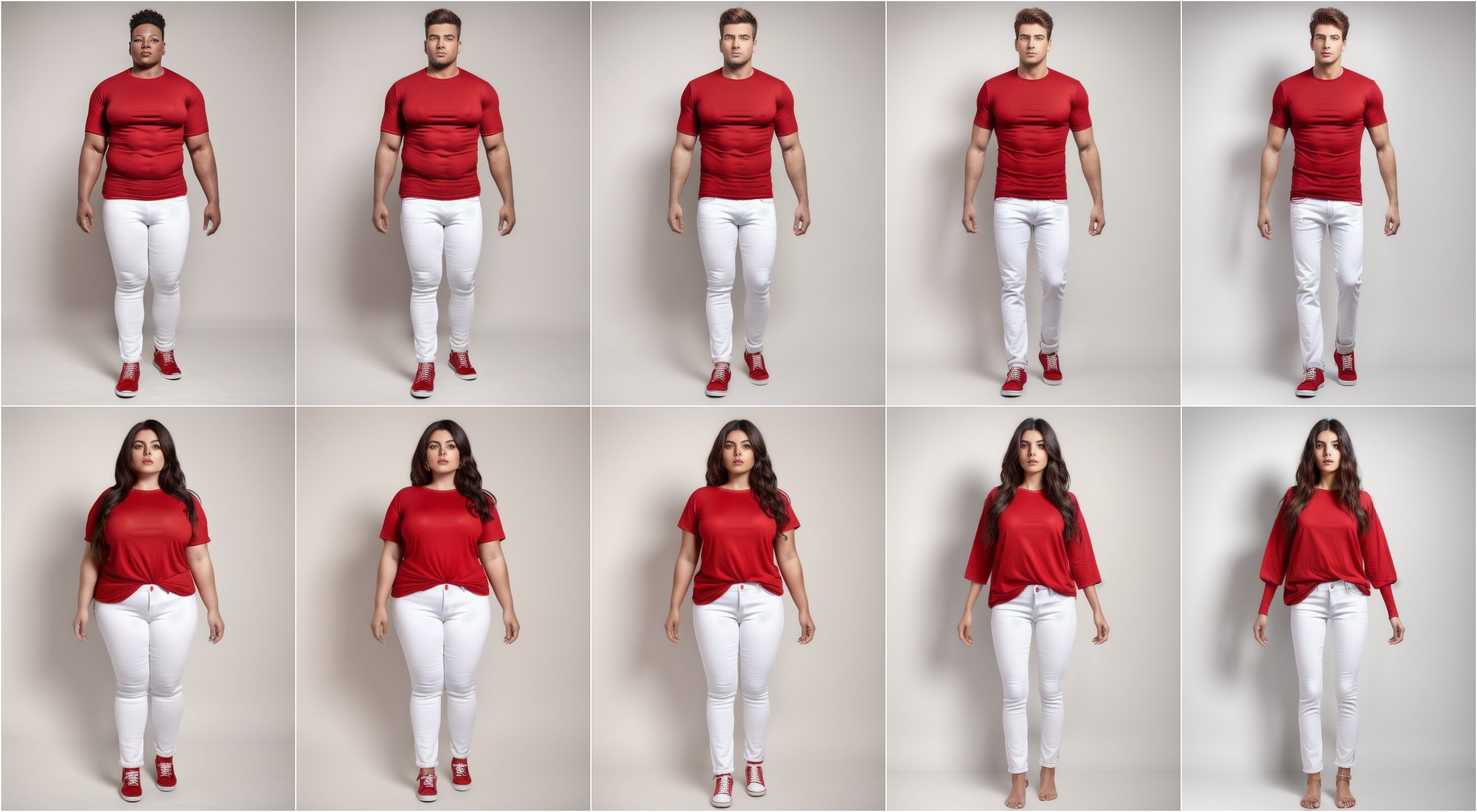}
    \caption{Varying shape parameters for fixed pose and prompt}
    \label{fig:shape-transition}
\end{figure}

\begin{figure}[t]
    \centering
    \includegraphics[width=\linewidth]{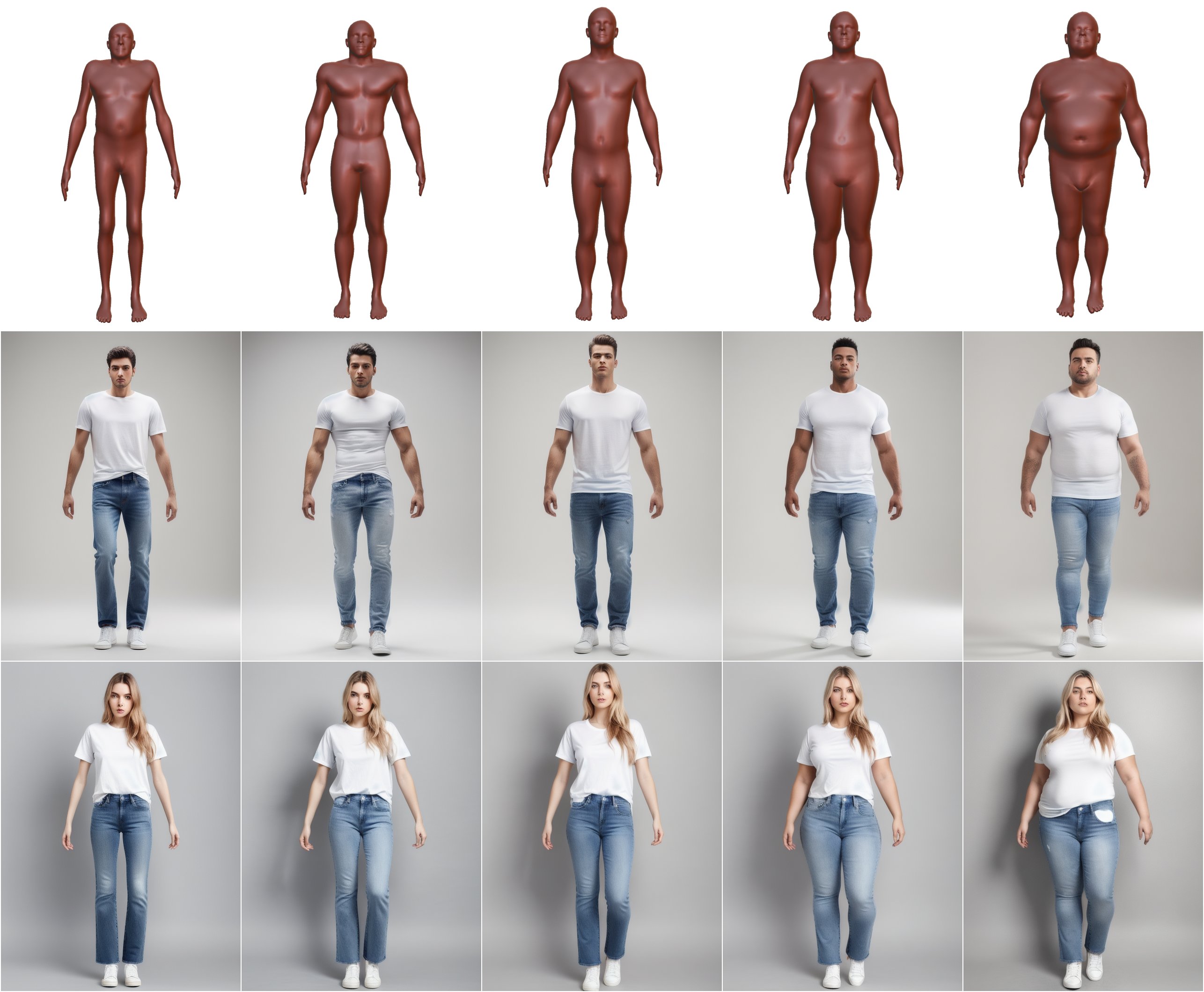}
    \caption{Sampling different shape parameters. The smpl shape (top row) is able to be accurately represented in the image of the clothed persons.}
    \label{fig:shapes-samples}
\end{figure}

\begin{table}[htbp]
    \centering
    \begin{tabular}{lcccc}
        \hline
        \multirow{2}{*}{\textbf{Method}} & \multicolumn{2}{c}{\textbf{PVET (SC)} $\downarrow$} & \multicolumn{2}{c}{\textbf{MPJPE (PA)} $\downarrow$} \\
        & \textbf{\small AS} & \textbf{\small AS Ext.} & \textbf{\small AS} & \textbf{\small AS Ext.} \\
        \hline
        ControlNet\cite{zhangAddingConditionalControl2023} & 16.1 & 20.7 & 148.8 & 98.5 \\
        T2I-Adapter\cite{mouT2IAdapterLearningAdapters2023} & 15.8 & 20.8 & \textbf{118.7} & \underline{93.6}  \\
        Champ\cite{shenhaozhuChampControllableConsistent2024} & - & 17.6 & - & \textbf{83.0} \\
        $C_\text{SMPL}$-ft-attn + CN & 14.9 & 16.0 & 135.3 & 98.1 \\
        $C_\text{SMPL}$-ft-attn + T2I & \underline{14.3} & \textbf{15.2} & \underline{119.7} & 94.1 \\[0.5ex]
        \hdashline
        \textit{Ablation Variants:} \\
        \quad $C_\text{SMPL}$-ft-all + CN & 14.8 & 16.5 & 135.5 & 98.6 \\
        \quad  $C_\text{SMPL}$-ft-all + T2I & \textbf{14.2} & \underline{15.7} & 120.1 & 94.5\\
        \hline
        GT Images \cite{SURREAL} & - & 12.1 & - & 75.3 \\
        \hline
    \end{tabular}
    \caption{SMPL shape and pose generation accuracy in millimeters. We show the scale and translation corrected \ac{PVET} and \ac{MPJPE-PA} between input SMPL params and SMPL meshes extracted using HierProb \cite{sengupta2021hierarchical} on the images generated from our evaluation set. AS-Ext uniformly samples body shapes w.r.t obesity and its SMPL models are rendered from a fixed frontal perspective enabling comparison with Champ~\cite{shenhaozhuChampControllableConsistent2024}.}
    \label{tab:pose_shape_comparison}
\end{table}

\begin{figure}
    \centering
    \includegraphics[width=\linewidth]{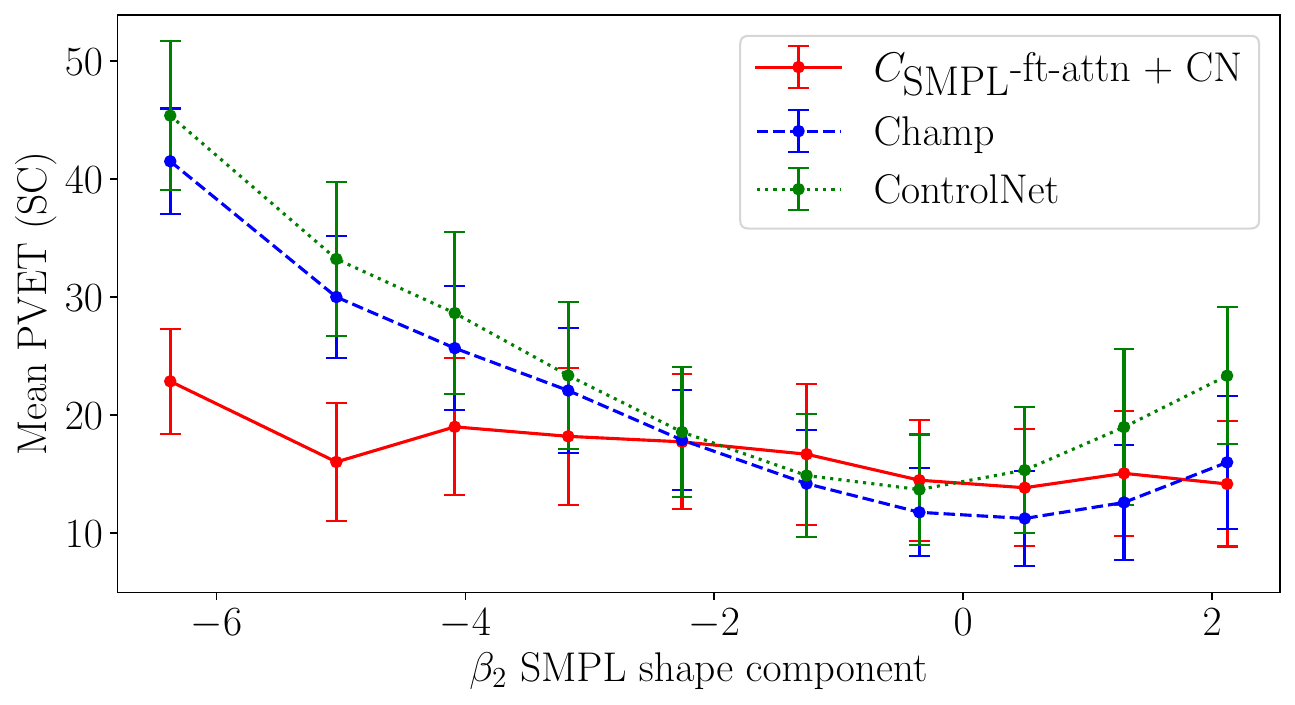}
    \caption{SMPL shape accuracy for extreme shapes. We show Per-Vertex-Error in T-Pose (PVET) for different shape extremes. The second shape component ($\beta_2$) strongly correlates with obesity. Error bars show the standard deviation.}
    \label{fig:pvetsc_vs_obesity}
\end{figure}

\input{fig_comparisonCNet}

\subsection{Shape and Pose Accuracy}

We assess pose and shape accuracy by generating images for an evaluation set of SMPL models and  measuring the distance of predicted SMPL parameters to ground-truth.

\textbf{Study Setup.}
We first created an evaluation dataset of 5000 SMPL models, combining poses from the AIST~\cite{AIST} dance dataset with shapes from the SURREAL~\cite{SURREAL} dataset. This combination dataset (\textbf{AS}) addresses the limitations of each dataset: AIST provides diverse poses from different viewpoints but lacks shape diversity, while SURREAL offers varied shapes but limited pose diversity. Using AS, we generated images using our proposed approach, its configurations, and baselines ControlNet and T2I-Adapter. Using \ac{HPH}~\cite{sengupta2021hierarchical}, we estimated the SMPL models from the generated images and compared these to the original ground-truth SMPL models.

SURREAL contains real body shape measurements, which are approximately normal distributed around a mean slim shape, with very limited samples for obese body types.
To fairly evaluate model performance on all body types, we create an extended dataset (AS-Ext) with added samples for obese shapes.
Please refer to the supplementary material for details on evaluation dataset creation.
Furthermore, to enable comparison with the state-of-the-art approach for SPML-based reference-image control Champ~\cite{shenhaozhuChampControllableConsistent2024}, we rendered the SMPL models in AS-Ext in a full-frame frontal view to align inputs of our method and Champ in absence of the ground-truth camera matrices. %
The fixed-view rendered SMPL meshes additionally allow us to establish a baseline noise level for AS-Ext by direct application of \ac{HPH}.

\textbf{Metrics.} 
For pose accuracy, we measure the \acl{MPJPE-PA}. 
For shape accuracy, we measure the scale and translation corrected  \acl{PVET}~\cite{sengupta2020synthetic}. All values are given in mm.

\textbf{Results.} Our results (\cref{tab:pose_shape_comparison}) show significant improvements in shape accuracy over ControlNet and T2I-Adapter, and Champ on AS-Ext.
Stronger deviation from the mean shape, e.g., obese shapes, degrade the accuracy of ControlNet and Champ, as shown in \cref{fig:pvetsc_vs_obesity}, while our method achieves relatively constant performance. Interestingly, SMPL-conditioned Champ~\cite{shenhaozhuChampControllableConsistent2024} shows only marginal improvement over the shape-unaware ControlNet in such cases,  while our model faithfully adheres to the conditioning.
T2I-Adapter demonstrates better adherence to poses than ControlNet, reflected in lower \ac{MPJPE-PA} values, which is consistent when adding T2I-guidance to our SMPL-finetuned models. Adding SMPL-guidance enhances pose adherence in ControlNet-generated images, while remaining on-par with T2I-Adapter.

\begin{figure}[htbp]
    \centering
    \begin{minipage}[t]{\linewidth}
        \centering
        \begin{subfigure}[t]{0.245\textwidth}
            \centering
            \includegraphics[width=\textwidth]{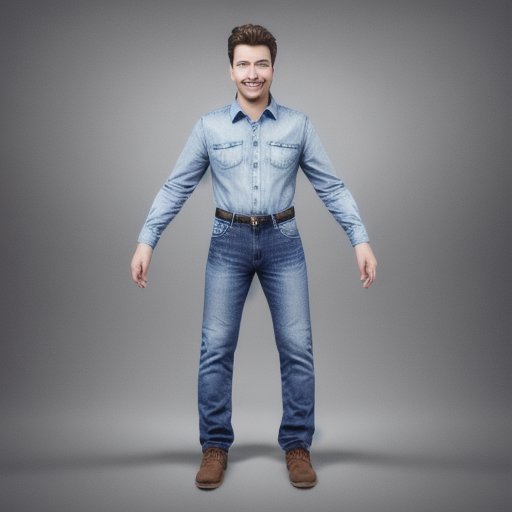}
            \subcaption{Ref. Image}
            \label{fig:ref_img}
        \end{subfigure}%
        \hfill
        \begin{subfigure}[t]{0.245\textwidth}
            \centering
            \includegraphics[width=\textwidth]{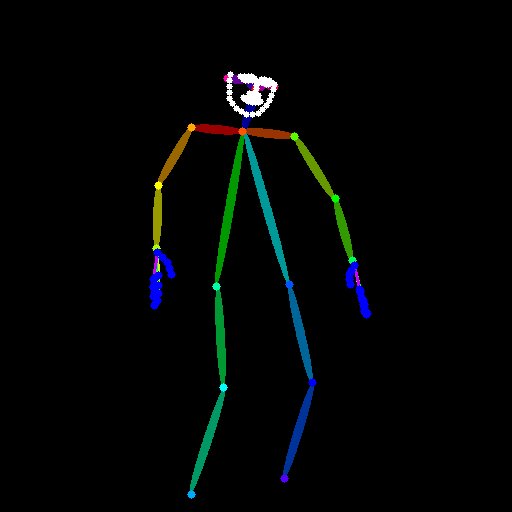}
            \subcaption{Pose Map}
            \label{fig:pose_map}
        \end{subfigure}
        \begin{subfigure}[t]{0.245\textwidth}
            \centering
            \includegraphics[width=\textwidth]{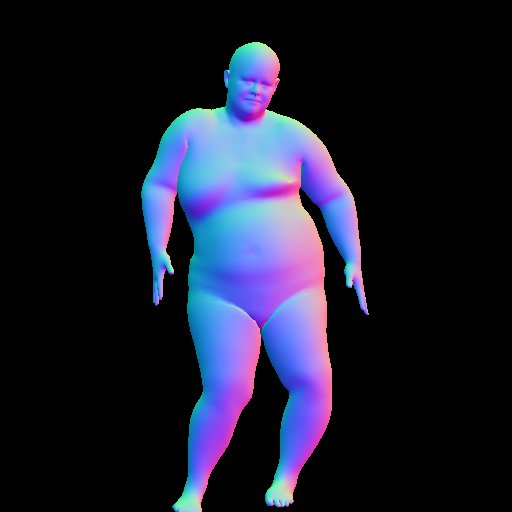}
            \subcaption{Normal Map}
            \label{fig:normal_map}
        \end{subfigure}%
        \hfill
        \begin{subfigure}[t]{0.245\textwidth}
            \centering
            \includegraphics[width=\textwidth]{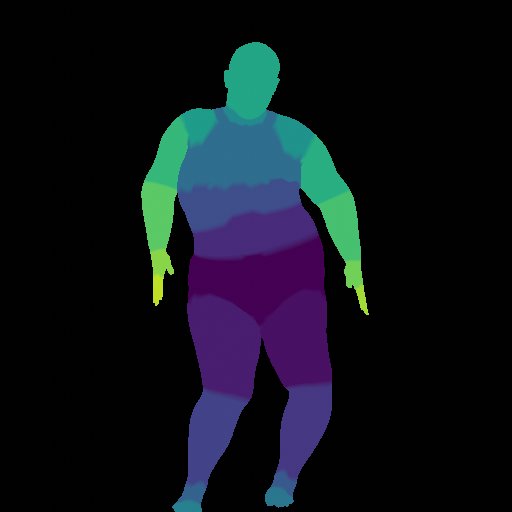}
            \subcaption{Semantic Map}
            \label{fig:semantic_map}
        \end{subfigure}
    \end{minipage}%
    \\
    \begin{minipage}[t]{\linewidth}
        \centering
        \begin{subfigure}[t]{0.49\textwidth}
            \centering
            \includegraphics[width=\textwidth]{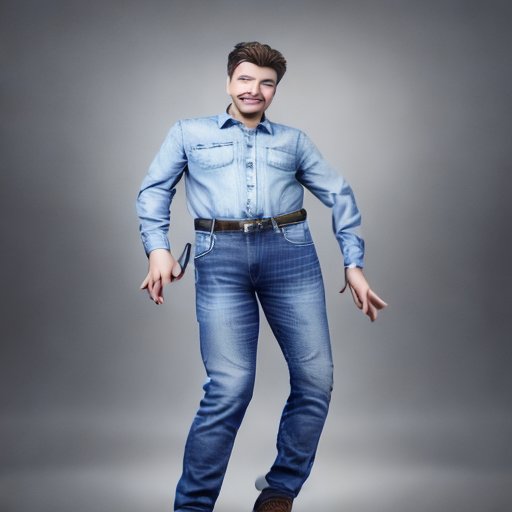}
            \subcaption{\centering SMPL-based control of reference image with Champ~\cite{shenhaozhuChampControllableConsistent2024}}
            \label{fig:champ_output}
        \end{subfigure}
        \begin{subfigure}[t]{0.49\textwidth}
            \centering
            \includegraphics[width=\textwidth]{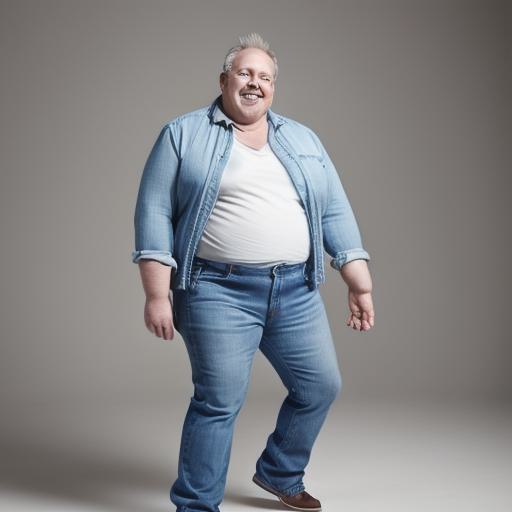}
            \subcaption{\centering SMPL-based text-to-image generation with ours}
            \label{fig:ours_output}
        \end{subfigure}
    \end{minipage}
    \caption{Comparison to SMPL-based reference image control using Champ~\cite{shenhaozhuChampControllableConsistent2024}. Champ uses a SMPL-derived pose map (b), normal map (c), semantic map (d), depth map \& mask (not shown) to manipulate a reference image (a). Our reference-free method (f) uses pose map (b) and SMPL parameters as inputs.}
    \label{fig:semantic_maps_champ}
\end{figure}

\subsection{Visual Comparisons}
\label{subsec:visualcomparison}

In \cref{fig:shape-transition}, we transition between two SMPL shape configurations, primarily affecting mid-body fat distribution, to illustrate a change from overweight to slim. Notice the background and person's appearance, aside from weight, remain stable with a fixed seed, prompt, and pose.
In \cref{fig:shapes-samples}, we sample different shape parameters to showcase various body proportions for the same prompt: ``a man/woman with a white t-shirt and jeans in front of a neutral background.'' 
In \cref{fig:comparison_ours_controlnet}, we compare our method to ControlNet and T2I. Using a pose from AIST \cite{AIST}, we apply slim and overweight shapes to generate images. For ControlNet and T2I, we include ``chubby'' in the prompt to create the overweight image. The rows are organized with prompts ``a man hiking'' and ``still of a football player in a FIFA game.'' Our method consistently adheres to shape information, whereas ControlNet's and T2I's interpretation of ``chubby'' varies. For instance, the first row shows a significantly overweight person, while the shape of the football player is only minimally altered. Using other words signifying overweight creates similar results. Additionally, our method shows improved pose and background stability, as verified by an experiment on masked-out foregrounds and LPIPS measurements (see supplemental material). We also observe that while metrically T2I achieves slightly better results for both the original and combined with $C_{\text{SMPL}}$ (\cref{tab:merged_comparison,tab:pose_shape_comparison}), it is more prone to body artifacts than its ControlNet counterpart.
In \cref{fig:semantic_maps_champ} we show an example from Champ\cite{shenhaozhuChampControllableConsistent2024}, and compare to ours in terms of shape adherence. We generate a reference image with SD of frontal view a person in neutral pose.
Champ then receives semantic maps derived from the SMPL mesh viewed from the same perspective as the reference image (\cref{fig:semantic_maps_champ}). Our approach receives the SMPL model in parameter form and generates content according to a prompt instead of a reference image. Visually, ours is able to achieve a better and more natural body shape adherence than Champ.

\subsection{Animation}
Our method can also be applied to generate animated sequences, by using the control-signal to drive frame transistion-consistent animations using modified versions such as AnimateDiff \cite{guoAnimateDiffAnimateYour2024}. We showcase examples in the supplemental material.

%% file: fig_comparisonCNet.tex
\newcommand{\comparisonfigsize}{0.124\textwidth}

\begin{figure*}[htbp]
    \centering
    \begin{subfigure}[b]{\comparisonfigsize}
        \includegraphics[width=\textwidth]{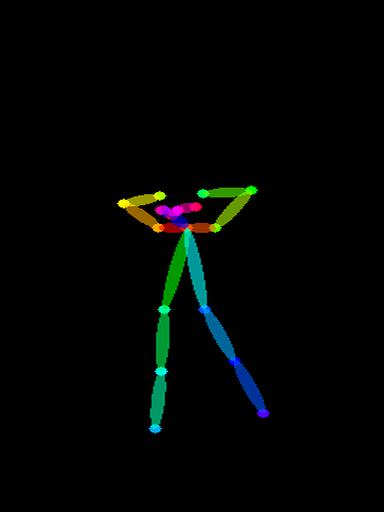}
    \end{subfigure}\hfill%
    \begin{subfigure}[b]{\comparisonfigsize}
        \includegraphics[width=0.5\textwidth, clip, trim={192 0 192 0}]{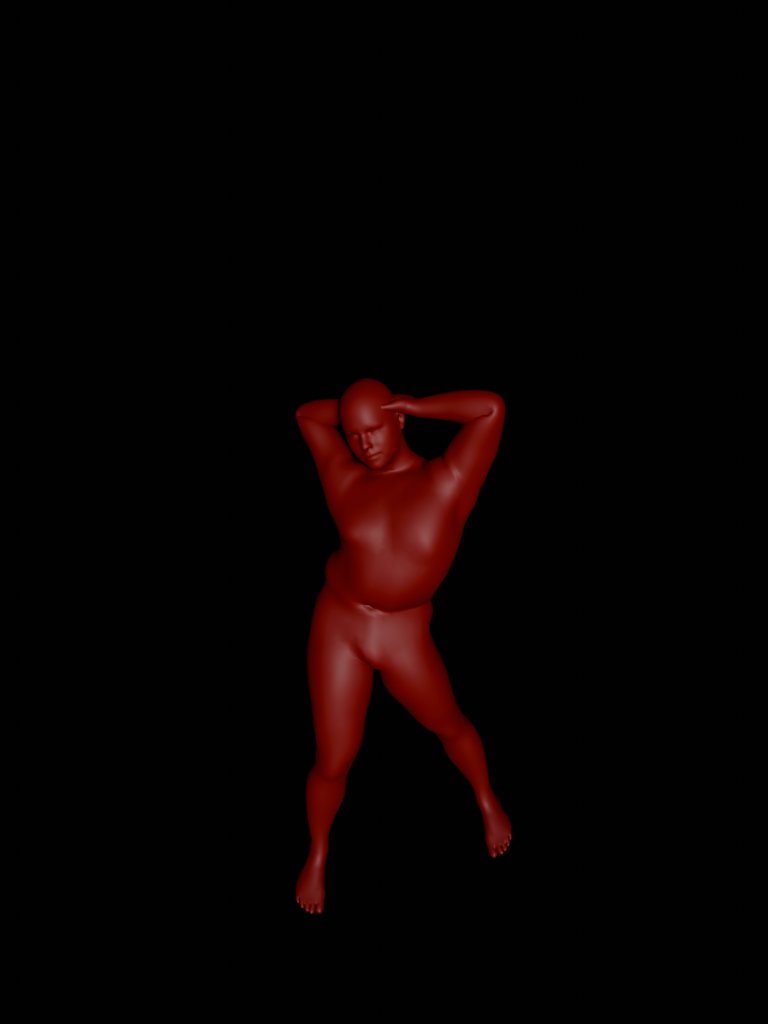}%
        \includegraphics[width=0.5\textwidth, clip, trim={192 0 192 0}]{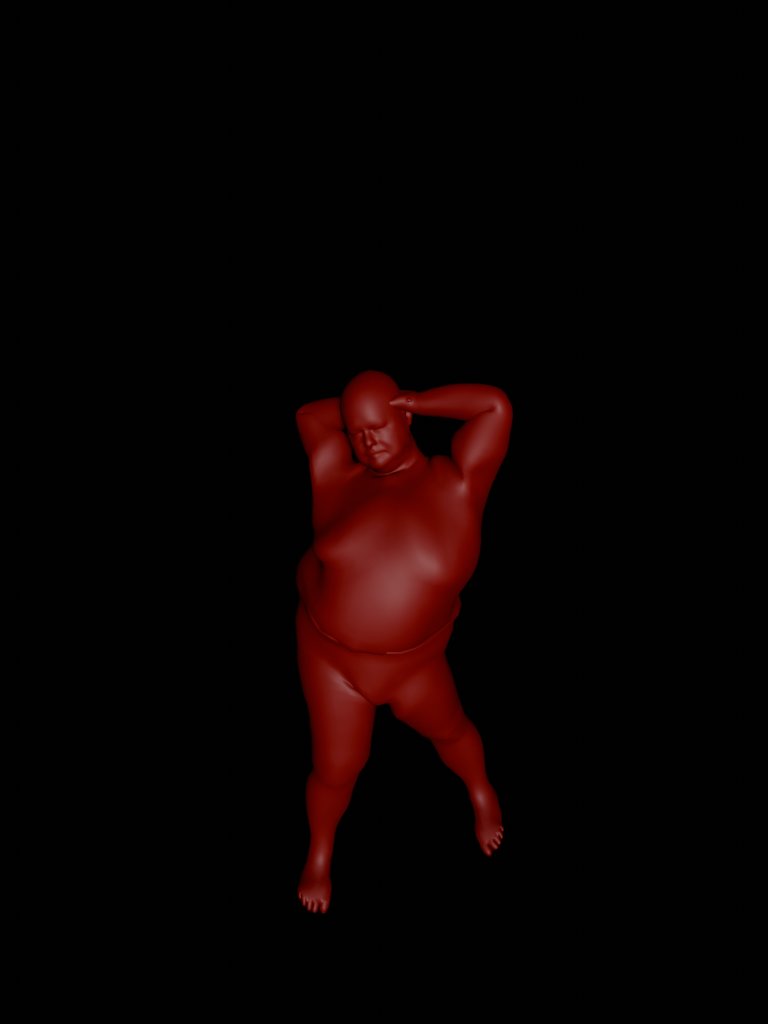}
    \end{subfigure}\hfill%
    \begin{subfigure}[b]{\comparisonfigsize}
        \includegraphics[width=\textwidth]{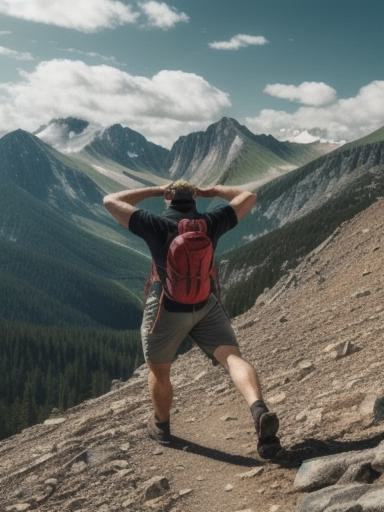}
    \end{subfigure}\hfill%
    \begin{subfigure}[b]{\comparisonfigsize}
        \includegraphics[width=\textwidth]{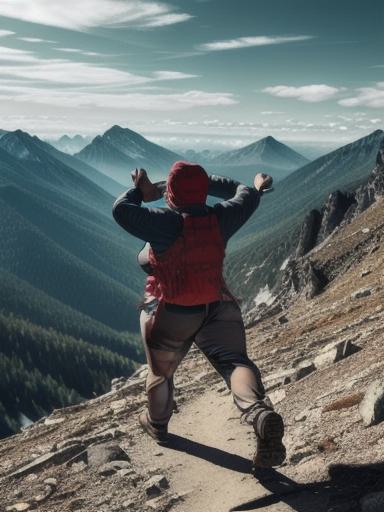}
    \end{subfigure}\hfill%
    \begin{subfigure}[b]{\comparisonfigsize}
        \includegraphics[width=\textwidth]{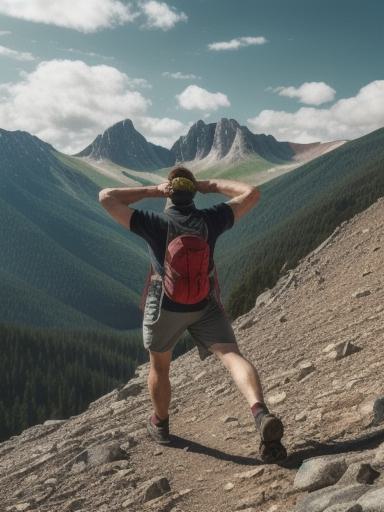}
    \end{subfigure}\hfill%
    \begin{subfigure}[b]{\comparisonfigsize}
        \includegraphics[width=\textwidth]{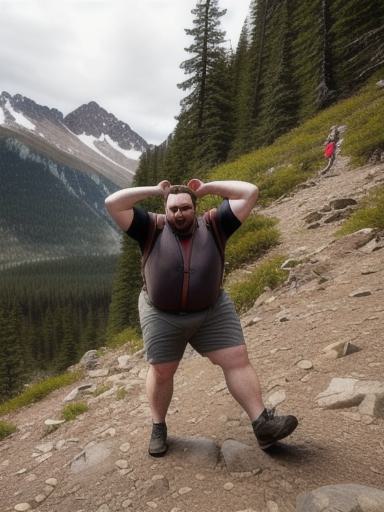}
    \end{subfigure}\hfill%
    \begin{subfigure}[b]{\comparisonfigsize}
        \includegraphics[width=\textwidth]{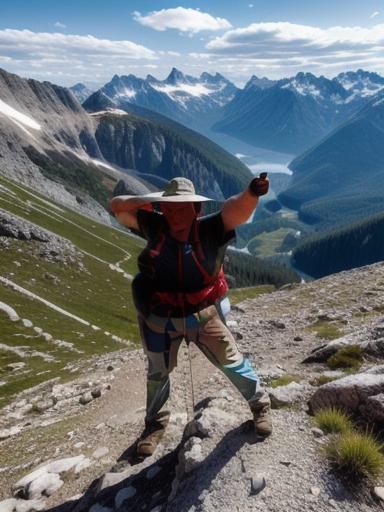}
    \end{subfigure}\hfill%
    \begin{subfigure}[b]{\comparisonfigsize}
        \includegraphics[width=\textwidth]{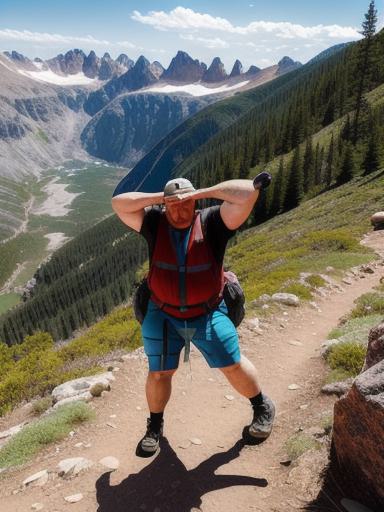}
    \end{subfigure}

    \begin{subfigure}[b]{\comparisonfigsize}
        \includegraphics[width=\textwidth]{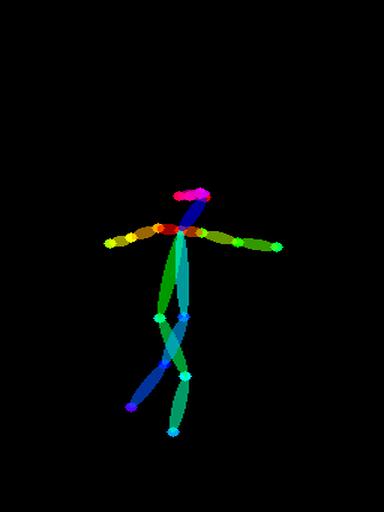}
        \caption{Pose}
    \end{subfigure}\hfill%
    \begin{subfigure}[b]{\comparisonfigsize}
        \includegraphics[width=0.5\textwidth, clip, trim={192 0 192 0}]{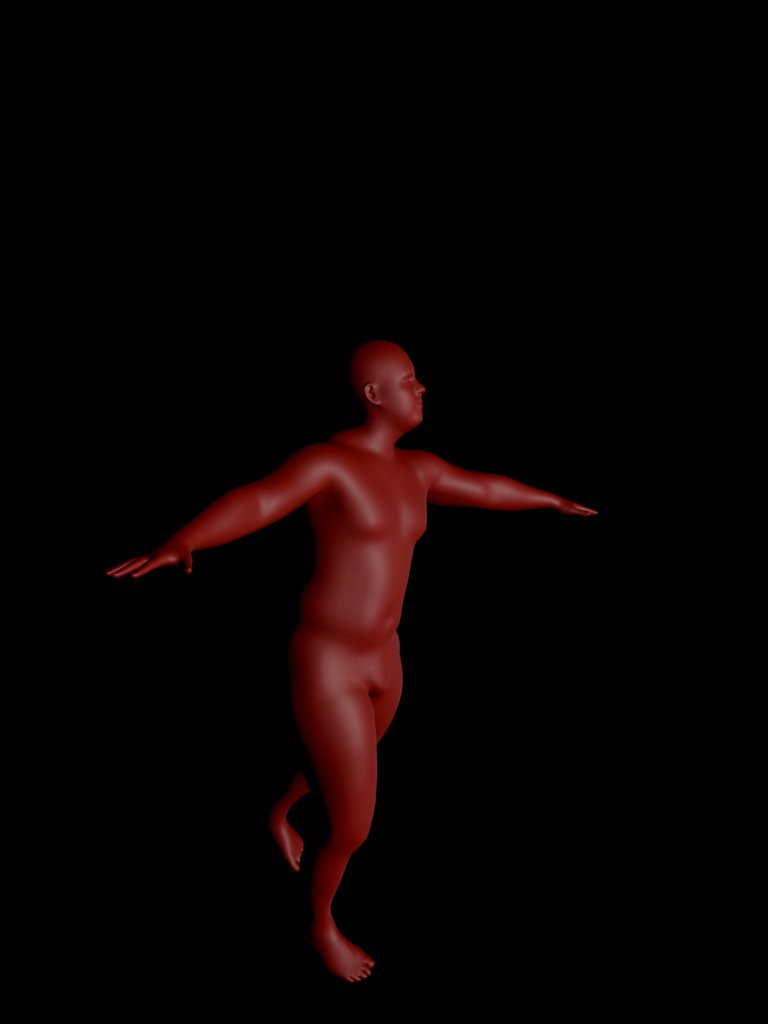}%
        \includegraphics[width=0.5\textwidth, clip, trim={192 0 192 0}]{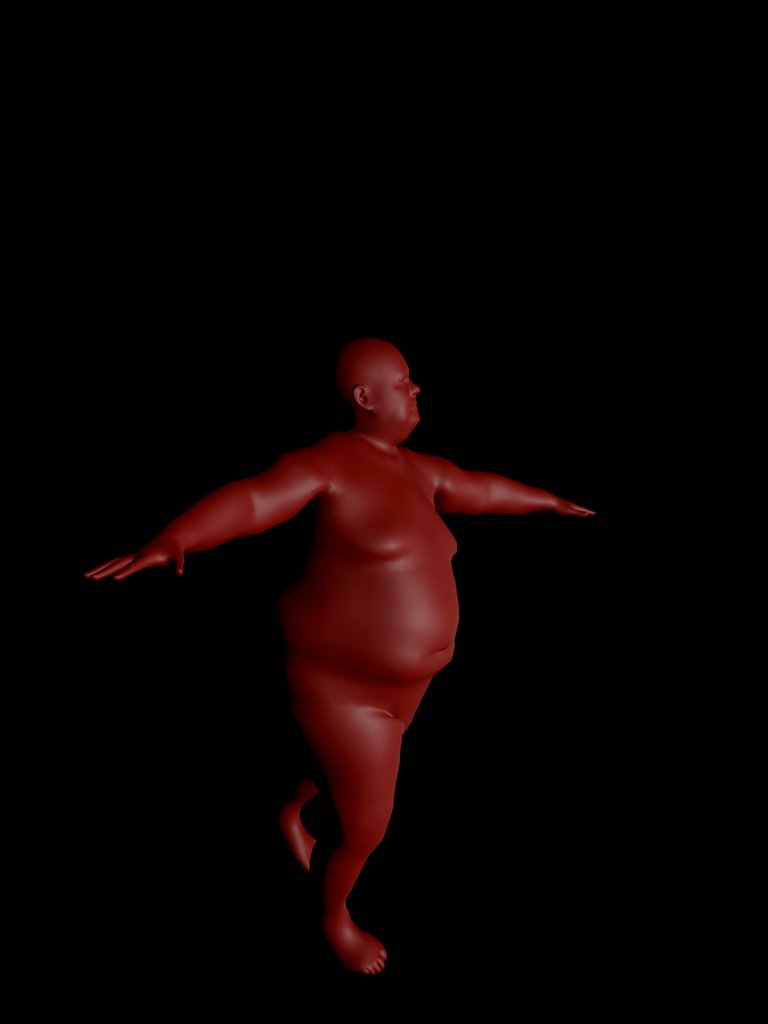}
        \caption{SPML $s_1, s_2$}
    \end{subfigure}\hfill%
    \begin{subfigure}[b]{\comparisonfigsize}
        \includegraphics[width=\textwidth]{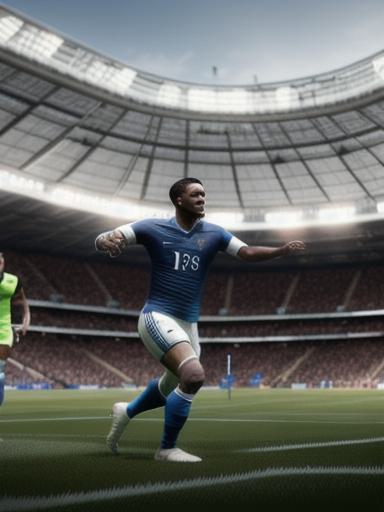}
        \caption{$\text{Ours}_\text{CN}(s_1)$ }
    \end{subfigure}\hfill%
    \begin{subfigure}[b]{\comparisonfigsize}
        \includegraphics[width=\textwidth]{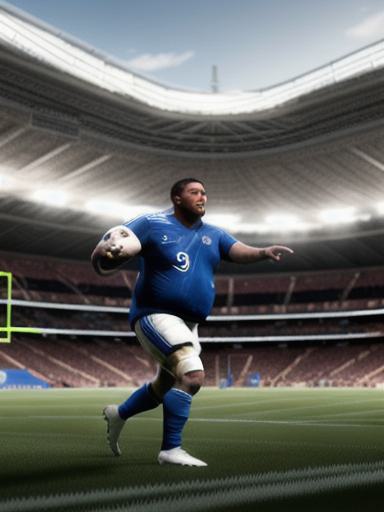}
        \caption{$\text{Ours}_\text{CN}(s_2)$}
    \end{subfigure}\hfill%
    \begin{subfigure}[b]{\comparisonfigsize}
        \includegraphics[width=\textwidth]{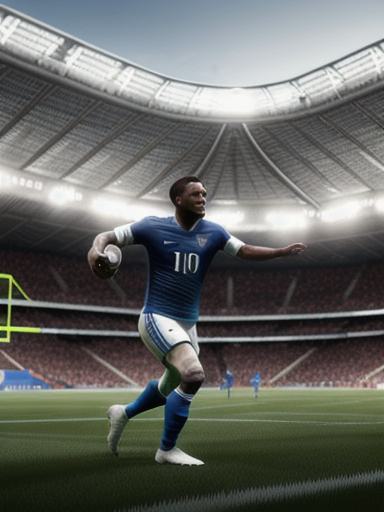}
        \caption{CN slim}
    \end{subfigure}\hfill%
    \begin{subfigure}[b]{\comparisonfigsize}
        \includegraphics[width=\textwidth]{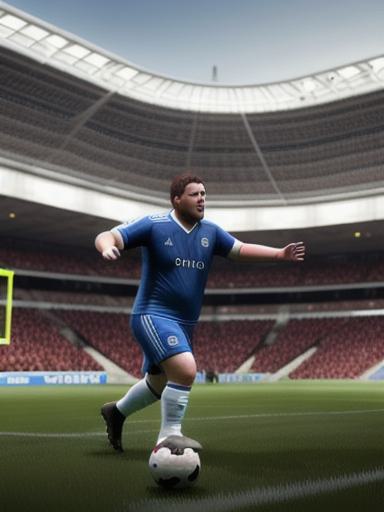}
        \caption{CN not slim}
    \end{subfigure}\hfill%
    \begin{subfigure}[b]{\comparisonfigsize}
        \includegraphics[width=\textwidth]{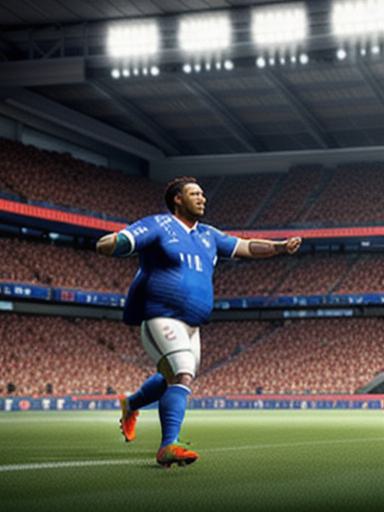}
        \caption{$\text{Ours}_\text{T2I}(s_2)$}
    \end{subfigure}\hfill%
    \begin{subfigure}[b]{\comparisonfigsize}
        \includegraphics[width=\textwidth]{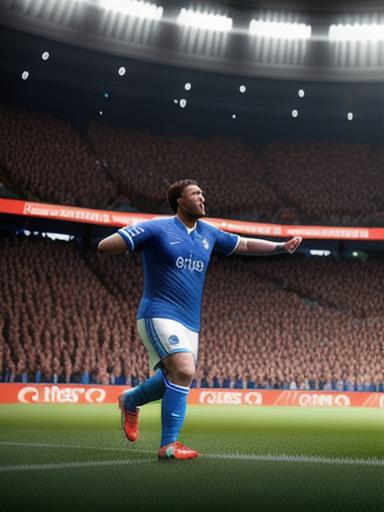}
        \caption{T2I not slim}
    \end{subfigure}

    \caption{Comparison of pose and body shape control. We compare ours, ControlNet \cite{zhangAddingConditionalControl2023} (CN) and T2I \cite{mouT2IAdapterLearningAdapters2023} using slim and overweight body shapes (for ours) or prompts (for CN and T2I). Note that $\text{Ours}_\text{CN}$ refers to $C_\text{SMPL}$ + CN, and $\text{Ours}_\text{T2I}$ refers to $C_\text{SMPL}$ + T2I.  Overall our method displays better stability and accuracy of shapes. }
    \label{fig:comparison_ours_controlnet}
\end{figure*}

%% file: guidance_ablations.tex
\subsection{Ablating Guidance Vector Composition}
\label{subsec:guidanceablation}

\begin{figure}[t]
    \setlength\tabcolsep{0.5pt}
    \adjustboxset{width=\linewidth,valign=c}
    {\renewcommand{\arraystretch}{1}%
    \centering
    \hspace{-5pt}\begin{tabular}{@{}l p{0.38\linewidth} p{0.19\linewidth}<{\centering} p{0.38\linewidth}<{\centering}}
        & \footnotesize{$w_1$} \hspace{0.14cm} \footnotesize{0.0} \hspace{1cm} \footnotesize{7.5} \hspace{1.0cm} & \footnotesize{0.0} & \footnotesize{0.0} \hspace{1.0cm} \footnotesize{7.5} \\
        \rotatebox[origin=c]{90}{\footnotesize{\hspace{0.7cm} 7.5 \hspace{1.55cm} 0.0 \hspace{0.35cm} $w_2$}}
        & \includegraphics[width=\linewidth]{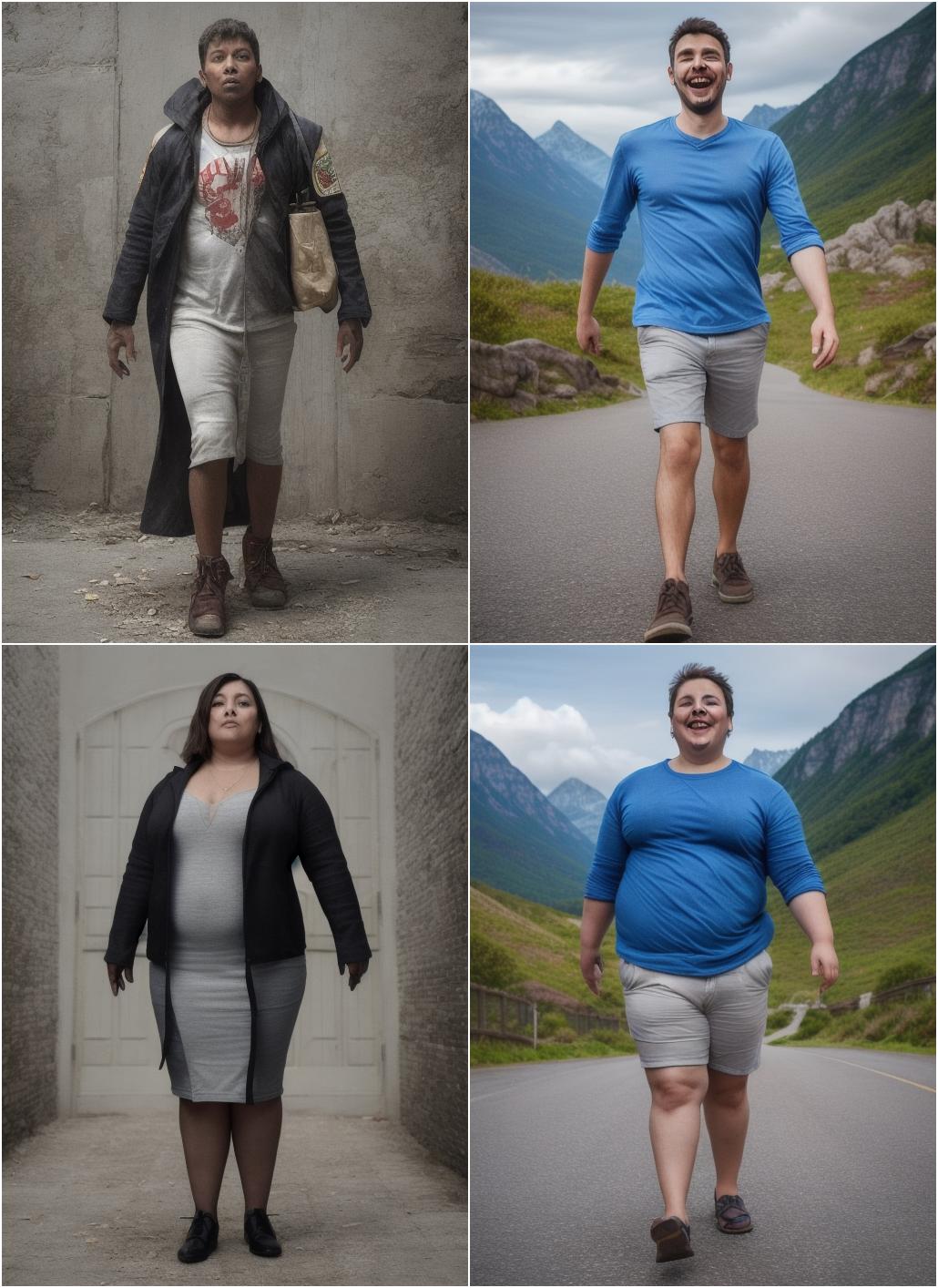}
        & \includegraphics[width=\linewidth]{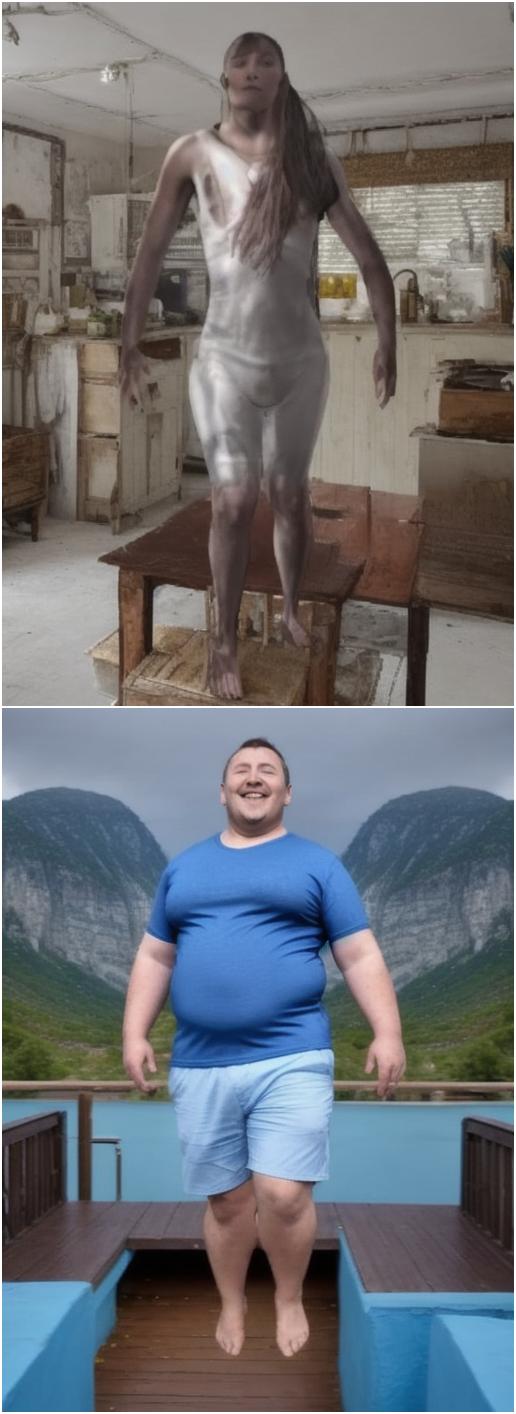}
        & \includegraphics[width=\linewidth]{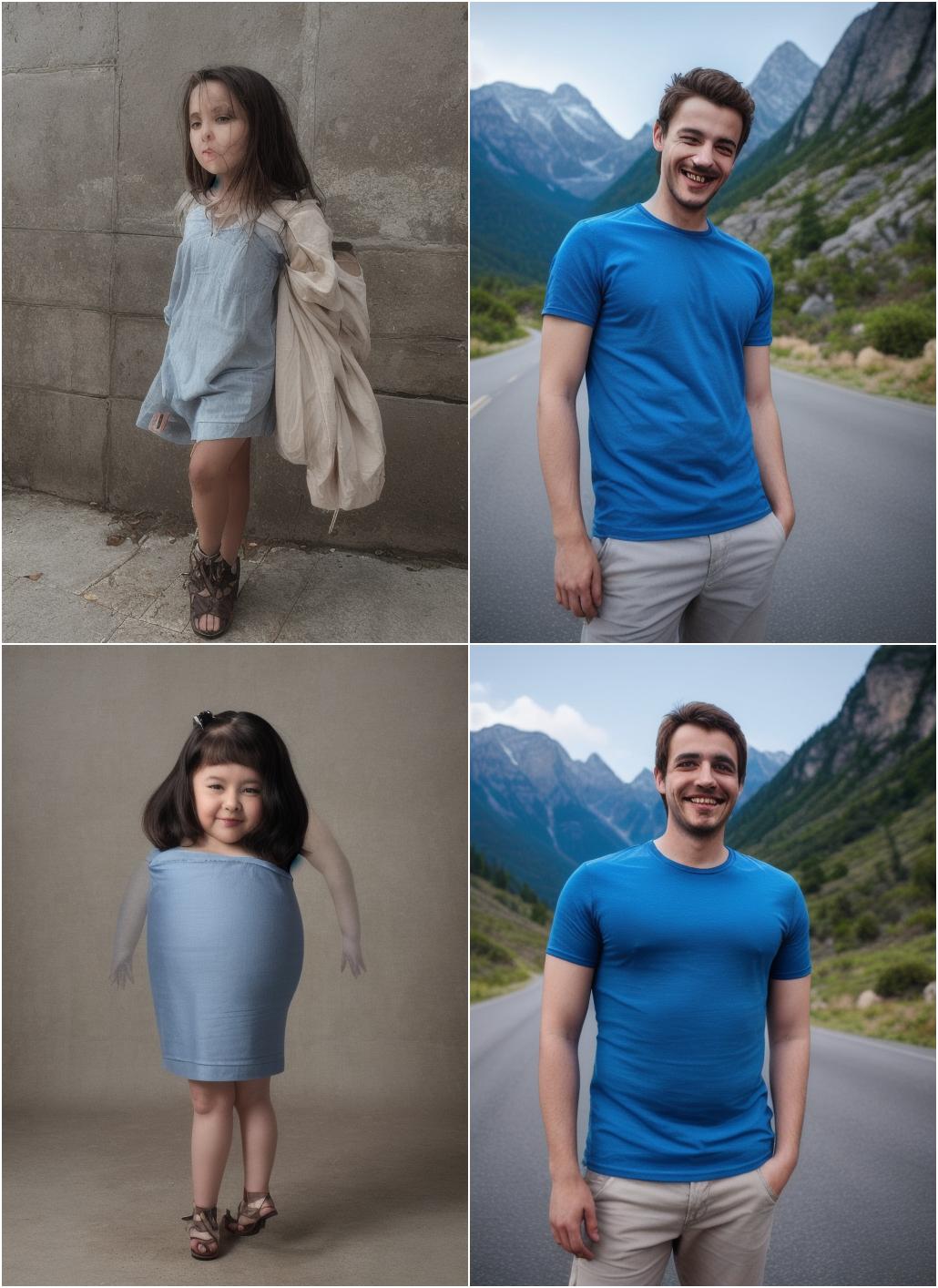} \\
        \vspace{1pt} \\
        & (a) Full Approach & (b) No DA & (c) No pose guidance
    \end{tabular}
    }
    \caption{Comparing guidance configurations and the isolated effects of individual guidance vectors. (a) The full approach: $\epsilon_{\text{SD}}(\emptyset, \boldsymbol{c}_o) + w_1 \boldsymbol{\vec{g}}_{c_p,c_o} + w_2 \boldsymbol{\vec{g}}_{c_s,c_o}$ (b) Ablated model w/o domain adaptation (DA): $\epsilon_{\text{Syn}}(\emptyset, \emptyset, \boldsymbol{c}_o) + w_2 \boldsymbol{\vec{g}}_{c_p,c_s,c_o}$ (c) No pose guidance (i.e., using SD instead of  $C_\text{SD}$): $\epsilon_{\text{SD}}(\emptyset) + w_1 \boldsymbol{\vec{g}}_{c_p} + w_2 \boldsymbol{\vec{g}}_{c_s,c_o}$}
    \label{fig:guidance_ablations}
\end{figure}

Our classifier-free guidance-based approach consists of a composition of multiple models with different possible input condition configurations. In the following, we explore ablations of \cref{eq:cfg_equations}.
In addition to the two guidance vectors $\boldsymbol{\Vec{g}_{c_p,c_o}}$ and $\boldsymbol{\Vec{g}_{c_s,c_o}}$ we ablate configurations using guidance vectors without pose-conditioning ($\boldsymbol{\Vec{g}_{c_p}}$) and without domain adaption, for which we use the ft-extra-attn which receives prompt in addition to pose and body ($\boldsymbol{\Vec{g}_{c_p, c_s, c_o}}$), defined as:
\begin{align*}
    \boldsymbol{\Vec{g}_{c_p}} &= \epsilon_{\text{SD}}(\boldsymbol{c}_p) - \epsilon_{\text{SD}}(\emptyset) \\
    \boldsymbol{\Vec{g}_{c_p, c_s, c_o}} &= \epsilon_{\text{Syn}}(\boldsymbol{c}_p, \boldsymbol{c}_s, \boldsymbol{c}_o) - \epsilon_{\text{Syn}}(\emptyset, \emptyset, \boldsymbol{c}_o)
\end{align*}
 
 In \cref{fig:guidance_ablations} we evaluate the effects of the composed guidance vectors by generating image grids using the prompt "A man wearing a blue shirt with a happy expression in front of a scenic and cinematic environment, best quality, photography". We compare configurations by selectively setting guidance scales ($w_1,w_2$) to 0.0, effectively disabling guidance components.
\cref{fig:guidance_ablations}a shows the configuration of our full approach.
The first row and column illustrate the influence of isolated guidance vectors.
The relatively stable background when altering $w_2$ indicates independence between the vectors, allowing separate scaling of adherence to the text prompt and SMPL conditioning.
\cref{fig:guidance_ablations}b demonstrates the impact of not shifting to the original data distribution, effectively running original classifier-free guidance in the synthetic data domain.
The domain gap from synthetic data results in compromised visual fidelity, with flat textures and floating effects, thus demonstrating that classifier-free guidance within the fine-tuned model’s domain realizes both SMPL conditioning and the text prompt but at the cost of visual quality.
The importance of the original 2d pose-conditioned ControlNet is visualized in \cref{fig:guidance_ablations}c.
When running the inference only using the base Stable Diffusion U-Net, the shape and pose condition is not satisfactorily realized.
In Sections B.4 and B.5 of the supplemental material we further demonstrate the negative impact of prompt-conditions in $\epsilon_{\text{Syn}}$ and the visual influence of the guidance scales.

%% file: conclusion.tex
\subsection{Limitations} %
Our approach, while effective in many scenarios, has notable limitations in generating body shapes that conflict with the prompt’s content or implied context. 
For example, specifying athletic professions can in some cases neutralize the generation of obese body shapes. This issue is also present in the original SD model when specifying body weight via prompt (see \cref{subsec:visualcomparison}),  suggesting a bias in SD training data or the model's learned representations. %
Our approach struggles to reconcile such conflicting information, leading to less accurate or unintended results. Please refer to the supplementary material for an example.

\section{Conclusion}
\label{sec:conclusion}

Our proposed method enables conditional control and generation of diverse human shapes and poses in pretrained text-to-image diffusion models by fine-tuning a ControlNet-based architecture on a 3D human parametric model (SMPL). To address the issue of limited real-world data, we train on synthetic data. To overcome the domain gap from training on less realistic and diverse synthetic scenes, we propose a domain adaptation technique using guidance-based isolation and composition. %
Our results show that composing the isolated SMPL condition with a domain guidance network can satisfactorily adapt the visual appearance from the synthetic to the original LDM domain.  
The interchangeability of ControlNet and T2I-Adapter for guidance domain adaptation suggests the potential for generalization in composing isolated conditions and domain guidance networks, possibly allowing multiple synthetically-trained conditions to be composed together. Future research should explore the generalization of our guidance domain adaptation technique to other datasets, tasks, and domains.

%% file: background.tex
\section{Background}
\label{sec:background}

\textbf{Latent Diffusion Models.}
Diffusion models add increments of random noise to input data during a variance-preserving Markov diffusion process~\cite{sohl2015deep,ho2020denoising,song2020score}, and then learn to reverse the diffusion process by denoising to construct desired data samples. \acp{LDM}~\cite{rombach2022high} perform diffusion and denoising within a low-dimensional latent space $\boldsymbol{z}$, into which a Variational Autoencoder (VAE)\cite{kingma2013auto} encodes an image $\boldsymbol{z}_0 = \mathcal{E}(I)$ using a pretrained encoder $\mathcal{E}(\cdot)$. Formally, the noisy latent representation of  $\boldsymbol{z}_t$ is obtained by adding noise to $z_o$ in every step $t$ until $q(\boldsymbol{z}_T \mid \boldsymbol{z}_0)$ approximates a Gaussian distribution $\mathcal{N}(\boldsymbol{0},\boldsymbol{I})$.
In the denoising process, noise $\boldsymbol{\epsilon}_{\theta}(\boldsymbol{z}_t,t,\boldsymbol{c})$ is predicted for each timestep $t$ from $\boldsymbol{z}_t$ to $\boldsymbol{z}_{t-1}$. Here, $\boldsymbol{\epsilon}_{\theta}$ represents a noise predicting neural network, typically a U-Net~\cite{ronneberger2015u}, while $\boldsymbol{c}$ denotes conditioning information. The training loss $L$ minimizes the error between actual noise $\boldsymbol{\epsilon}\sim\mathcal{N}(\boldsymbol{0},\boldsymbol{I})$ and predicted noise $\boldsymbol{\epsilon}_{\theta}$:
\begin{equation}
\label{eq:diffusion_loss}
L = \mathbb{E}_{\mathcal{E}(I), \boldsymbol{c_{\text{p}}},\boldsymbol{\epsilon},t}\left[\omega(t) \lVert \boldsymbol{\epsilon}-\boldsymbol{\epsilon}_{\theta}(\boldsymbol{z_t}, t, \boldsymbol{c_{\text{p}}}) \rVert_{2}^{2} \right], t=1,...,T
\end{equation}
Here, $\omega(t)$ is a hyperparameter that adjusts the loss weighting at each timestep, and $c_{\text{p}}$ are text prompt embeddings from CLIP in the case of text-to-image models such as Stable Diffusion \cite{rombach2022high}. After training, the model can progressively denoise from $\boldsymbol{z}_T \sim \mathcal{N}(\boldsymbol{0},\boldsymbol{I})$ to $\boldsymbol{z}_0$ using a fast diffusion sampler~\cite{song2020denoising, lu2022dpm}, with the final $\boldsymbol{z}_0$ decoded back into the image space $I$ using a frozen decoder $\mathcal{D}(\cdot)$.

\textbf{ControlNet.}
ControlNet ~\cite{zhangAddingConditionalControl2023} integrates spatially localized, task-specific image conditions into LDMs. It does this by cloning the blocks of a pretrained LDM, retraining these blocks, and then adding them to the original model output using zero convolution. These ControlNet blocks $\mathcal{C}$ modify the latent output features  $\boldsymbol{f}$ of each U-Net decoder block:
\begin{equation}
\boldsymbol{f}_{\text{c}} = \boldsymbol{f} + \mathcal{Z}\left(\mathcal{C}(\boldsymbol{x}, \boldsymbol{c}_{\text{f}})
\right)
\label{eq:controlnet_feature}
\end{equation}
Here, $\mathcal{Z}$ denotes zero-convolution and $\boldsymbol{c}_{\text{f}}$ represents the task-specific spatial condition, which is incorporated in the fine-tuning of ControlNet blocks by minimizing the loss  \cref{eq:diffusion_loss} with $\boldsymbol{\epsilon}_{\theta}(\boldsymbol{z}_t,t,\boldsymbol{c_{\text{p}}},\{\boldsymbol{f}^i_{\text{c}}\})$, i.e., conditioned on all ControlNet block output features.

\textbf{SMPL model.}
The Skinned Multi-Person Linear (SMPL) model~\cite{SMPL:2015}, is widely used in computer graphics and vision for anatomically plausible and visually realistic human body deformations across various shapes and poses. SMPL combines a parametric shape space, capturing individual body shape variations, with a pose space encoding human joint positioning. The model uses low-dimensional parameters for pose, represented in joint angles $\theta_\text{SMPL} \in \mathbb{R}^{24 \times 3 \times 3}$, and shape, represented in principal shape variation directions $\beta_\text{SMPL} \in \mathbb{R}^{10}$, to produce a 3D mesh representation ($M \in \mathbb{R}^{3 \times N}$) with $N = 6890$ vertices. A vertex weight evaluates the relationships between vertices and body joints.

%% file: supplemental_content.tex
\input{suppl_fig_tablemacros}

\begin{figure}[htbp]
    \begin{subfigure}[b]{0.49\linewidth}
        \centering
        \includegraphics[width=\textwidth]{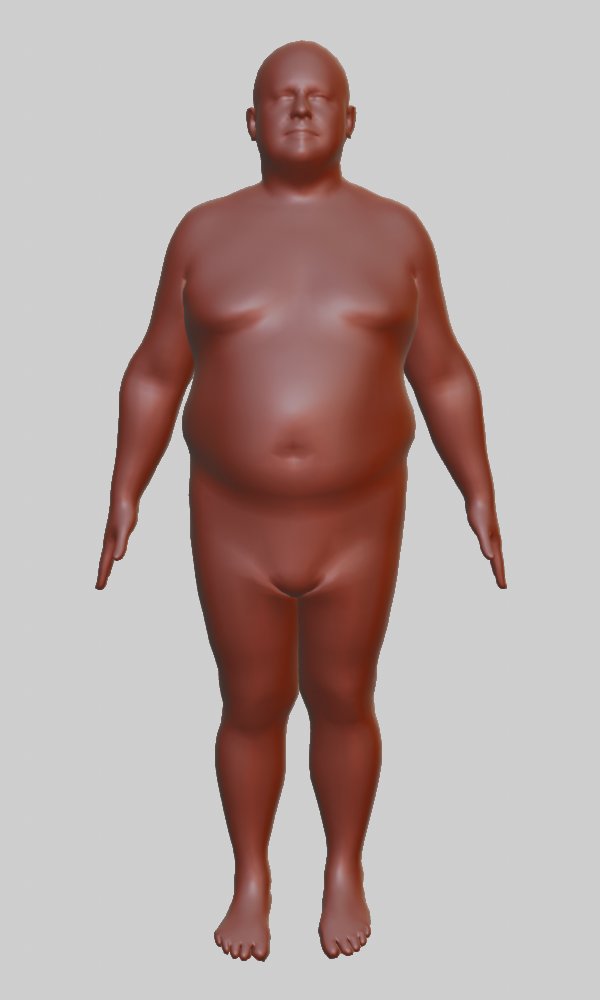}
        \caption{$\beta_2 = -4.0$}
    \end{subfigure}
    \begin{subfigure}[b]{0.49\linewidth}
        \centering
        \includegraphics[width=\textwidth]{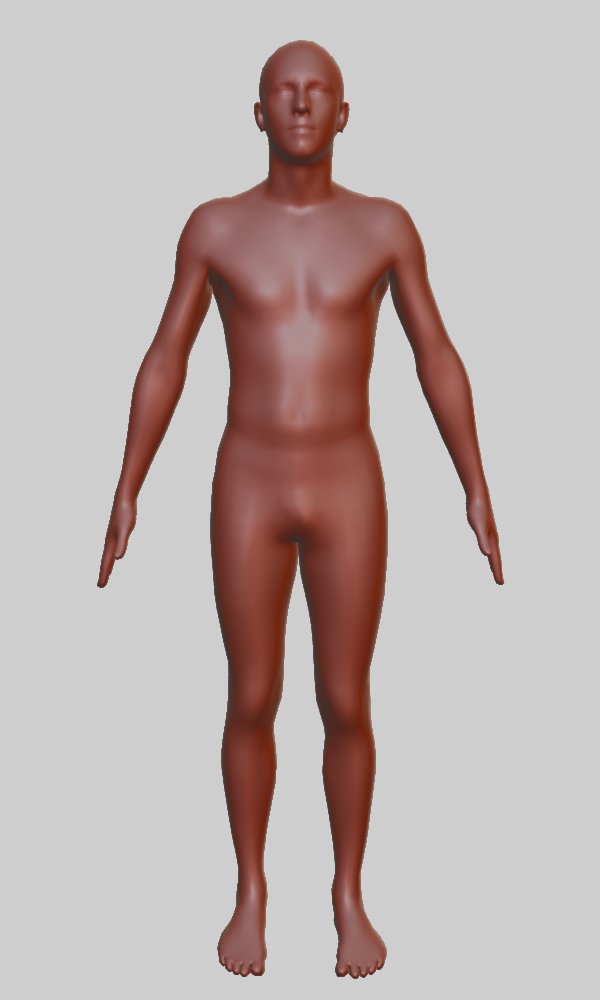}
        \caption{$\beta_2 = 2.0$}
    \end{subfigure}
    \caption{Varying the second shape component $\beta_2$ of the SMPL\cite{SMPL:2015} model. This component is highly correlated with body mass, which we use to evaluate the methods across different body types.}
    \label{fig:beta_component}
\end{figure}

\section{Evaluation Details and Additional Results}
\subsection{Shape and Pose Accuracy}
The SMPL\cite{SMPL:2015} model encodes body shape using shape components $\beta = \{\beta_1, \beta_2, ..., \beta_{10}\}$, derived through a principal component analysis of body meshes. For some of these components identifiable and distinct effects on body shape can be determined.
Specifically, $\beta_1$ component strongly correlates with overall height, while $\beta_2$ strongly correlates with body mass (\cref{fig:beta_component}).
The SURREAL~\cite{SURREAL} dataset provides body shape parameters collected from in-the-wild data.
Although the data is of high quality and diversity, it contains limited samples for overweight and obese body types, with only about 4\% of samples having a $\beta_2 \leq -2$. 
To fairly evaluate methods across body shapes, we augment the evaluation dataset (AS) to approximate a uniform shape distribution with respect to $\beta_2$.
Specifically, for our extended analysis dataset (AS-Ext) we generate 500 pose-shape pairs for each increment of 1 of $\beta_2$ values between -6 and -2, using random poses from AIST~\cite{AIST} and shapes with corresponding $\beta_2$ sampled from SURREAL~\cite{SURREAL}.

\begin{figure*}[t]
    \centering
        \begin{subfigure}[b]{0.19\textwidth}
        \centering
        \includegraphics[width=\textwidth]{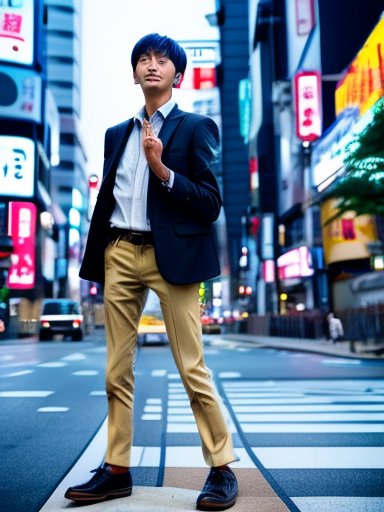}
    \end{subfigure}
    \begin{subfigure}[b]{0.19\textwidth}
        \centering
        \includegraphics[width=\textwidth]{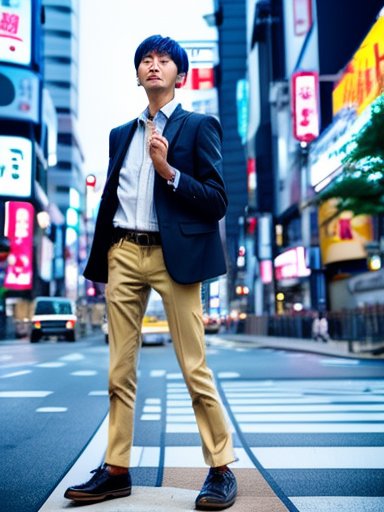}
    \end{subfigure}
    \begin{subfigure}[b]{0.19\textwidth}
        \centering
        \includegraphics[width=\textwidth]{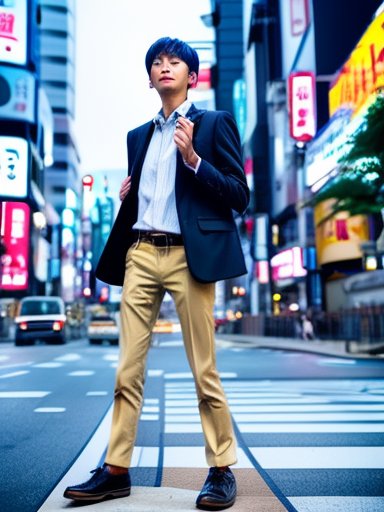}
    \end{subfigure}
    \begin{subfigure}[b]{0.19\textwidth}
        \centering
        \includegraphics[width=\textwidth]{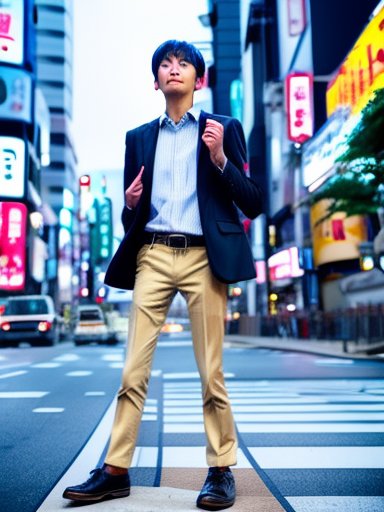}
    \end{subfigure}
    \begin{subfigure}[b]{0.19\textwidth}
        \centering
        \includegraphics[width=\textwidth]{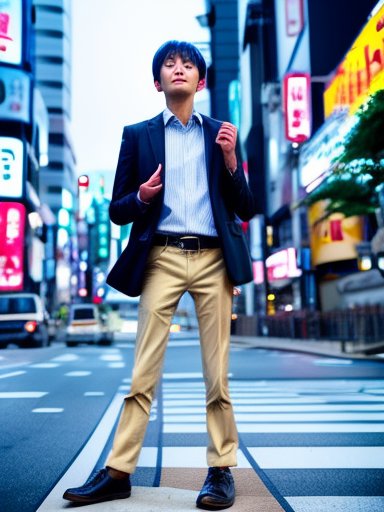}
    \end{subfigure}\\
    
    \begin{subfigure}[b]{0.19\textwidth}
        \centering
        \includegraphics[width=\textwidth]{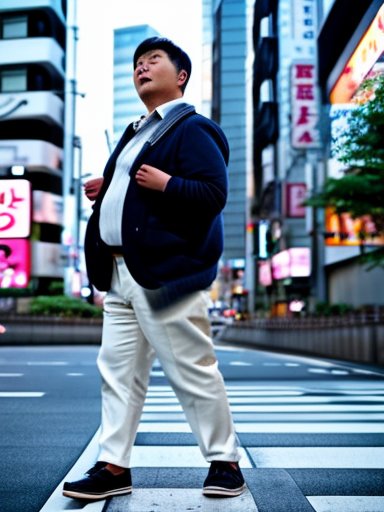}
        \caption{Frame 1}
    \end{subfigure}
    \begin{subfigure}[b]{0.19\textwidth}
        \centering
        \includegraphics[width=\textwidth]{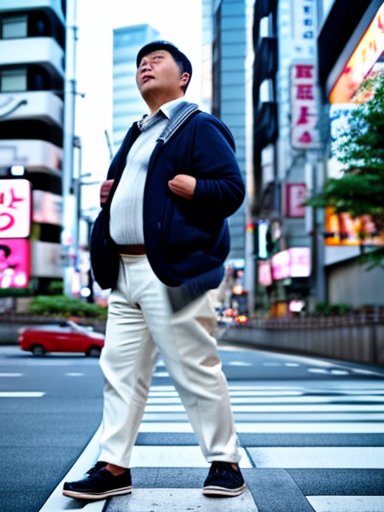}
        \caption{Frame 4}
    \end{subfigure}
    \begin{subfigure}[b]{0.19\textwidth}
        \centering
        \includegraphics[width=\textwidth]{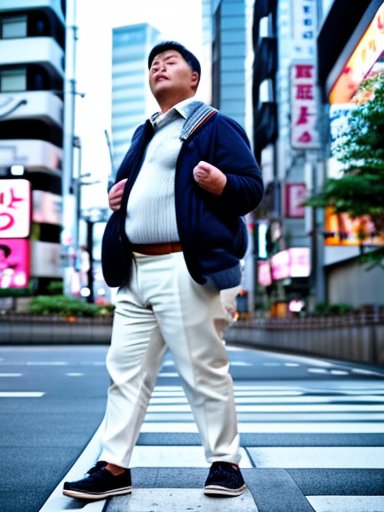}
        \caption{Frame 8}
    \end{subfigure}
    \begin{subfigure}[b]{0.19\textwidth}
        \centering
        \includegraphics[width=\textwidth]{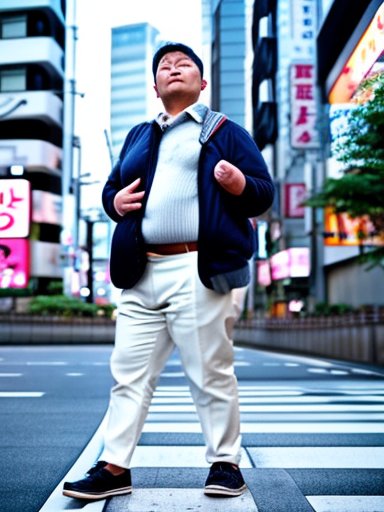}
        \caption{Frame 12}
    \end{subfigure}
    \begin{subfigure}[b]{0.19\textwidth}
        \centering
        \includegraphics[width=\textwidth]{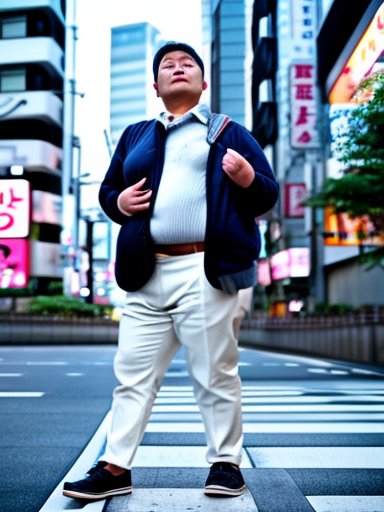}
        \caption{Frame 16}
    \end{subfigure}
    \caption{Frames from a SMPL-controlled AnimateDiff\cite{guoAnimateDiffAnimateYour2024} clip, for slim (top row) and overweight (bottom row) body types.}
    \label{fig:animation_frames}
\end{figure*}

\subsection{Background Stability}
To measure background stability, we generated images using three prompts for 100 poses sampled from AIST, resulting in 300 prompt+pose samples. We use two prompts from Fig. 6 of the main paper, and ``a man dancing in the desert''. For each sample, we generated images under both fit and not-fit shape conditions using our approach and ControlNet. We then used MaskRCNN \cite{he2017mask} to mask out the person in each image. By computing the average LPIPS \cite{zhang2018perceptual} between the masked versions of fit and not-fit pairs, we quantified the perceptual loss between generated backgrounds when only the shape condition varied. ControlNet achieved an average LPIPS of 10.26, while our method improved on this with an average LPIPS of 9.13.

\begin{figure*}[ht]
    \centering
    \fivexonetable{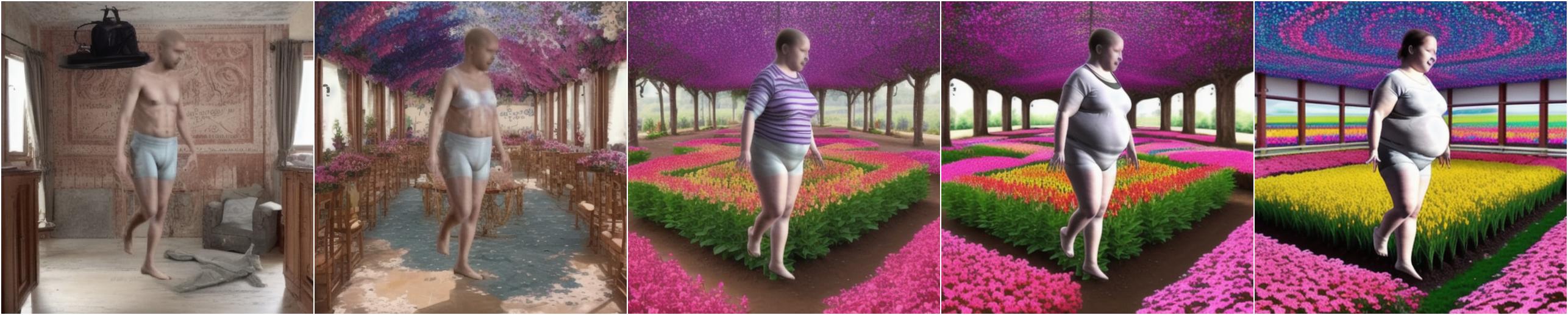}{3.4cm}
    \caption{Ablated configuration without domain adaptation and no guidance composition. The prompt condition is supplied to the Stable Diffusion U-Net. We show results for different scales $w$ in the ablated composition $\epsilon_{\text{Syn}}(\emptyset,\emptyset,\boldsymbol{c}_o) + w \boldsymbol{\Vec{g}_{c_p,c_s,c_o}}$. }
    \label{fig:guidance_no_composition}
\end{figure*}

\subsection{Animation}
AnimateDiff \cite{guoAnimateDiffAnimateYour2024} enables transition-consistent generation of frames, and can be controlled using a Control mechanism such as ControlNet. We use our approach to add 3d parametric SMPL control to the animations, and generate several clips with both slim and obese variants. See \cref{fig:animation_frames} for frames from our animation. We supply several animated clips in our supplemental video. The animation appears smooth, with no noteable degradation versus using the original slim-body type ControlNet. Faces appear less consistent, which is, however, a general limitation of the AnimateDiff method.

\subsection{Extended Guidance Ablation}
\begin{figure}
    \centering
    \twobytwotable{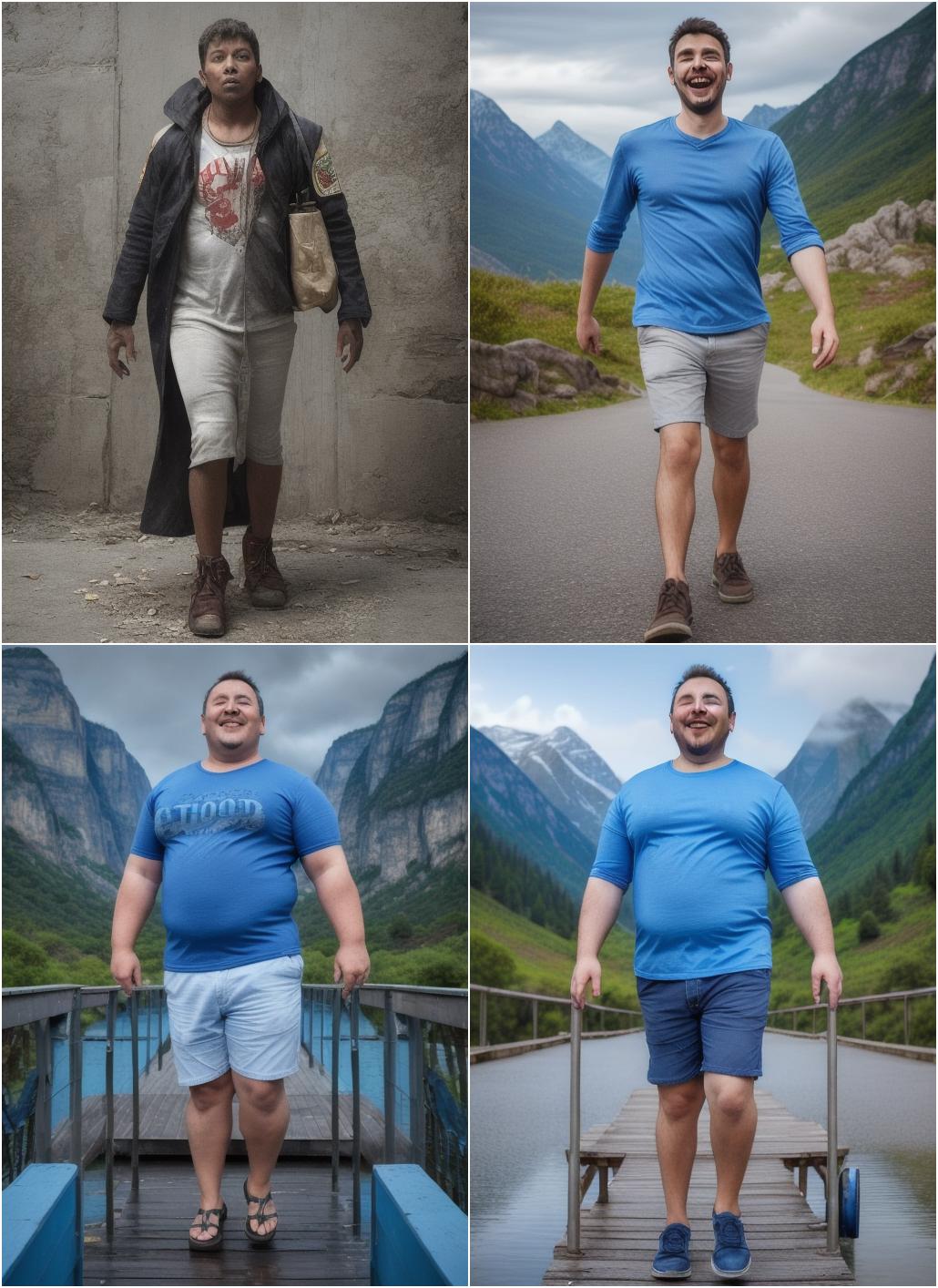}{3.4cm}{1cm}
    \caption{Ablation: including the prompt-condition into the SD U-Net of $C_{\text{SMPL}}$, i.e., $\epsilon_{\text{SD}}(\emptyset, \boldsymbol{c}_o)+w_1\boldsymbol{\Vec{g}_{c_p,c_o}} + w_2\boldsymbol{\Vec{g}_{c_p,c_s,c_o}}$}
    \label{fig:ablated_guidance_extra_prompt}
\end{figure}

Our proposed approach in \cref{subsec:smplguidance} always uses the empty prompt ($\emptyset$) for the attribute guidance SD U-Net's cross-attention layers during training and inference. That this is desirable can be seen in  \cref{fig:ablated_guidance_extra_prompt}, in which we use an ablated configuration which also uses the prompt in the SMPL-guidance SD UNet, resulting in the guidance vector $\boldsymbol{\Vec{g}_{c_p,c_s,c_o}}$. As the two guidance vectors can no longer be scaled independently, increasing the guidance scales leads to unstable background and exaggerated contrast.
In \cref{fig:guidance_no_composition} we also provide the prompt to the SD U-Net’s cross-attention layers during inference, but don't use guidance composition, i.e., only using $\epsilon_\text{Syn}$. We vary the combined prompt+smpl guidance scale ($w$) for this configuration. While the body shape is adhered to, the results exhibit a distinct synthetic appearance, which is exacerbated when increasing the shape adherence via $w$ (in contrast to increasing $w_2$ of our proposed approach). %

\subsection{Guidance Scale Analysis}
\begin{figure*}[ht]
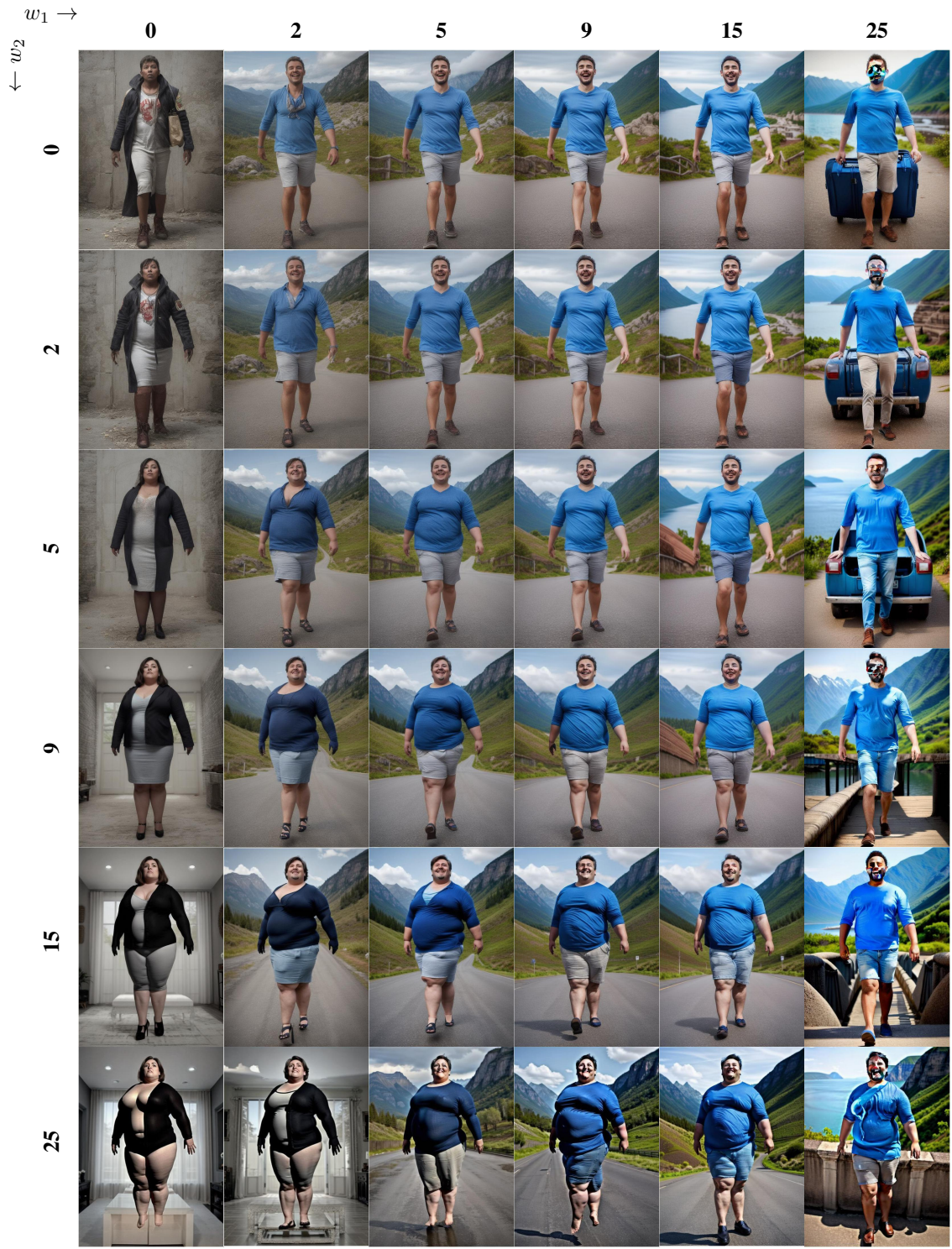

    \centering
    \sixxsixtable{originalguidance}{2.3cm}
    \caption{Varying guidance scales ($w_1,w_2$) in our proposed guidance composition: $\epsilon_{\text{SD}}(\emptyset,\boldsymbol{c}_o) + w_1 \boldsymbol{\Vec{g}_{c_p,c_o}} + w_2 \boldsymbol{\Vec{g}_{c_s,c_o}}$.}
    \label{fig:guidance_original}
\end{figure*}

Our paper generally uses guidance scales ($w_1, w_2$) of 7.5 for all images, unless otherwise specified. These scales can be varied independently to adjust the strength of each component. In \cref{fig:guidance_original}, we show an extended version of Fig 7a from the main paper, by varying the guidance scale between 0 and 25.
Moderate guidance scales below 25 yield high-fidelity images. Increasing the domain guidance scale ($w_1$) enhances prompt adherence but introduces unstable backgrounds and artifacts at $w_1 = 25$. Similarly, increasing $w_2$ exaggerates the SMPL shape, causing body artifacts at $w_2 = 25$. Higher $w_2$ values make images resemble the synthetic SURREAL dataset, while higher $w_1$ values decrease shape adherence. The diagonal ($w_1 = w_2$) provides the best combination of prompt adherence, shape adherence, and visual appearance.

\begin{figure*}[ht]
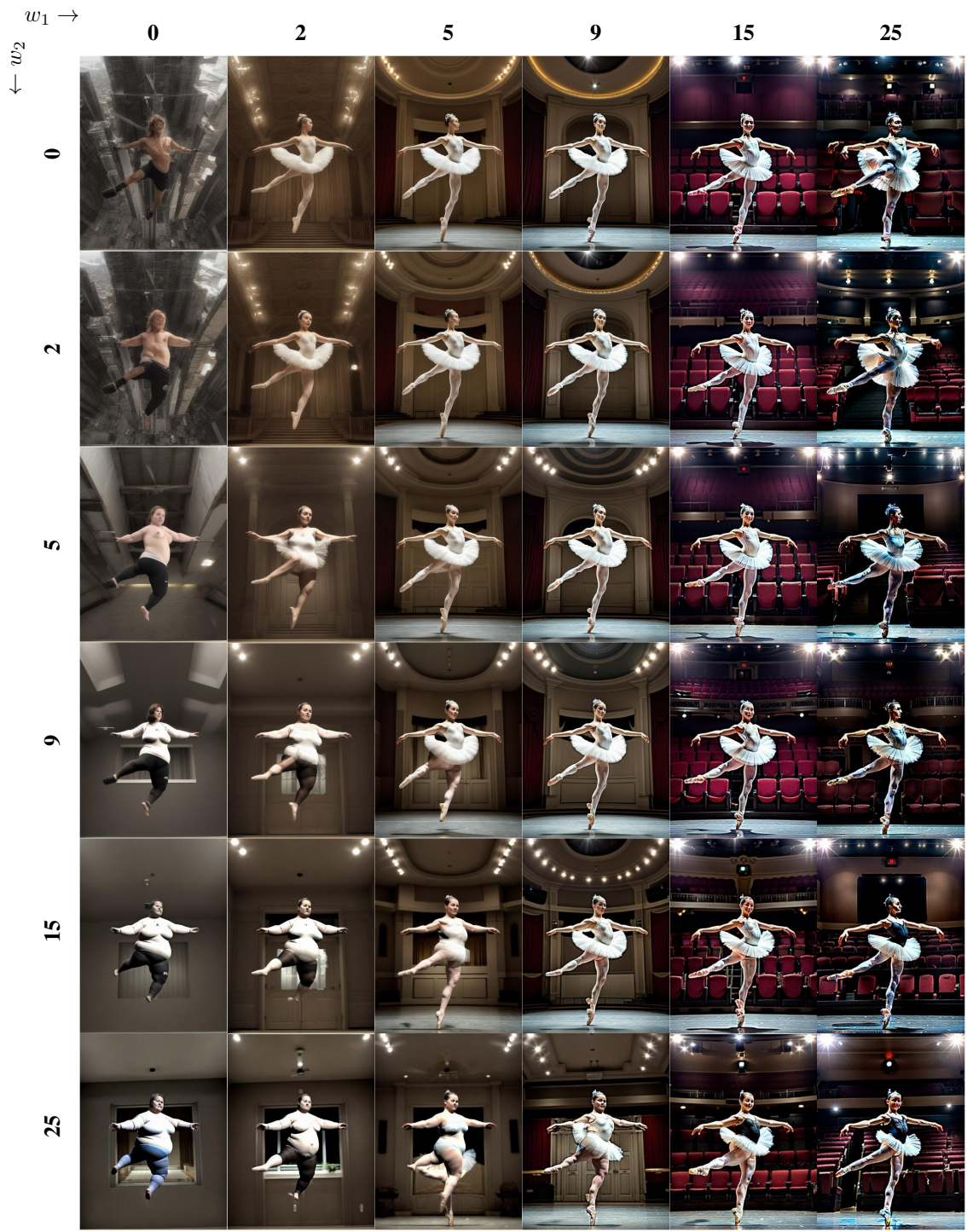

    \centering
    \sixxsixtable{dancer}{2.3cm}
   \caption{Limitation in Guidance Composition: The model struggles to reconcile conflicting information, such as obese SMPL-shape and slim ballet dancer produced by the pose-conditioned ControlNet.}
   \label{fig:compositionlimitation}
\end{figure*}

\cref{fig:compositionlimitation} uses the same guidance scale setup with the prompt ``a ballet dancer'' and a dancing pose from AIST \cite{AIST}. In this context, Stable Diffusion is biased towards generating slim body types, even with ``obese'' added to the prompt. As discussed in our limitations, our method similarly struggles to reconcile such conflicting information. Using an obese body shape for our SMPL-conditioned model only imparts the shape if $w_2$ is much larger than $w_1$, but this also results in a synthetic appearance. Intermediate scales also display artifacts in the body composition.  It must be noted that the severity of this limitation generally depends on factors such as the prompt, seed, and pose. It can work well in some contexts, e.g., the football player in Fig 6 of the main paper, likely due to specific training biases of Stable Diffusion.

\subsection{Examples from Quantitative Evaluation}
In Sections 5.2 and 5.3, we evaluated the fidelity and SMPL accuracy metrics of various model configurations and baseline models. We generated datasets using pose and shape inputs for each method, using a fixed seed. In \cref{fig:coco_eval_comparison,fig:aist_eval_comparison}, we provide visual examples from our experiments.

\cref{fig:coco_eval_comparison} presents outputs generated from the pose and caption inputs from the MSCOCO \cite{MSCOCO} dataset. 
As the models are only conditioned on high-level semantic information (text prompt, OpenPose semantic map and SMPL parameters), the differences in terms of visual content to the reference images (\cref{fig:coco_eval_comparison}a) are expected. We thus calculated visual fidelity in terms of Kernel Inception Distance (KID) to measure the dataset’s overall distance to the reference real-world dataset\cite{MSCOCO}.
As shown in the metrics, fine-tuning a ControlNet with additional cross-attention blocks and without domain adaptation (\cref{fig:coco_eval_comparison}c) introduces a synthetic appearance. In contrast, our guidance composition avoids domain shift (\cref{fig:coco_eval_comparison}d, e and g).
Additionally, it is evident that the domain guidance network (\cref{fig:coco_eval_comparison}f and h) controls the overall layout of the image, while our method introduces only slight changes in the generated person. Fine-tuning only the attention blocks (ft-attn) keeps the background more stable compared to fine-tuning all blocks.

\cref{fig:aist_eval_comparison} shows outputs generated using the AIST \cite{AIST}-SURREAL \cite{SURREAL} SMPL models (see Section 5.3). It is visible that our models adhere better to the shape. In some cases, our ControlNet-guided (ft+CN) models also correct wrong body orientations introduced by ControlNet, while T2I-Adapter does not seem to suffer from this issue.

\newcommand{\subwidth}{0.12\textwidth}

\begin{figure*}[htbp]
    \setlength{\tabcolsep}{0em}
    \renewcommand{\arraystretch}{0}
    \centering
        \begin{subfigure}[b]{\subwidth}
            \centering
            \includegraphics[width=\textwidth]{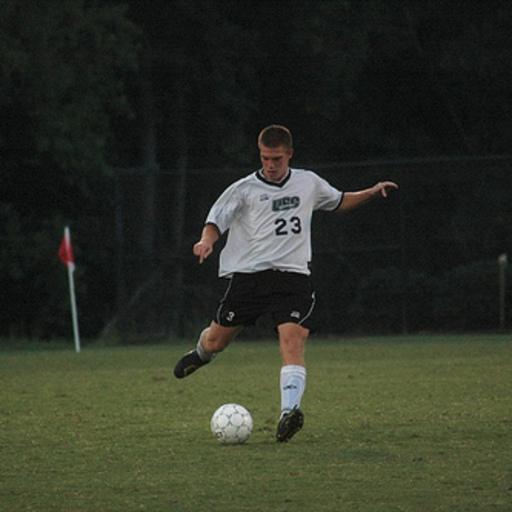}
        \end{subfigure}
        \begin{subfigure}[b]{\subwidth}
            \centering
            \includegraphics[width=\textwidth]{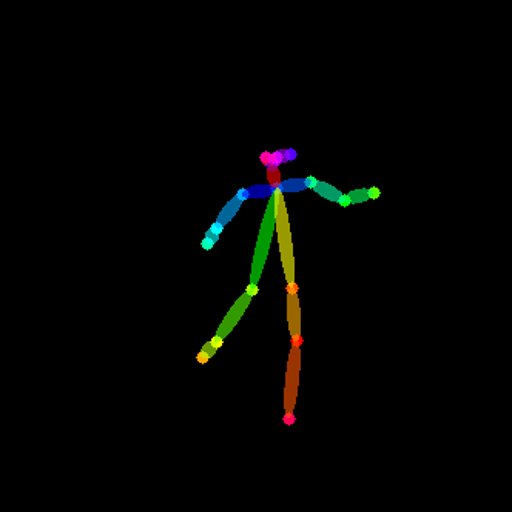}
        \end{subfigure}
        \begin{subfigure}[b]{\subwidth}
            \centering
            \includegraphics[width=\textwidth]{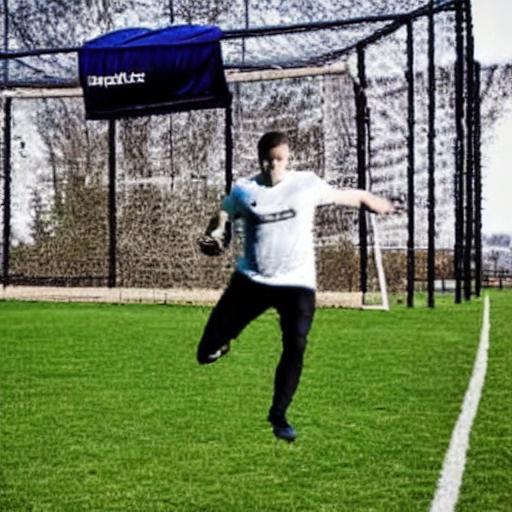}
        \end{subfigure}
        \begin{subfigure}[b]{\subwidth}
            \centering
            \includegraphics[width=\textwidth]{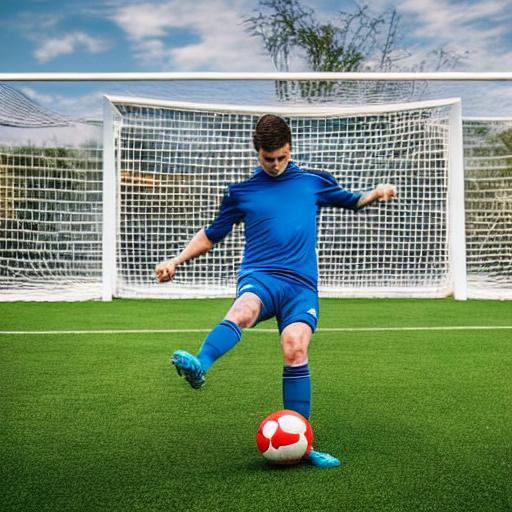}
        \end{subfigure}
        \begin{subfigure}[b]{\subwidth}
            \centering
            \includegraphics[width=\textwidth]{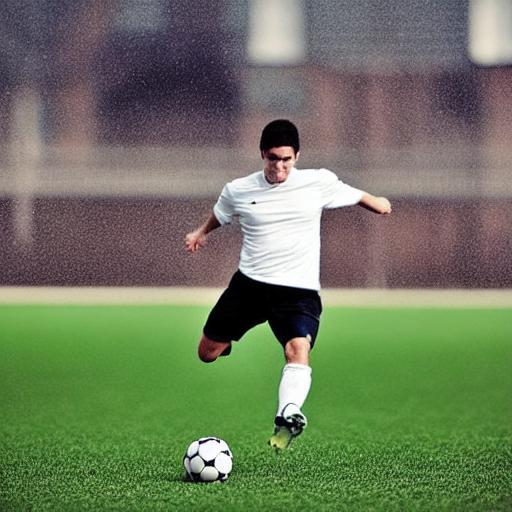}
        \end{subfigure}
        \begin{subfigure}[b]{\subwidth}
            \centering
            \includegraphics[width=\textwidth]{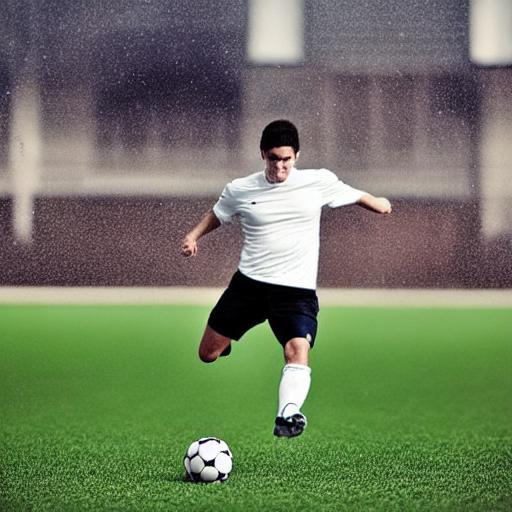}
        \end{subfigure} 
        \begin{subfigure}[b]{\subwidth}
            \centering
            \includegraphics[width=\textwidth]{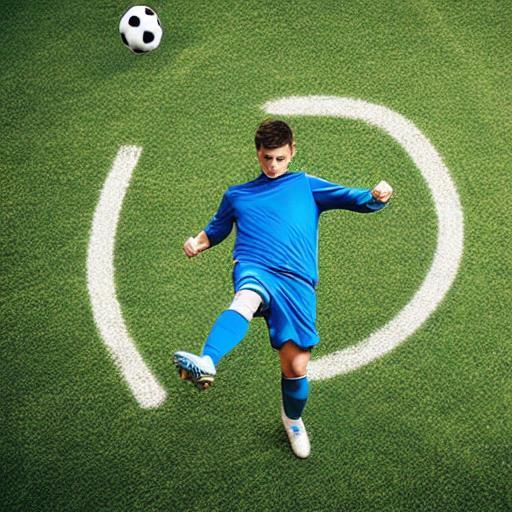}
        \end{subfigure}
        \begin{subfigure}[b]{\subwidth}
            \centering
            \includegraphics[width=\textwidth]{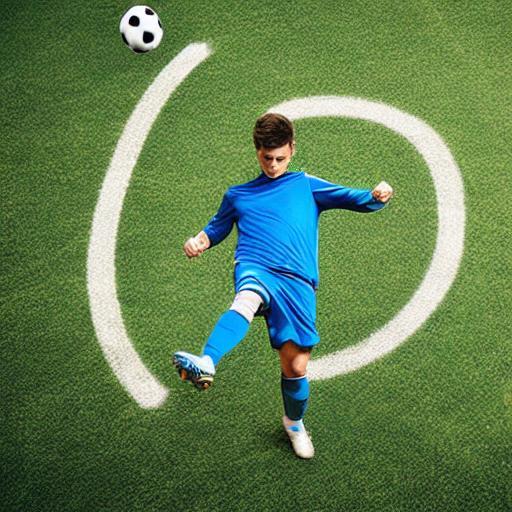}
        \end{subfigure}\\

        \begin{subfigure}[b]{\subwidth}
            \centering
            \includegraphics[width=\textwidth]{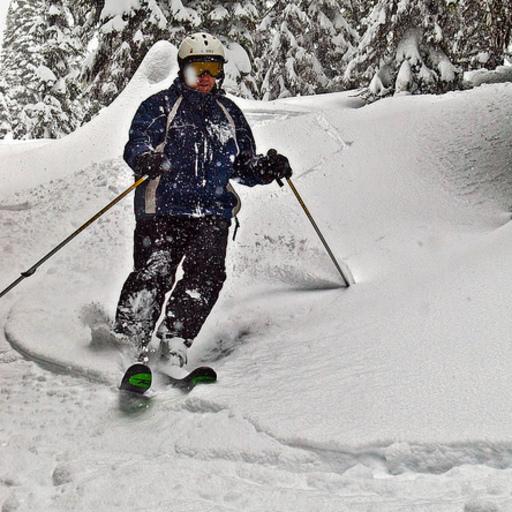}
        \end{subfigure}
        \begin{subfigure}[b]{\subwidth}
            \centering
            \includegraphics[width=\textwidth]{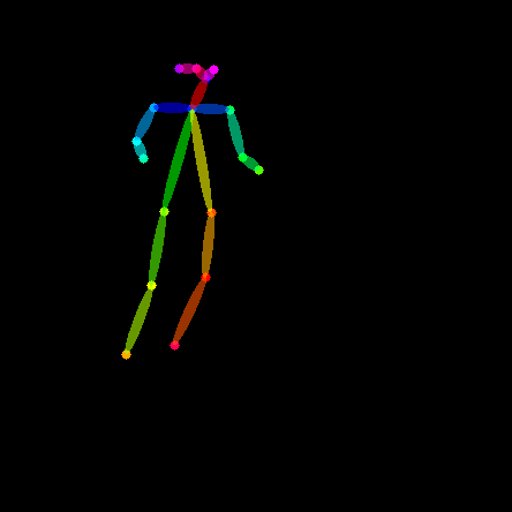}
        \end{subfigure}
        \begin{subfigure}[b]{\subwidth}
            \centering
            \includegraphics[width=\textwidth]{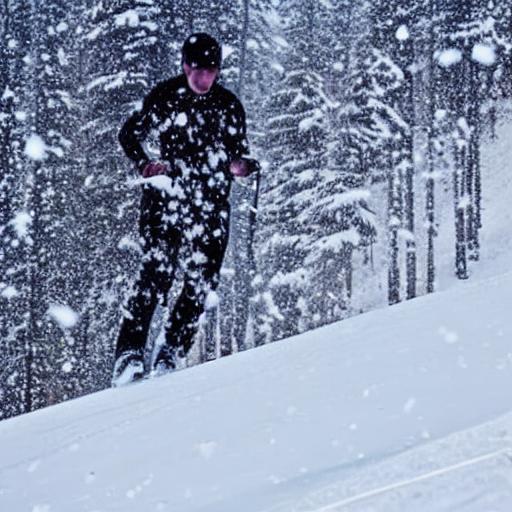}
        \end{subfigure}
        \begin{subfigure}[b]{\subwidth}
            \centering
            \includegraphics[width=\textwidth]{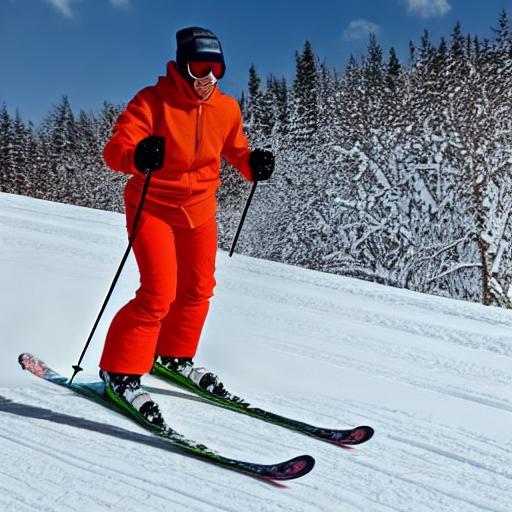}
        \end{subfigure}
        \begin{subfigure}[b]{\subwidth}
            \centering
            \includegraphics[width=\textwidth]{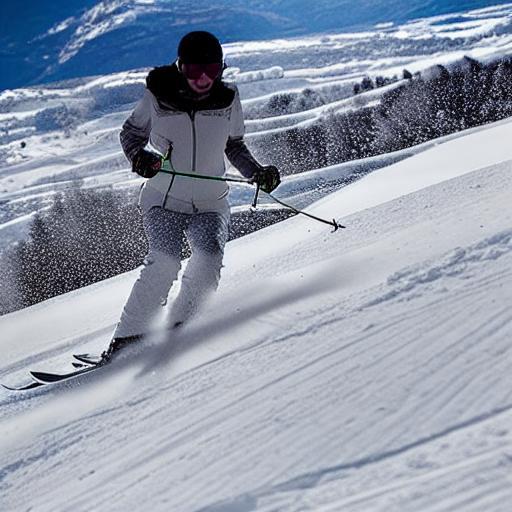}
        \end{subfigure}
        \begin{subfigure}[b]{\subwidth}
            \centering
            \includegraphics[width=\textwidth]{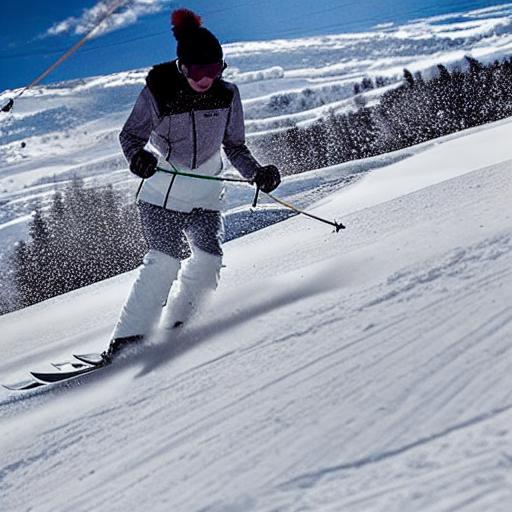}
        \end{subfigure}
        \begin{subfigure}[b]{\subwidth}
            \centering
            \includegraphics[width=\textwidth]{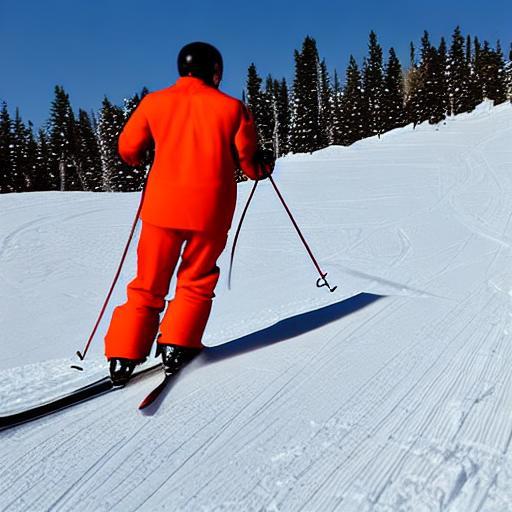}
        \end{subfigure}
        \begin{subfigure}[b]{\subwidth}
            \centering
            \includegraphics[width=\textwidth]{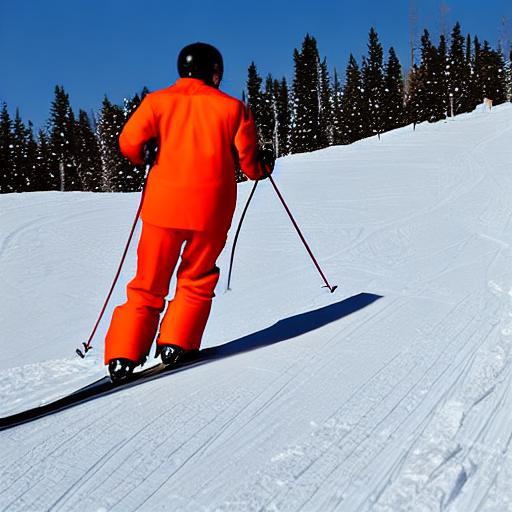}
        \end{subfigure} \\

        \begin{subfigure}[b]{\subwidth}
            \centering
            \includegraphics[width=\textwidth]{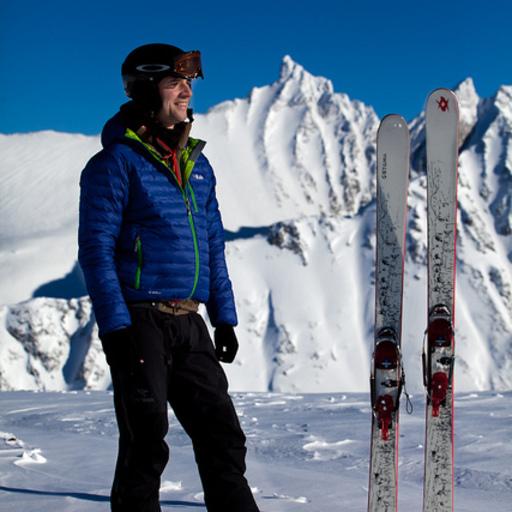}
        \end{subfigure}
        \begin{subfigure}[b]{\subwidth}
            \centering
            \includegraphics[width=\textwidth]{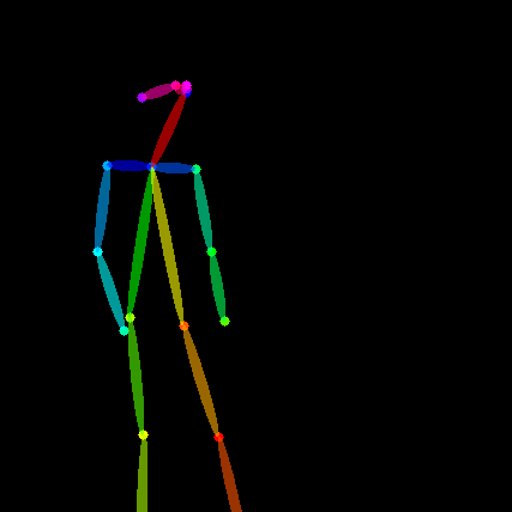}
        \end{subfigure}
        \begin{subfigure}[b]{\subwidth}
            \centering
            \includegraphics[width=\textwidth]{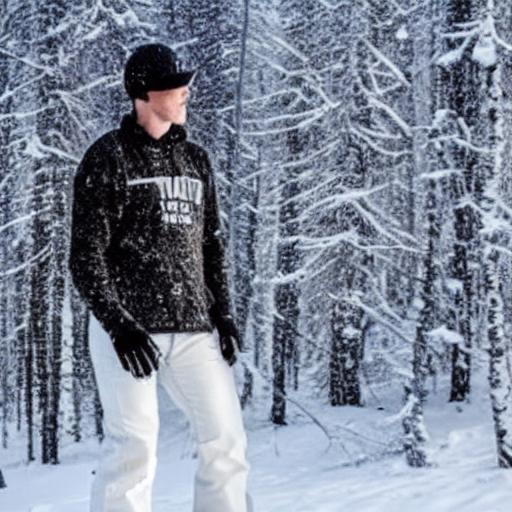}
        \end{subfigure}
        \begin{subfigure}[b]{\subwidth}
            \centering
            \includegraphics[width=\textwidth]{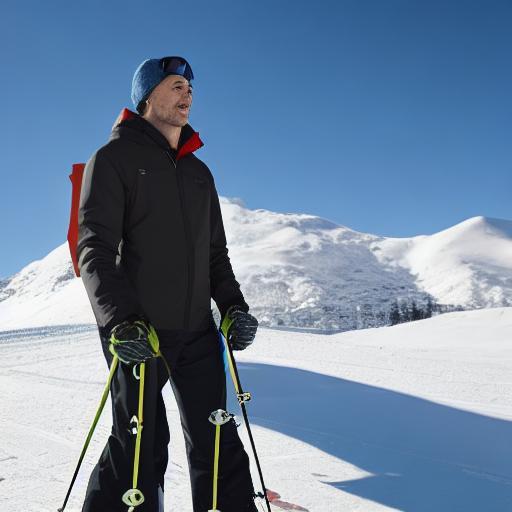}
        \end{subfigure}
        \begin{subfigure}[b]{\subwidth}
            \centering
            \includegraphics[width=\textwidth]{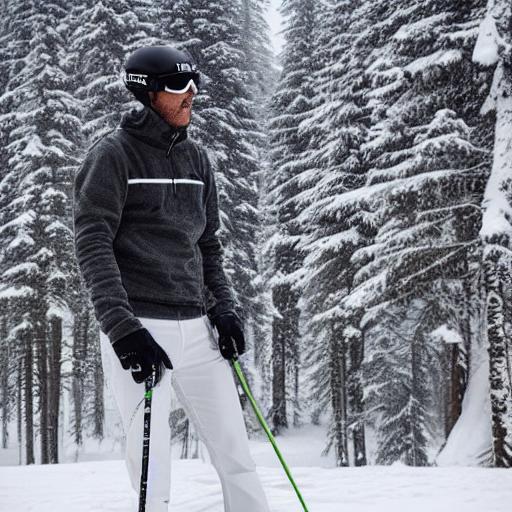}
        \end{subfigure}
        \begin{subfigure}[b]{\subwidth}
            \centering
            \includegraphics[width=\textwidth]{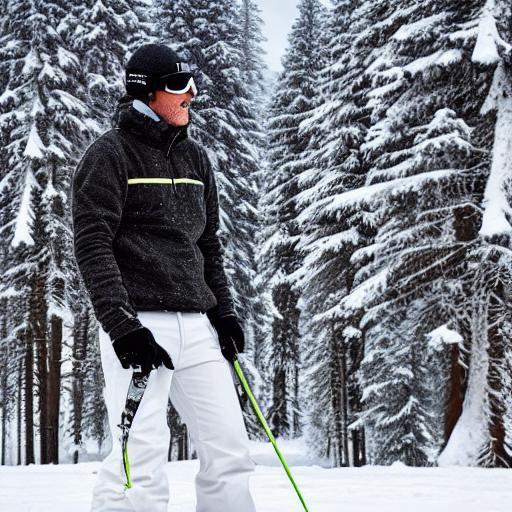}
        \end{subfigure}
        \begin{subfigure}[b]{\subwidth}
            \centering
            \includegraphics[width=\textwidth]{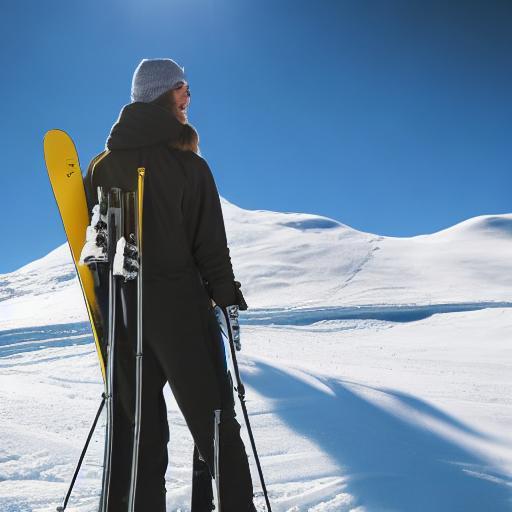}
        \end{subfigure}
        \begin{subfigure}[b]{\subwidth}
            \centering
            \includegraphics[width=\textwidth]{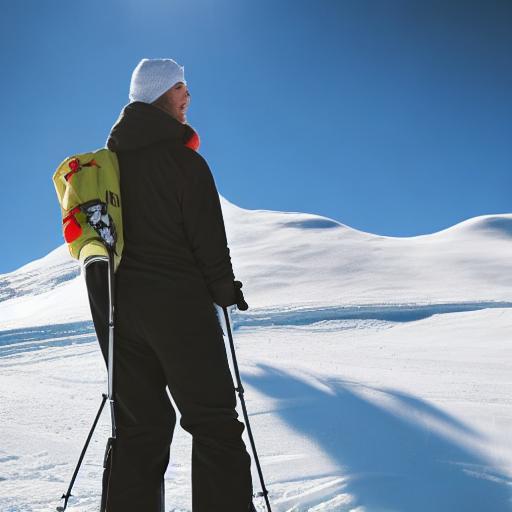}
        \end{subfigure} \\

        \begin{subfigure}[b]{\subwidth}
            \centering
            \includegraphics[width=\textwidth]{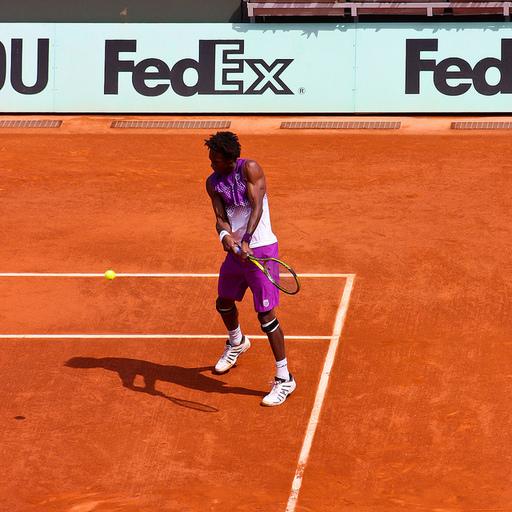}
        \end{subfigure}
        \begin{subfigure}[b]{\subwidth}
            \centering
            \includegraphics[width=\textwidth]{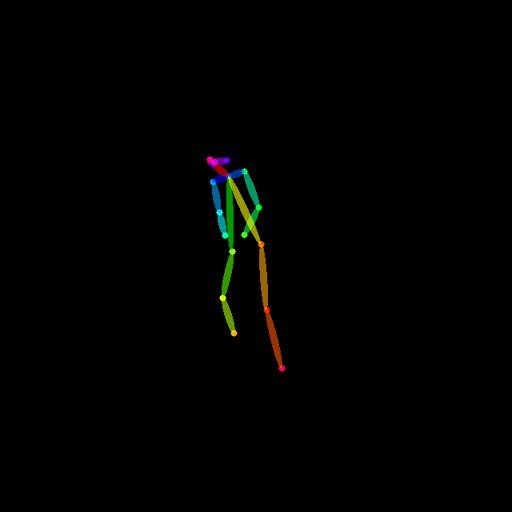}
        \end{subfigure}
        \begin{subfigure}[b]{\subwidth}
            \centering
            \includegraphics[width=\textwidth]{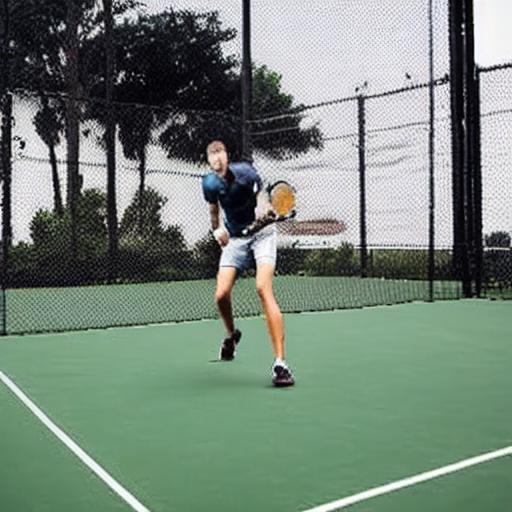}
        \end{subfigure}
        \begin{subfigure}[b]{\subwidth}
            \centering
            \includegraphics[width=\textwidth]{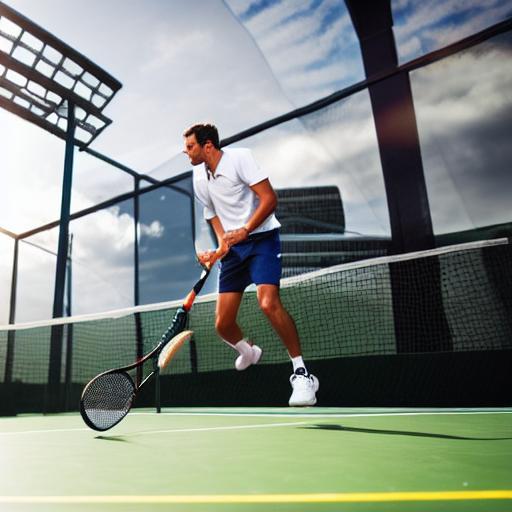}
        \end{subfigure}
        \begin{subfigure}[b]{\subwidth}
            \centering
            \includegraphics[width=\textwidth]{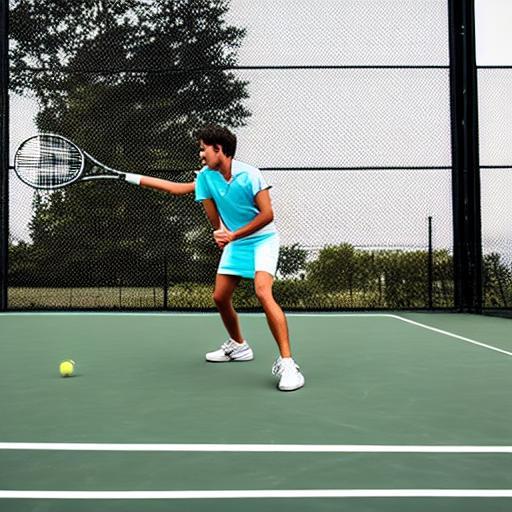}
        \end{subfigure}
        \begin{subfigure}[b]{\subwidth}
            \centering
            \includegraphics[width=\textwidth]{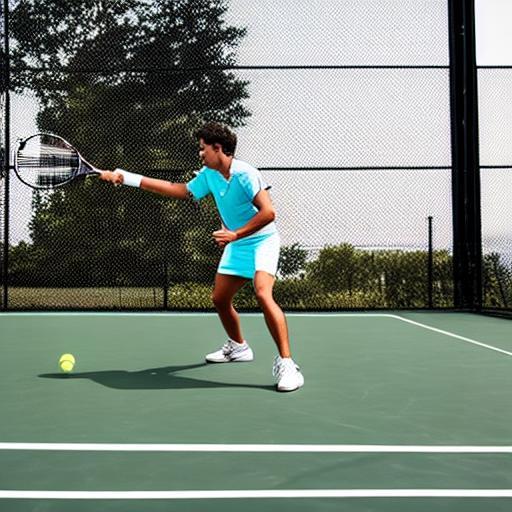}
        \end{subfigure}
        \begin{subfigure}[b]{\subwidth}
            \centering
            \includegraphics[width=\textwidth]{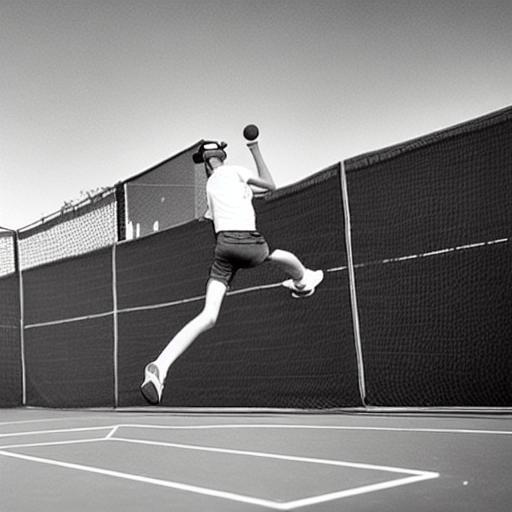}
        \end{subfigure}
        \begin{subfigure}[b]{\subwidth}
            \centering
            \includegraphics[width=\textwidth]{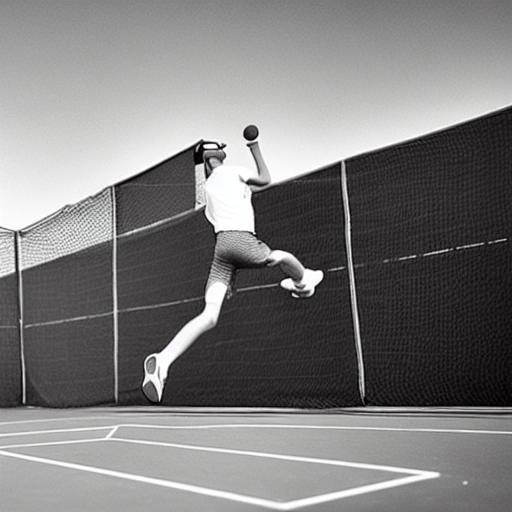}
        \end{subfigure} \\
        
        \begin{subfigure}[b]{\subwidth}
            \centering
            \includegraphics[width=\textwidth]{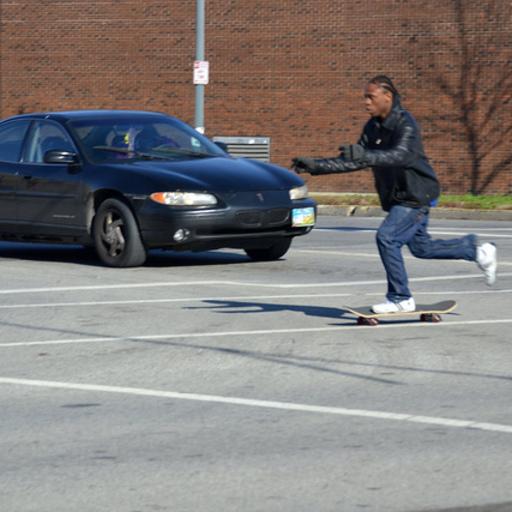}
        \end{subfigure}
        \begin{subfigure}[b]{\subwidth}
            \centering
            \includegraphics[width=\textwidth]{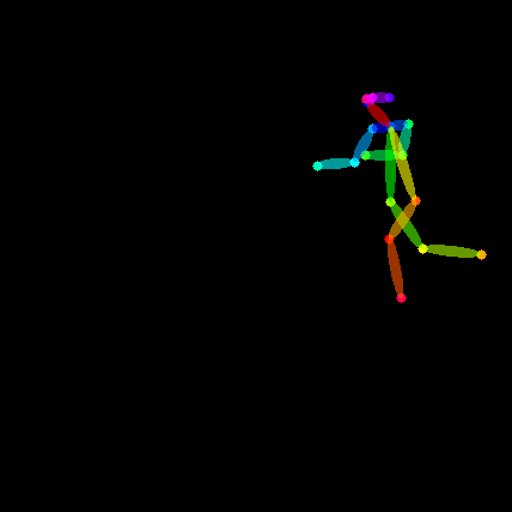}
        \end{subfigure}
        \begin{subfigure}[b]{\subwidth}
            \centering
            \includegraphics[width=\textwidth]{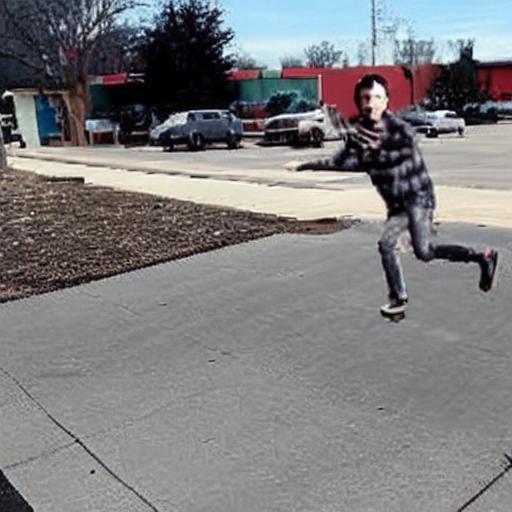}
        \end{subfigure}
        \begin{subfigure}[b]{\subwidth}
            \centering
            \includegraphics[width=\textwidth]{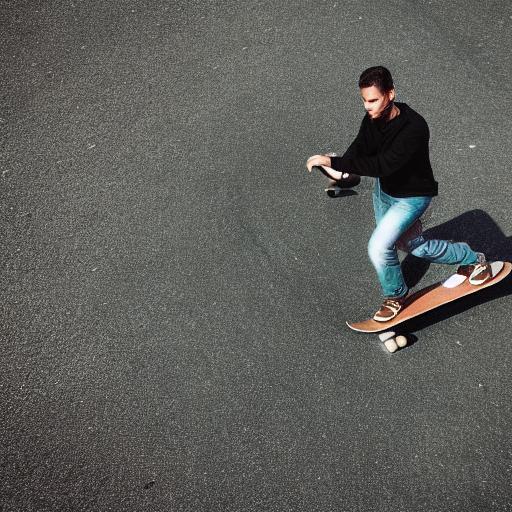}
        \end{subfigure}
        \begin{subfigure}[b]{\subwidth}
            \centering
            \includegraphics[width=\textwidth]{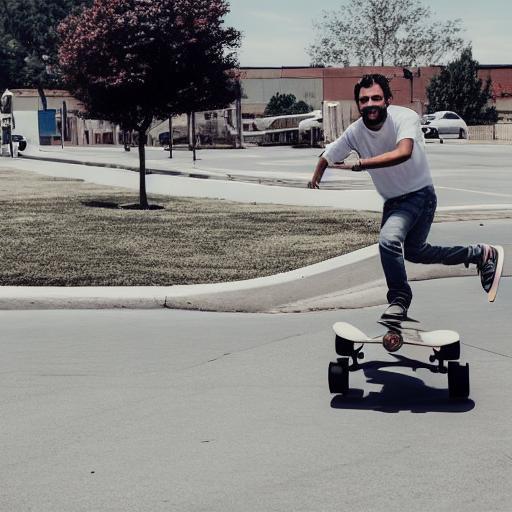}
        \end{subfigure}
        \begin{subfigure}[b]{\subwidth}
            \centering
            \includegraphics[width=\textwidth]{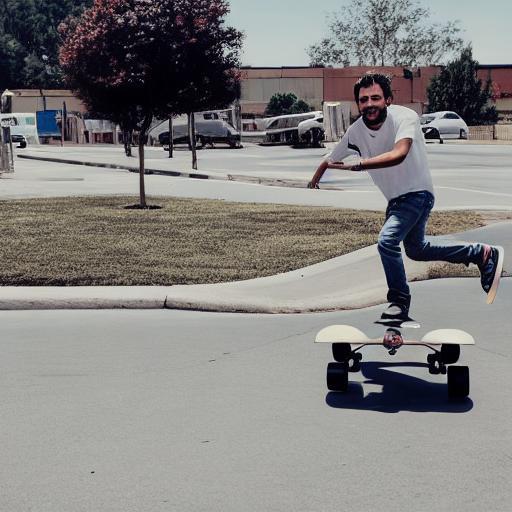}
        \end{subfigure}
        \begin{subfigure}[b]{\subwidth}
            \centering
            \includegraphics[width=\textwidth]{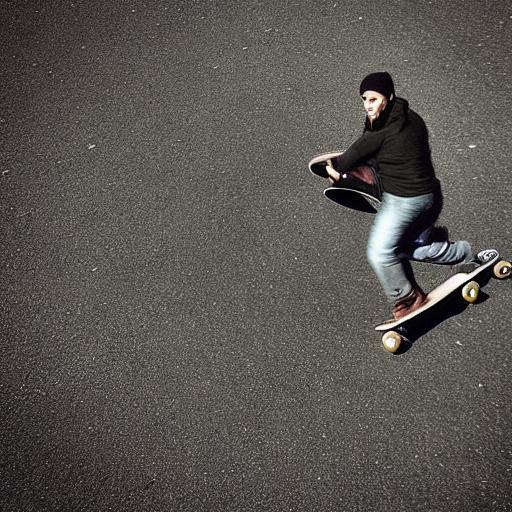}
        \end{subfigure}
        \begin{subfigure}[b]{\subwidth}
            \centering
            \includegraphics[width=\textwidth]{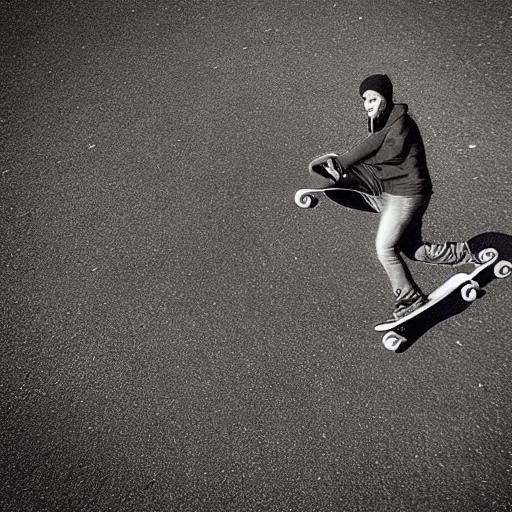}
        \end{subfigure} \\

        \begin{subfigure}[b]{\subwidth}
            \centering
            \includegraphics[width=\textwidth]{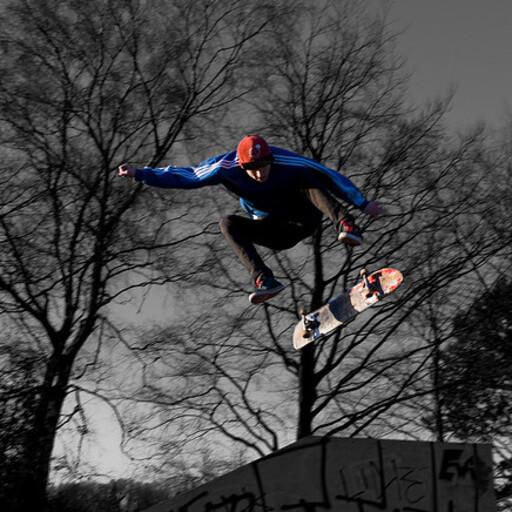}
        \end{subfigure}
        \begin{subfigure}[b]{\subwidth}
            \centering
            \includegraphics[width=\textwidth]{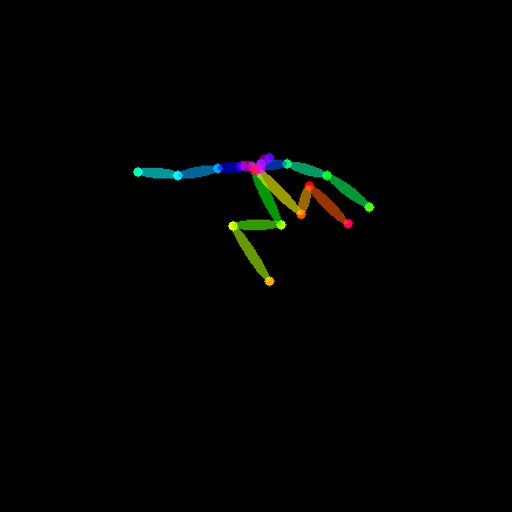}
        \end{subfigure}
        \begin{subfigure}[b]{\subwidth}
            \centering
            \includegraphics[width=\textwidth]{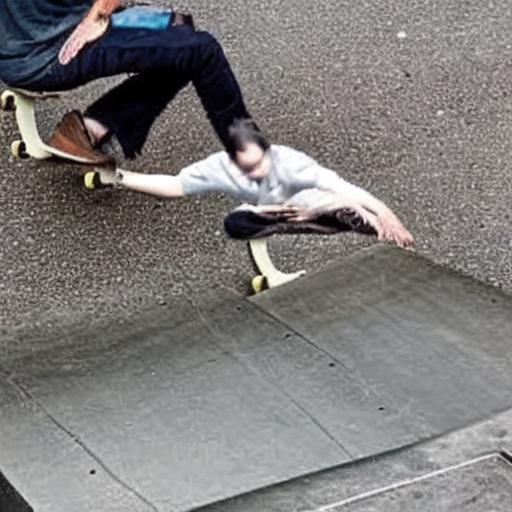}
        \end{subfigure}
        \begin{subfigure}[b]{\subwidth}
            \centering
            \includegraphics[width=\textwidth]{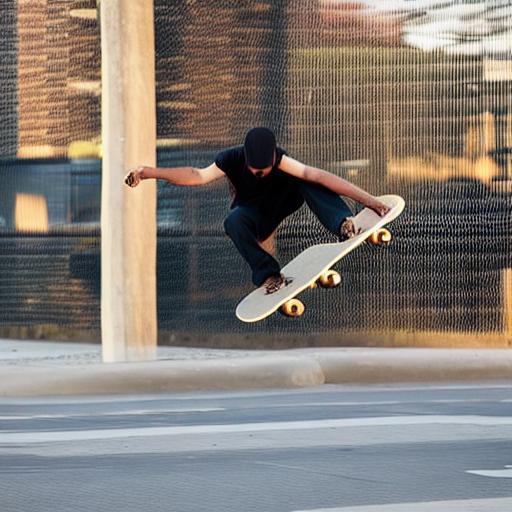}
        \end{subfigure}
        \begin{subfigure}[b]{\subwidth}
            \centering
            \includegraphics[width=\textwidth]{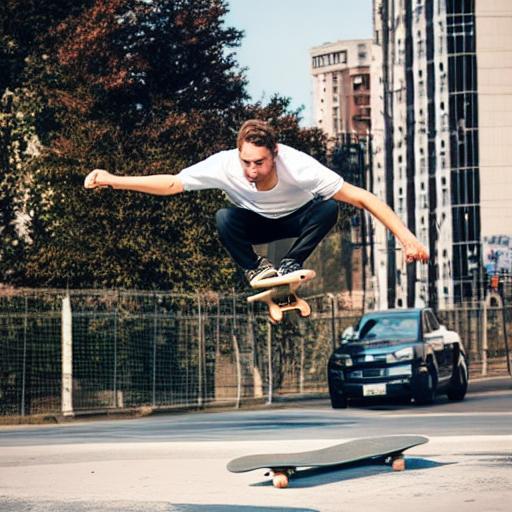}
        \end{subfigure}
        \begin{subfigure}[b]{\subwidth}
            \centering
            \includegraphics[width=\textwidth]{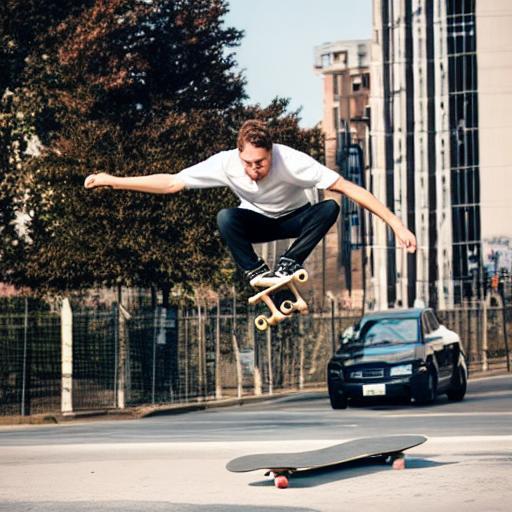}
        \end{subfigure}
        \begin{subfigure}[b]{\subwidth}
            \centering
            \includegraphics[width=\textwidth]{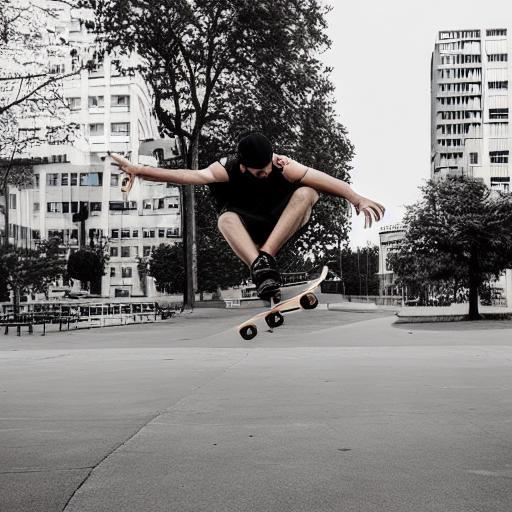}
        \end{subfigure}
        \begin{subfigure}[b]{\subwidth}
            \centering
            \includegraphics[width=\textwidth]{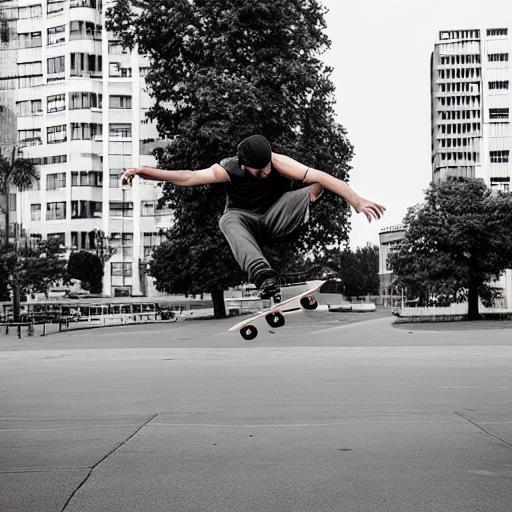}
        \end{subfigure} \\

        \begin{subfigure}[b]{\subwidth}
            \centering
            \includegraphics[width=\textwidth]{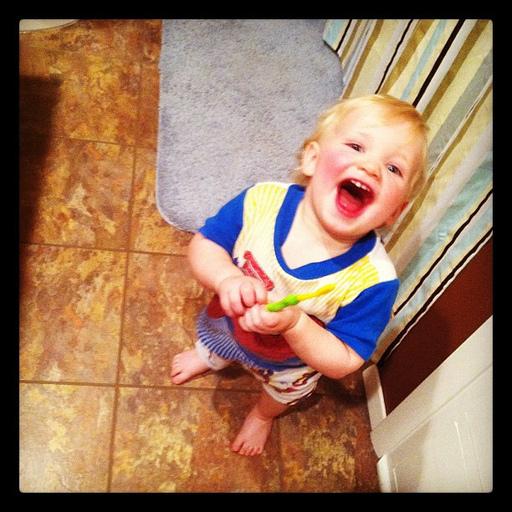}
            \caption{MSCOCO \cite{MSCOCO}}
        \end{subfigure}
        \begin{subfigure}[b]{\subwidth}
            \centering
            \includegraphics[width=\textwidth]{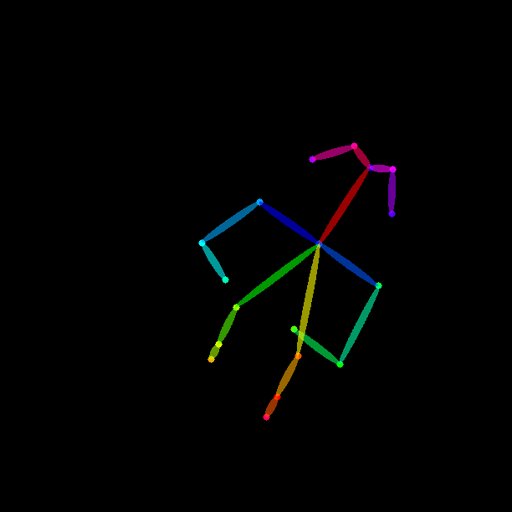}
            \caption{Pose}
        \end{subfigure}
        \begin{subfigure}[b]{\subwidth}
            \centering
            \includegraphics[width=\textwidth]{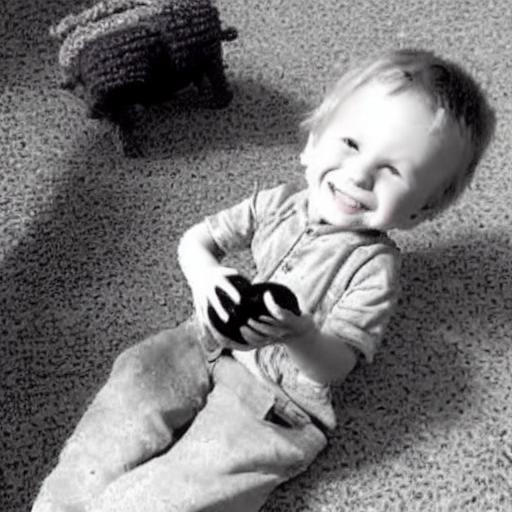}
            \caption{ft-extra-attn}
        \end{subfigure}
        \begin{subfigure}[b]{\subwidth}
            \centering
            \includegraphics[width=\textwidth]{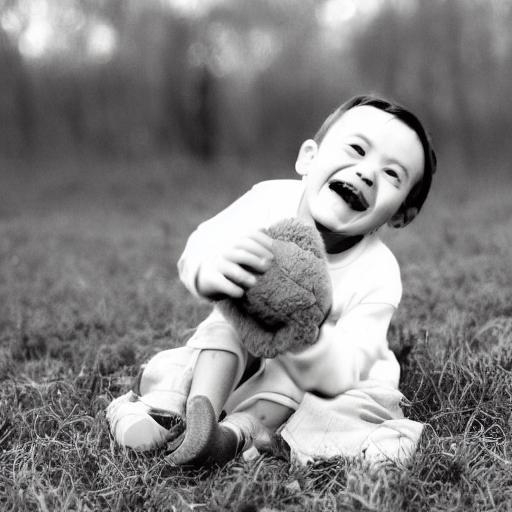}
            \caption{ft-all+CN}
        \end{subfigure}
        \begin{subfigure}[b]{\subwidth}
            \centering
            \includegraphics[width=\textwidth]{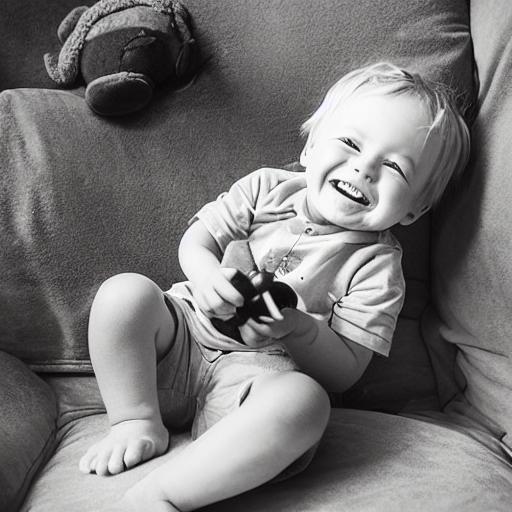}
            \caption{ft-attn+CN}
        \end{subfigure}
        \begin{subfigure}[b]{\subwidth}
            \centering
            \includegraphics[width=\textwidth]{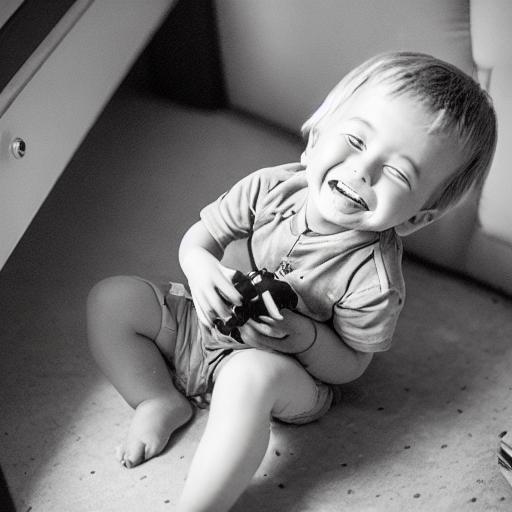}
            \caption{ControlNet}
        \end{subfigure}
        \begin{subfigure}[b]{\subwidth}
            \centering
            \includegraphics[width=\textwidth]{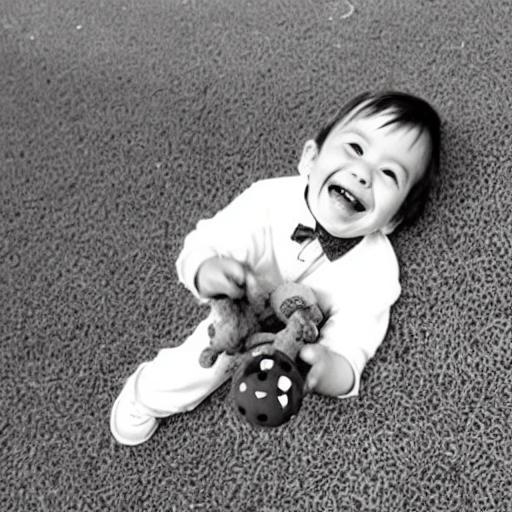}
            \caption{ft-attn+T2I}
        \end{subfigure}
        \begin{subfigure}[b]{\subwidth}
            \centering
            \includegraphics[width=\textwidth]{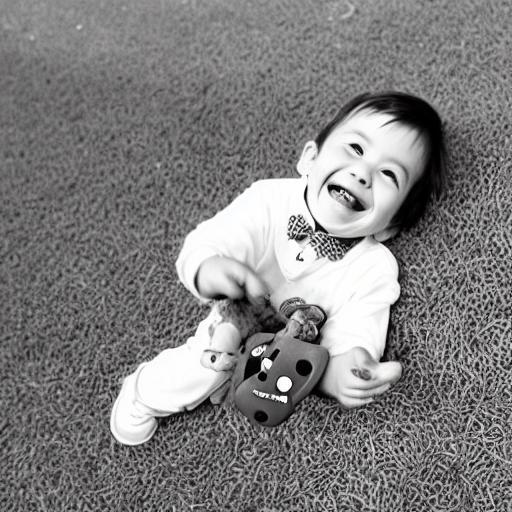}
            \caption{T2I-Adapter}
        \end{subfigure}
    \caption{Examples from the visual fidelity evaluation (Table 1, main paper). Our SMPL-finetuned (``ft-'') models receive the text, pose and SMPL inputs, while the standard ControlNet and T2I-Adapter only receive the text and pose input. The text prompt and OpenPose Keypoints are provided by the MSCOCO\cite{MSCOCO} dataset, while SMPL annotations are predicted using HierProbHuman\cite{sengupta2021hierarchical}.}
    \label{fig:coco_eval_comparison}
\end{figure*}

\begin{figure*}[htbp]
    \setlength{\tabcolsep}{0em}
    \renewcommand{\arraystretch}{0}
    \centering

        \begin{subfigure}[b]{\subwidth}
            \centering
            \includegraphics[width=\textwidth]{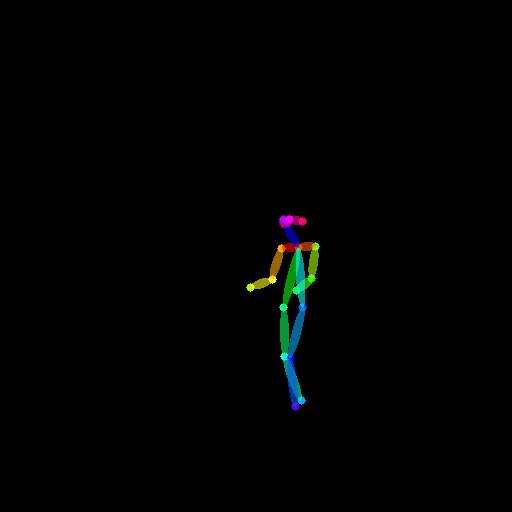}
        \end{subfigure}
        \begin{subfigure}[b]{\subwidth}
            \centering
            \includegraphics[width=\textwidth, height=\textwidth]{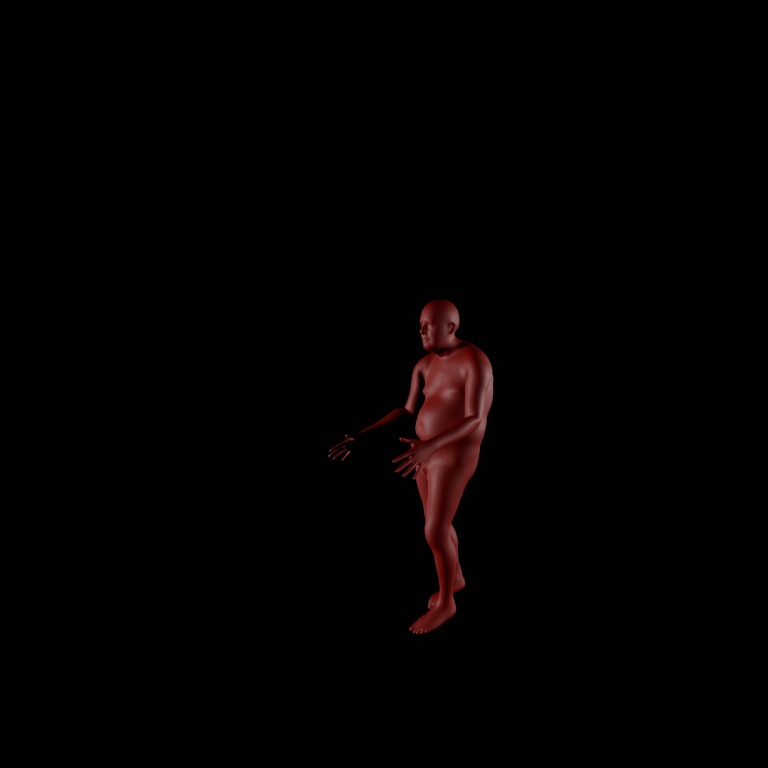}
        \end{subfigure}
        \begin{subfigure}[b]{\subwidth}
            \centering
            \includegraphics[width=\textwidth]{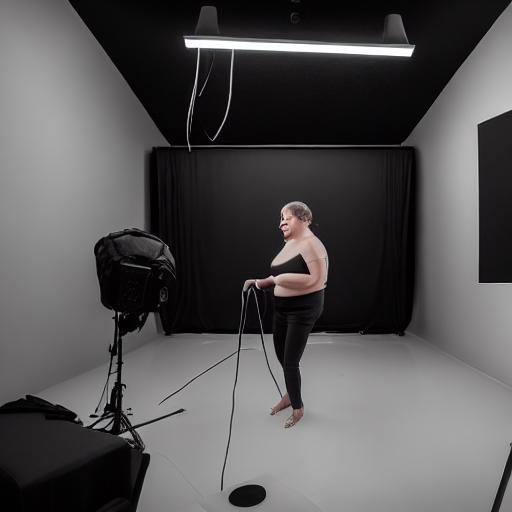}
        \end{subfigure}
        \begin{subfigure}[b]{\subwidth}
            \centering
            \includegraphics[width=\textwidth]{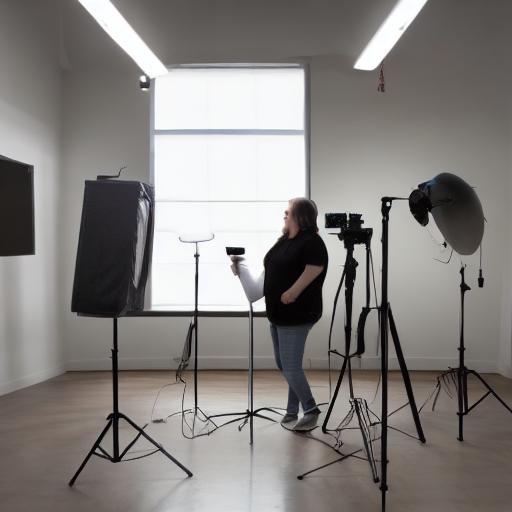}
        \end{subfigure}
        \begin{subfigure}[b]{\subwidth}
            \centering
            \includegraphics[width=\textwidth]{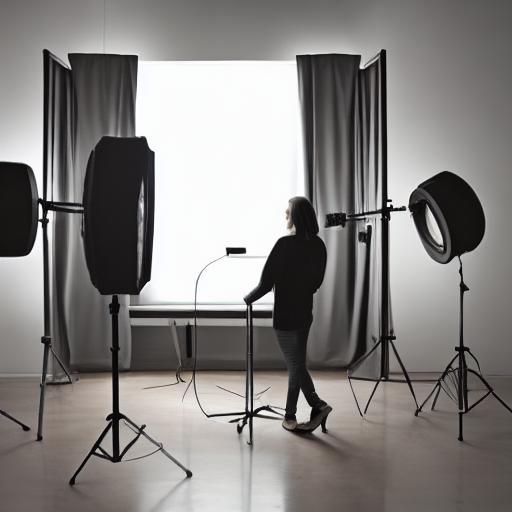}
        \end{subfigure}
        \begin{subfigure}[b]{\subwidth}
            \centering
            \includegraphics[width=\textwidth]{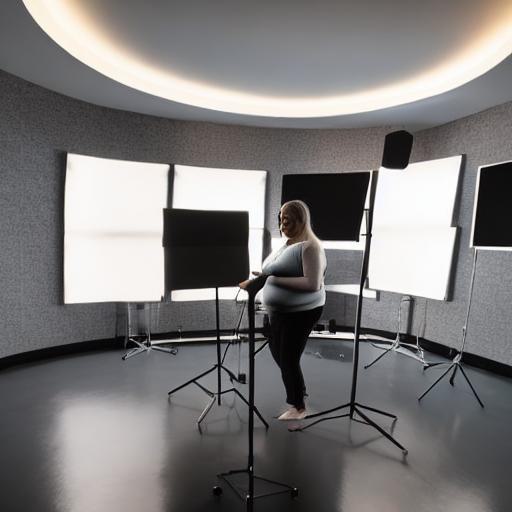}
        \end{subfigure}
        \begin{subfigure}[b]{\subwidth}
            \centering
            \includegraphics[width=\textwidth]{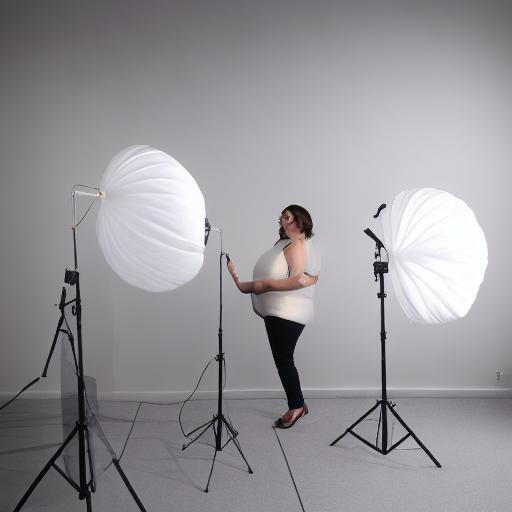}
        \end{subfigure} 
        \begin{subfigure}[b]{\subwidth}
            \centering
            \includegraphics[width=\textwidth]{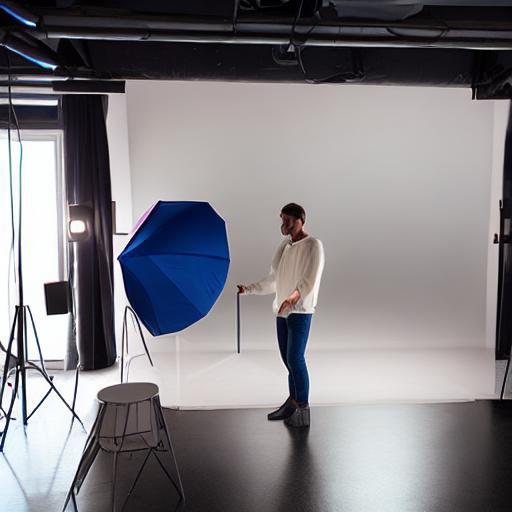}
        \end{subfigure}\\

        \begin{subfigure}[b]{\subwidth}
            \centering
            \includegraphics[width=\textwidth]{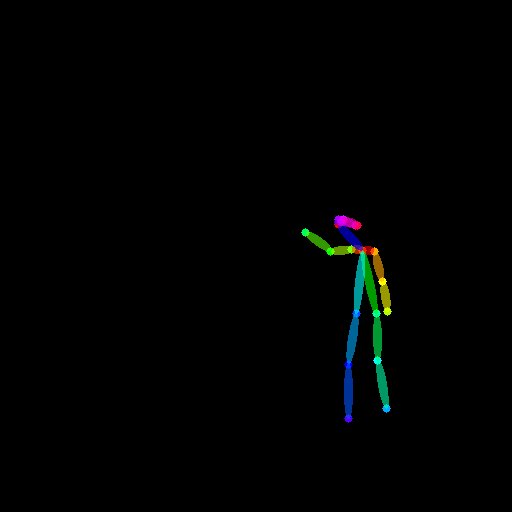}
        \end{subfigure}
        \begin{subfigure}[b]{\subwidth}
            \centering
            \includegraphics[trim={0cm 2.5cm 0cm 3.5cm},width=\textwidth, clip, height=\textwidth]{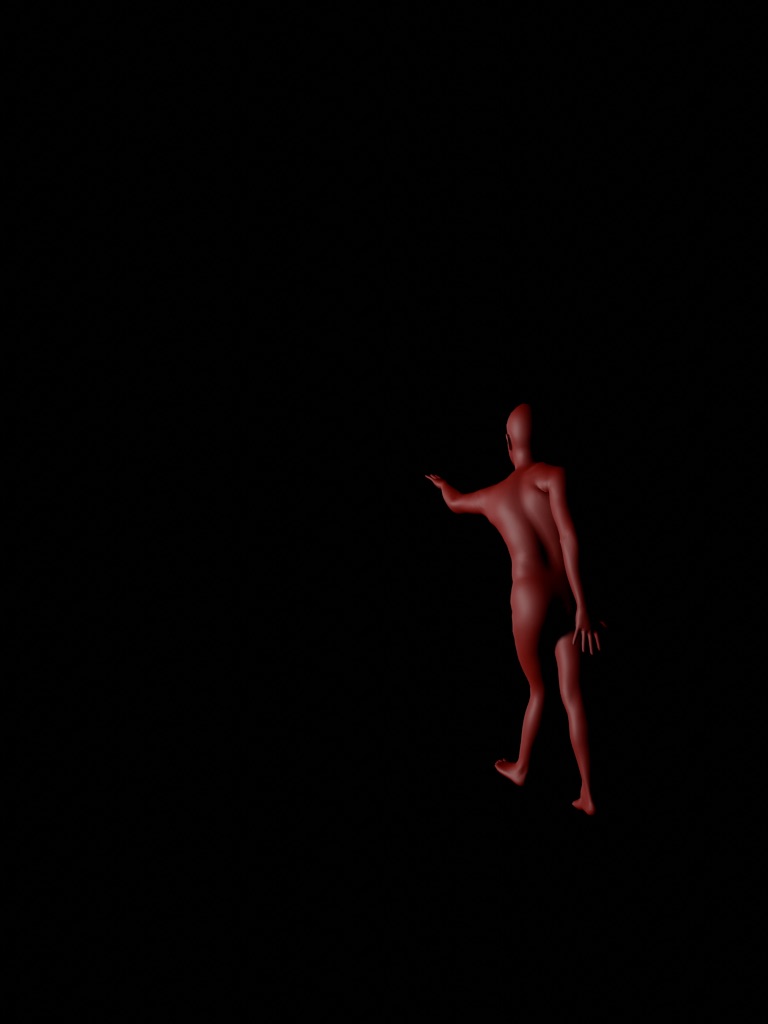}
        \end{subfigure}
        \begin{subfigure}[b]{\subwidth}
            \centering
            \includegraphics[width=\textwidth]{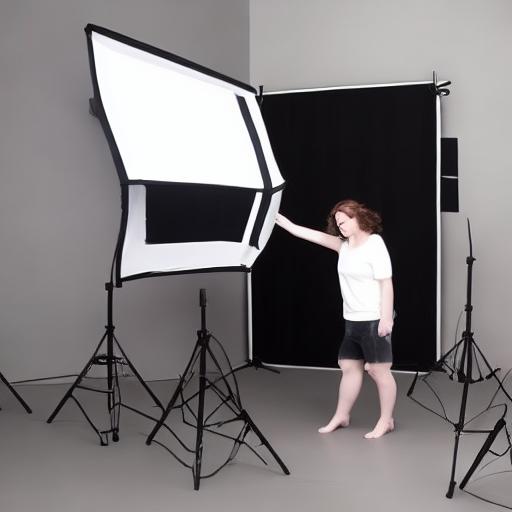}
        \end{subfigure}
        \begin{subfigure}[b]{\subwidth}
            \centering
            \includegraphics[width=\textwidth]{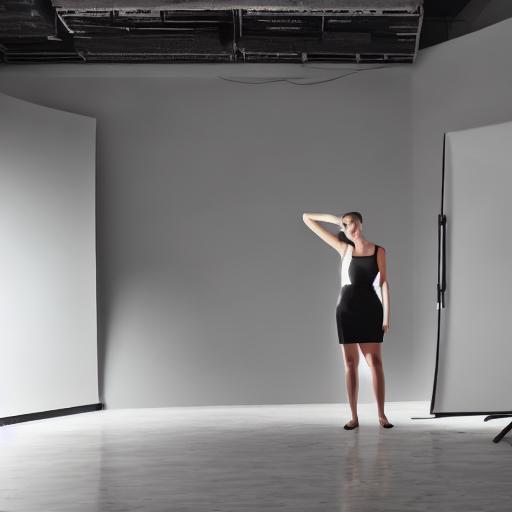}
        \end{subfigure}
        \begin{subfigure}[b]{\subwidth}
            \centering
            \includegraphics[width=\textwidth]{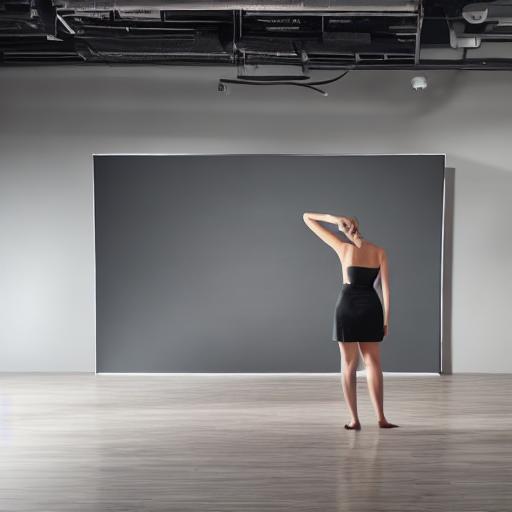}
        \end{subfigure}
        \begin{subfigure}[b]{\subwidth}
            \centering
            \includegraphics[width=\textwidth]{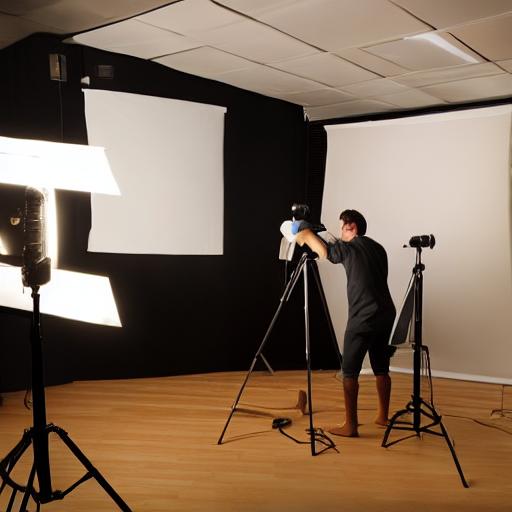}
        \end{subfigure}
        \begin{subfigure}[b]{\subwidth}
            \centering
            \includegraphics[width=\textwidth]{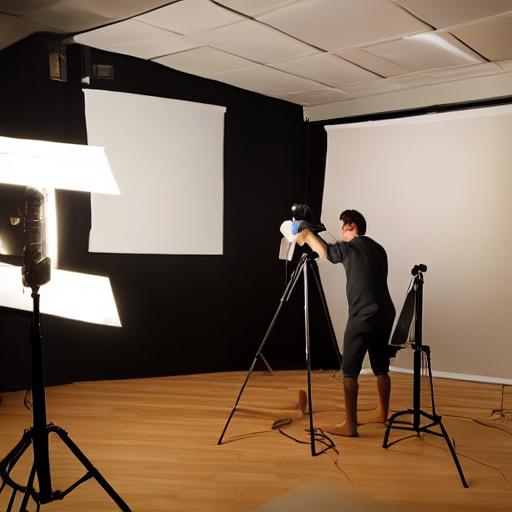}
        \end{subfigure}
        \begin{subfigure}[b]{\subwidth}
            \centering
            \includegraphics[width=\textwidth]{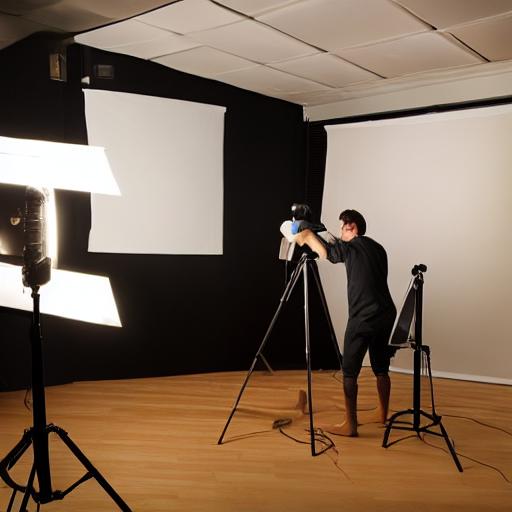}
        \end{subfigure} \\

        \begin{subfigure}[b]{\subwidth}
            \centering
            \includegraphics[width=\textwidth]{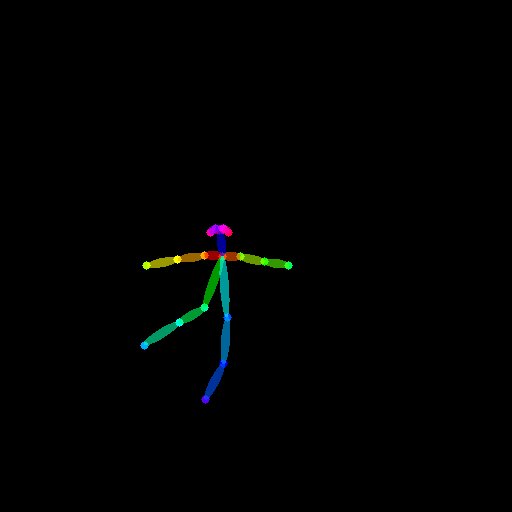}
        \end{subfigure}
        \begin{subfigure}[b]{\subwidth}
            \centering
            \includegraphics[trim={0cm 2.5cm 0cm 3.5cm},width=\textwidth, clip, height=\textwidth]{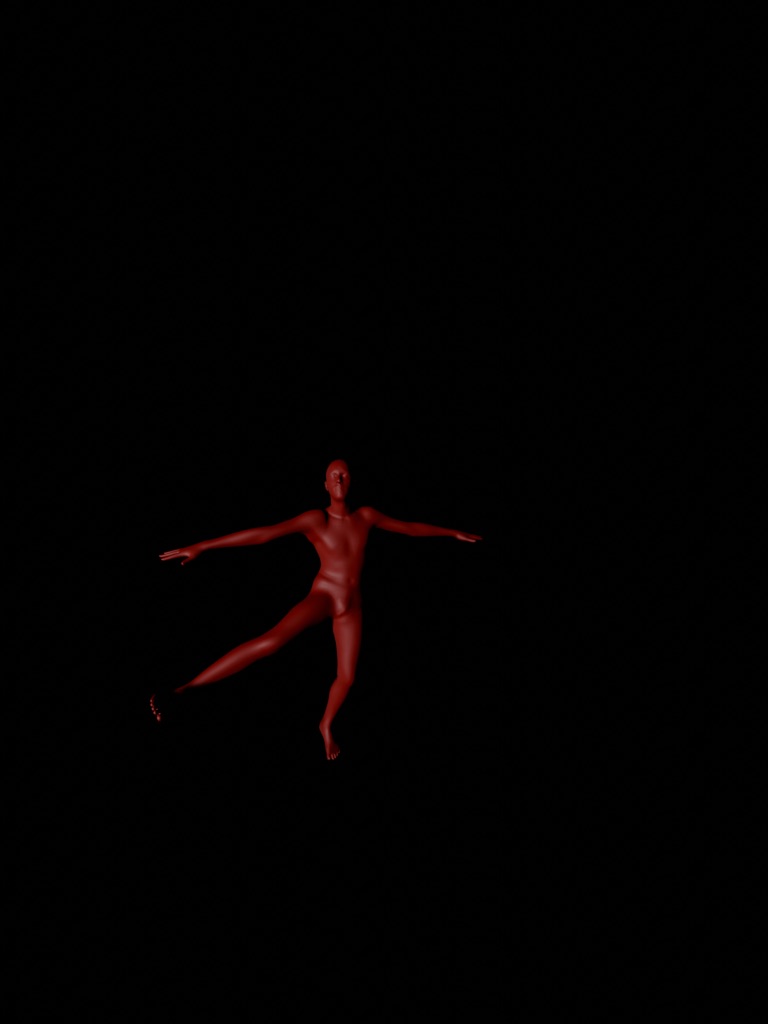}
        \end{subfigure}
        \begin{subfigure}[b]{\subwidth}
            \centering
            \includegraphics[width=\textwidth]{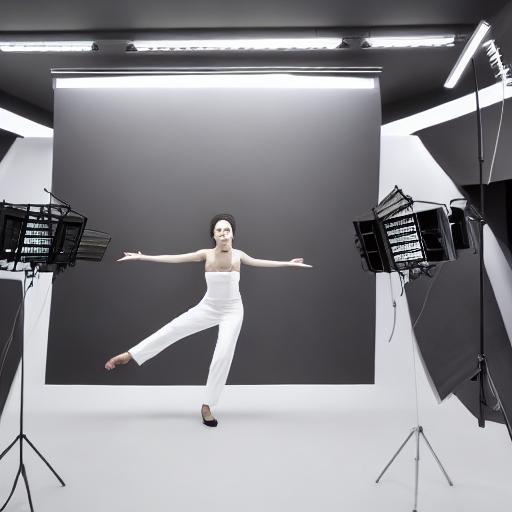}
        \end{subfigure}
        \begin{subfigure}[b]{\subwidth}
            \centering
            \includegraphics[width=\textwidth]{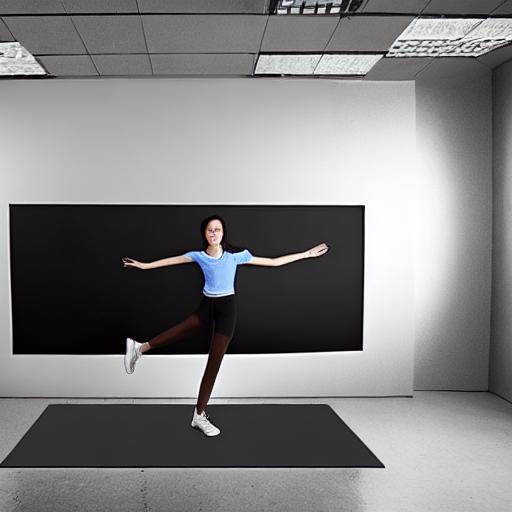}
        \end{subfigure}
        \begin{subfigure}[b]{\subwidth}
            \centering
            \includegraphics[width=\textwidth]{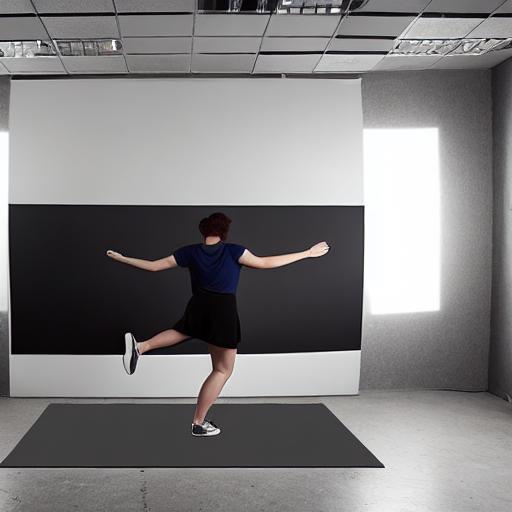}
        \end{subfigure}
        \begin{subfigure}[b]{\subwidth}
            \centering
            \includegraphics[width=\textwidth]{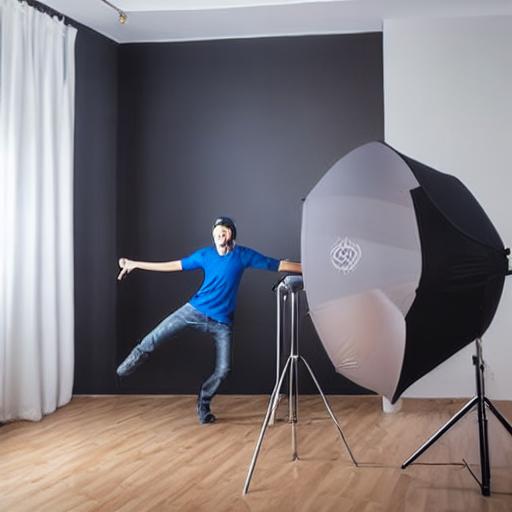}
        \end{subfigure}
        \begin{subfigure}[b]{\subwidth}
            \centering
            \includegraphics[width=\textwidth]{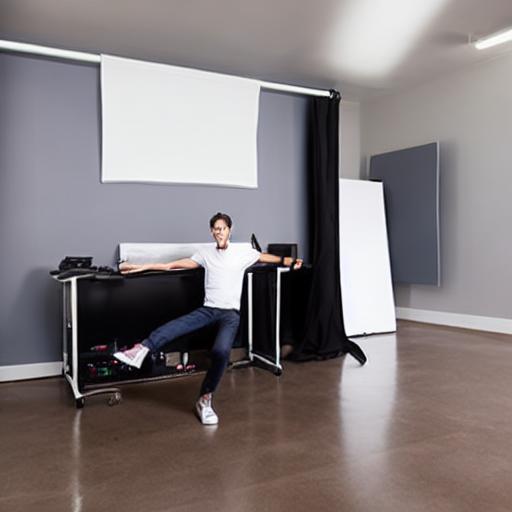}
        \end{subfigure}
        \begin{subfigure}[b]{\subwidth}
            \centering
            \includegraphics[width=\textwidth]{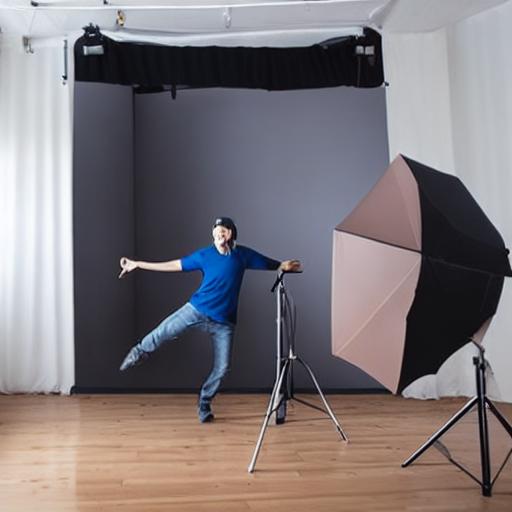}
        \end{subfigure} \\

        \begin{subfigure}[b]{\subwidth}
            \centering
            \includegraphics[width=\textwidth]{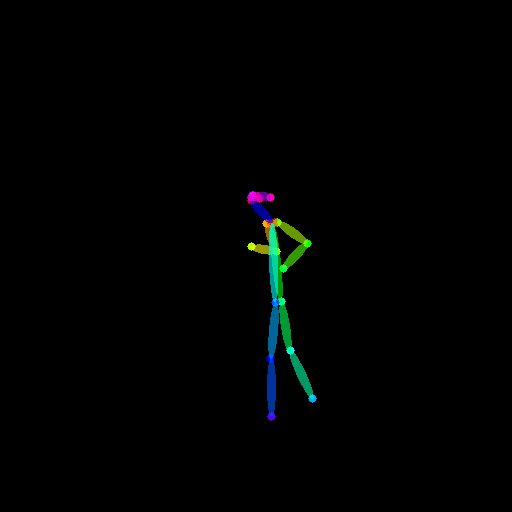}
        \end{subfigure}
        \begin{subfigure}[b]{\subwidth}
            \centering
            \includegraphics[trim={0cm 2.5cm 0cm 3.5cm},width=\textwidth, clip, height=\textwidth]{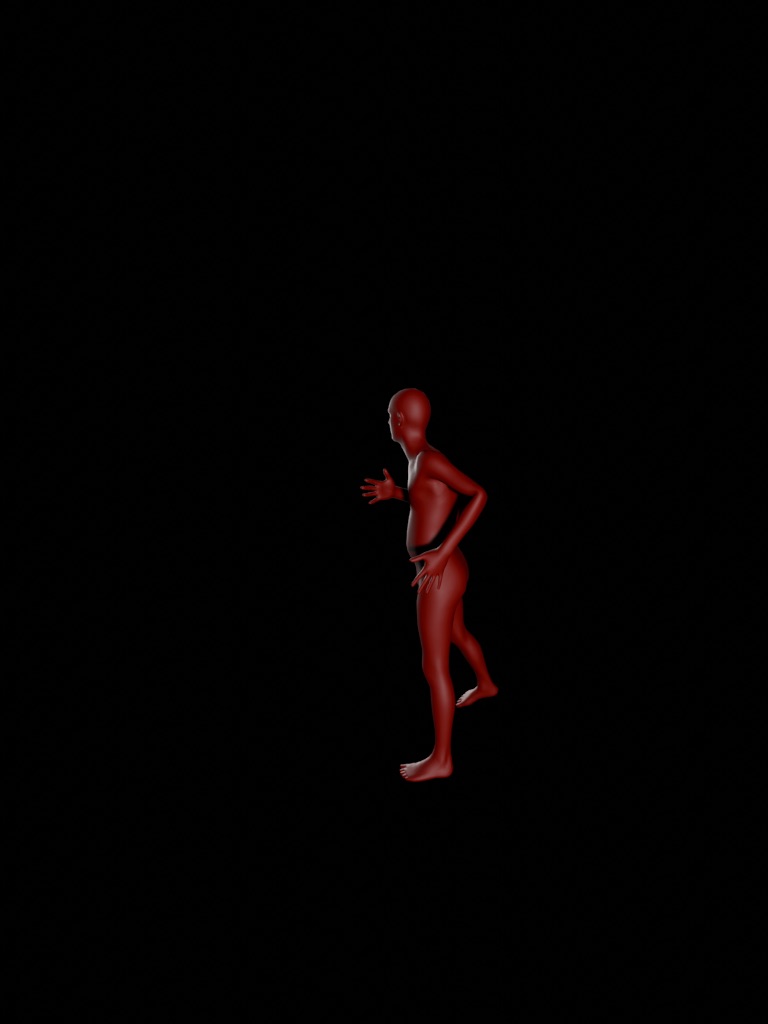}
        \end{subfigure}
        \begin{subfigure}[b]{\subwidth}
            \centering
            \includegraphics[width=\textwidth]{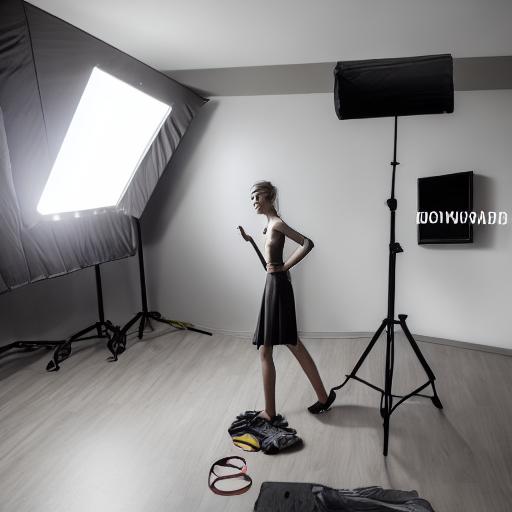}
        \end{subfigure}
        \begin{subfigure}[b]{\subwidth}
            \centering
            \includegraphics[width=\textwidth]{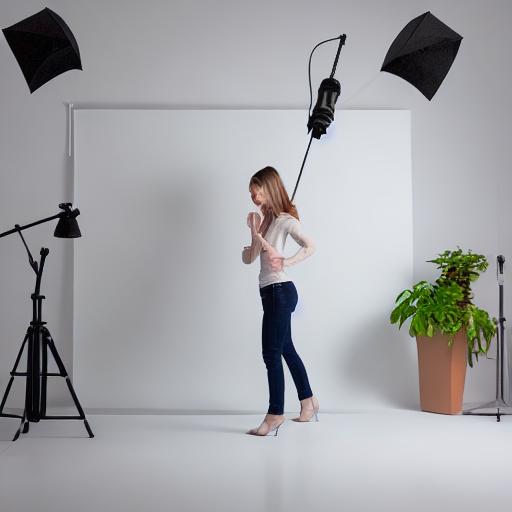}
        \end{subfigure}
        \begin{subfigure}[b]{\subwidth}
            \centering
            \includegraphics[width=\textwidth]{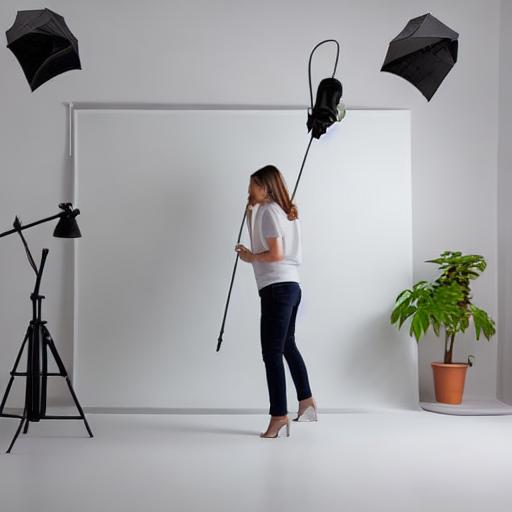}
        \end{subfigure}
        \begin{subfigure}[b]{\subwidth}
            \centering
            \includegraphics[width=\textwidth]{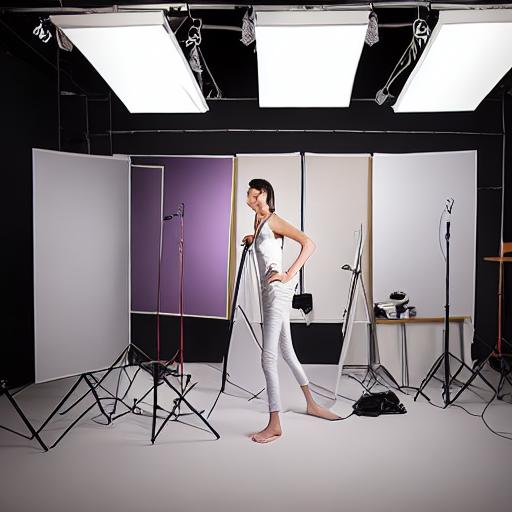}
        \end{subfigure}
        \begin{subfigure}[b]{\subwidth}
            \centering
            \includegraphics[width=\textwidth]{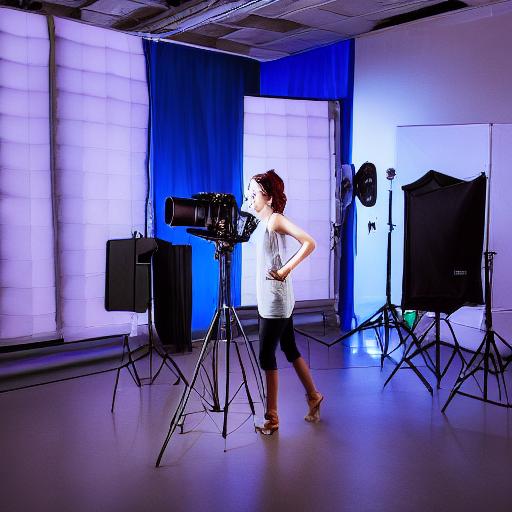}
        \end{subfigure}
        \begin{subfigure}[b]{\subwidth}
            \centering
            \includegraphics[width=\textwidth]{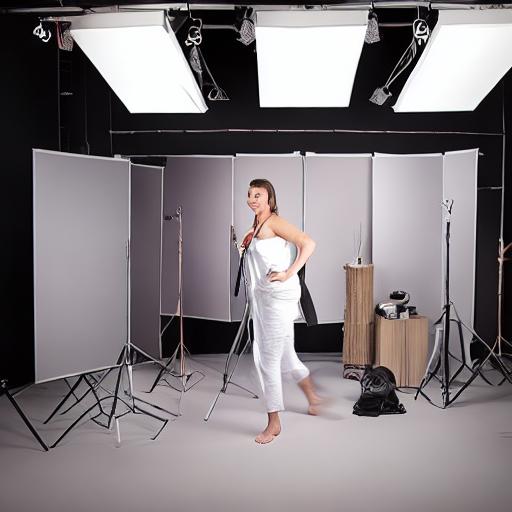}
        \end{subfigure} \\
        
        \begin{subfigure}[b]{\subwidth}
            \centering
            \includegraphics[width=\textwidth]{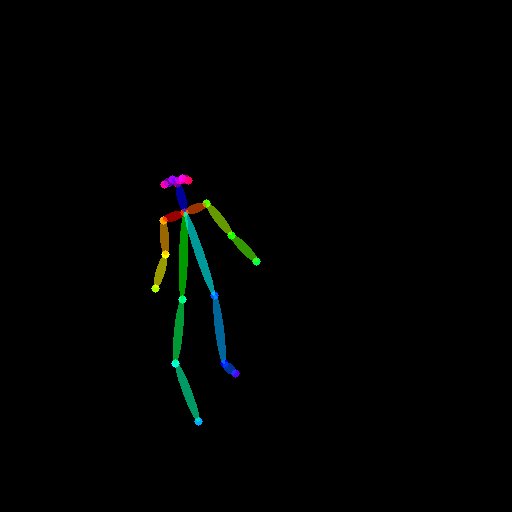}
        \end{subfigure}
        \begin{subfigure}[b]{\subwidth}
            \centering
            \includegraphics[trim={0cm 2.5cm 0cm 3.5cm},width=\textwidth, clip, height=\textwidth]{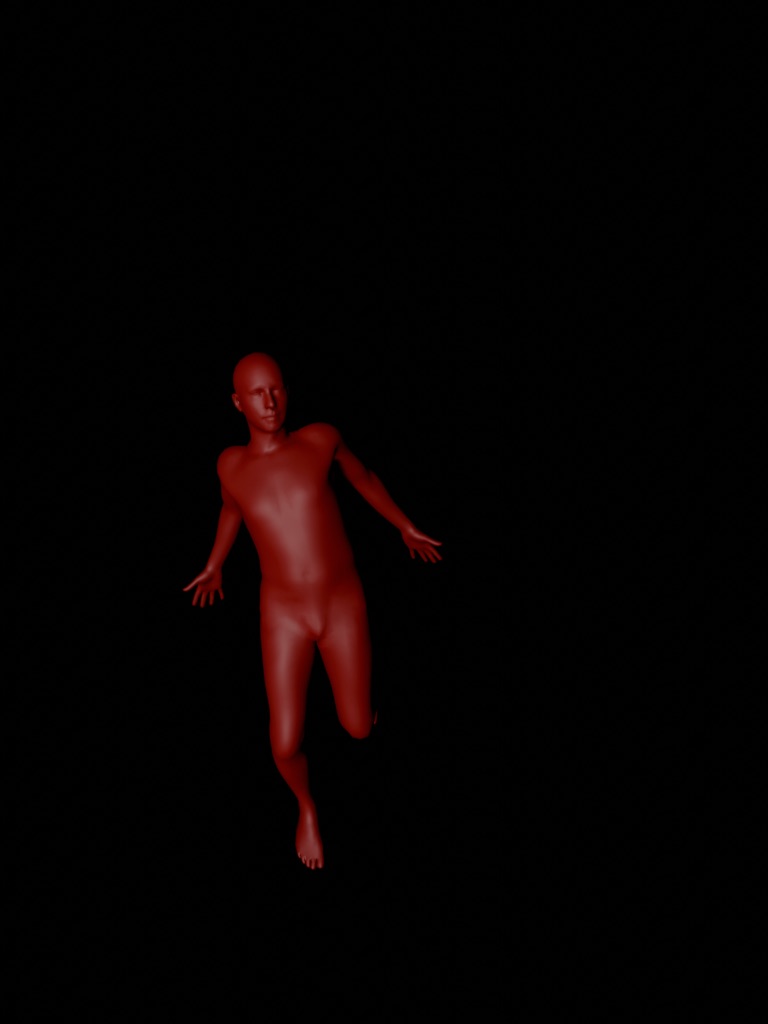}
        \end{subfigure}
        \begin{subfigure}[b]{\subwidth}
            \centering
            \includegraphics[width=\textwidth]{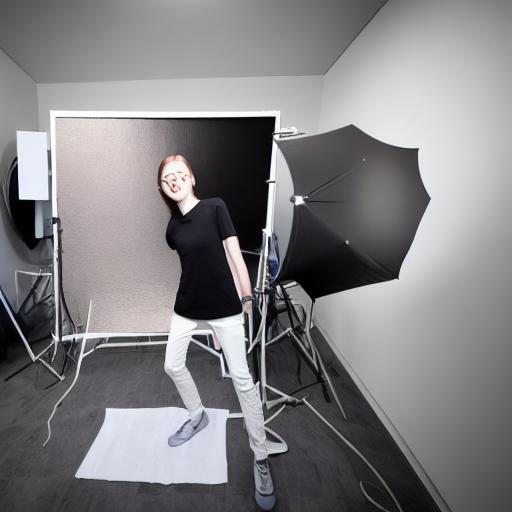}
        \end{subfigure}
        \begin{subfigure}[b]{\subwidth}
            \centering
            \includegraphics[width=\textwidth]{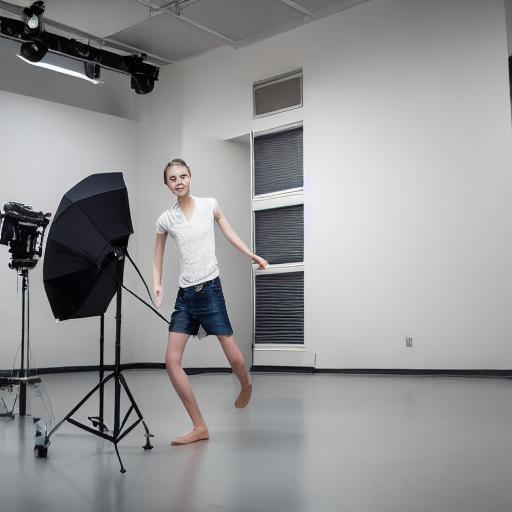}
        \end{subfigure}
        \begin{subfigure}[b]{\subwidth}
            \centering
            \includegraphics[width=\textwidth]{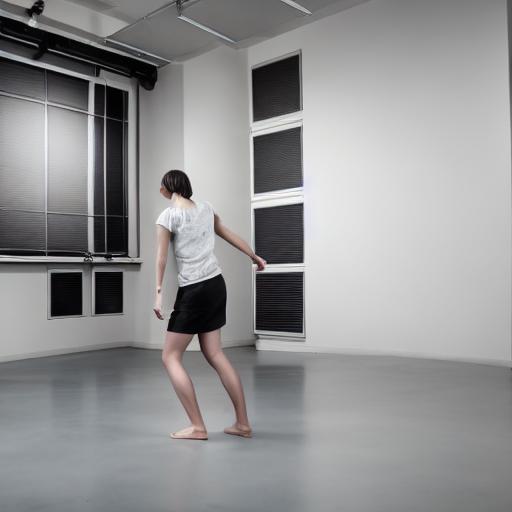}
        \end{subfigure}
        \begin{subfigure}[b]{\subwidth}
            \centering
            \includegraphics[width=\textwidth]{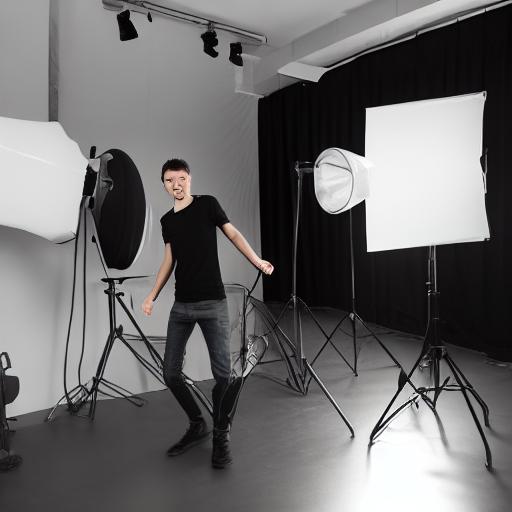}
        \end{subfigure}
        \begin{subfigure}[b]{\subwidth}
            \centering
            \includegraphics[width=\textwidth]{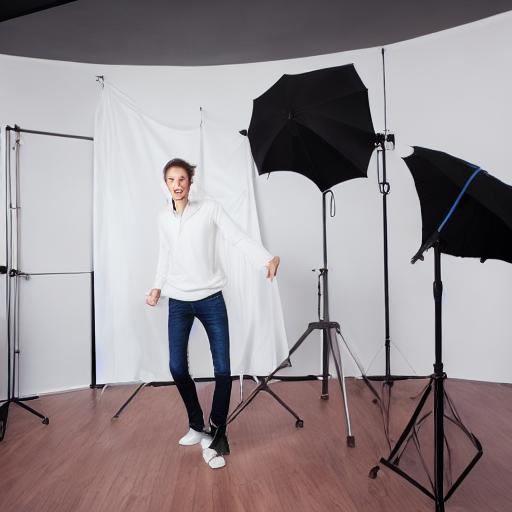}
        \end{subfigure}
        \begin{subfigure}[b]{\subwidth}
            \centering
            \includegraphics[width=\textwidth]{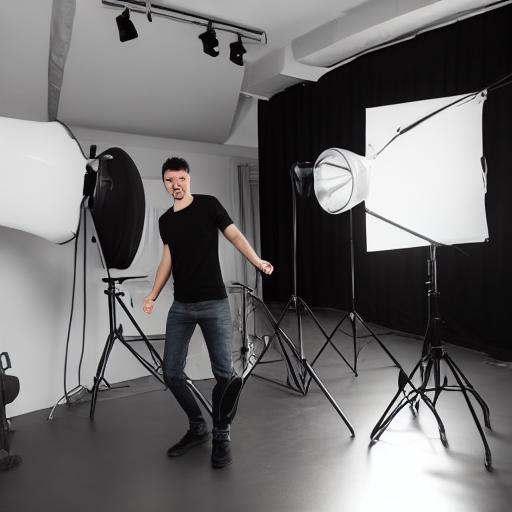}
        \end{subfigure} \\

        \begin{subfigure}[b]{\subwidth}
            \centering
            \includegraphics[width=\textwidth]{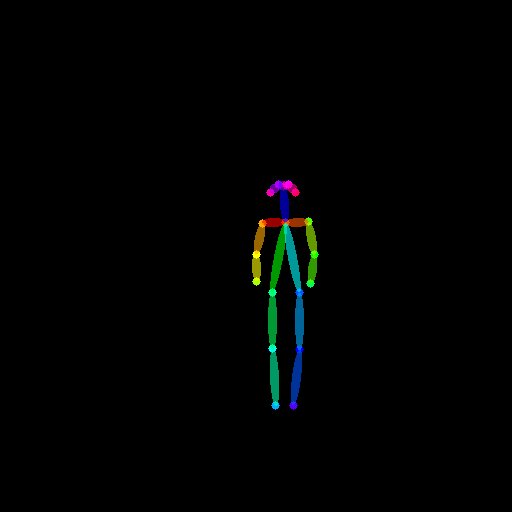}
        \end{subfigure}
        \begin{subfigure}[b]{\subwidth}
            \centering
            \includegraphics[trim={0cm 2.5cm 0cm 3.5cm},width=\textwidth, clip, height=\textwidth]{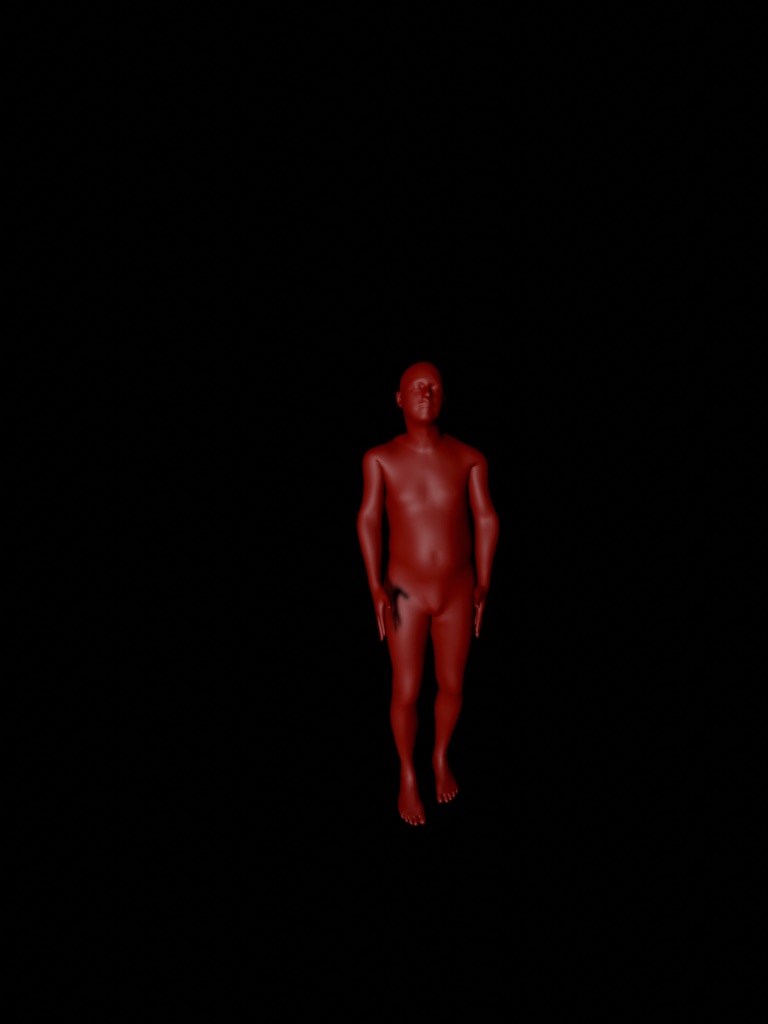}
        \end{subfigure}
        \begin{subfigure}[b]{\subwidth}
            \centering
            \includegraphics[width=\textwidth]{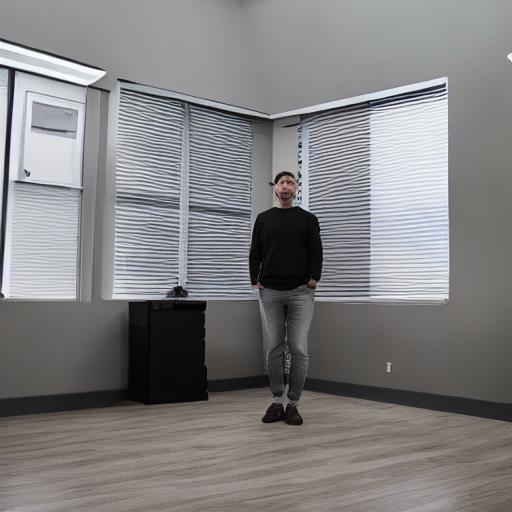}
        \end{subfigure}
        \begin{subfigure}[b]{\subwidth}
            \centering
            \includegraphics[width=\textwidth]{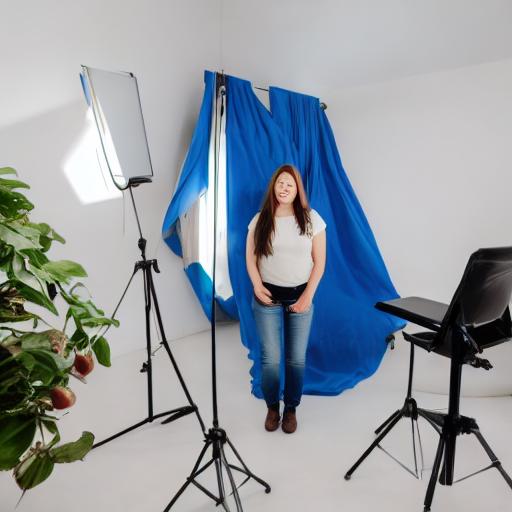}
        \end{subfigure}
        \begin{subfigure}[b]{\subwidth}
            \centering
            \includegraphics[width=\textwidth]{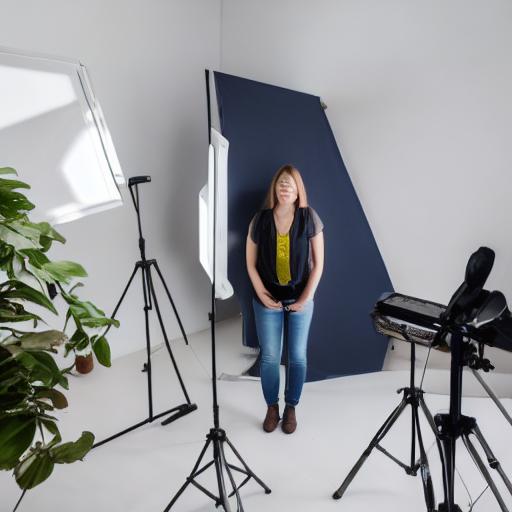}
        \end{subfigure}
        \begin{subfigure}[b]{\subwidth}
            \centering
            \includegraphics[width=\textwidth]{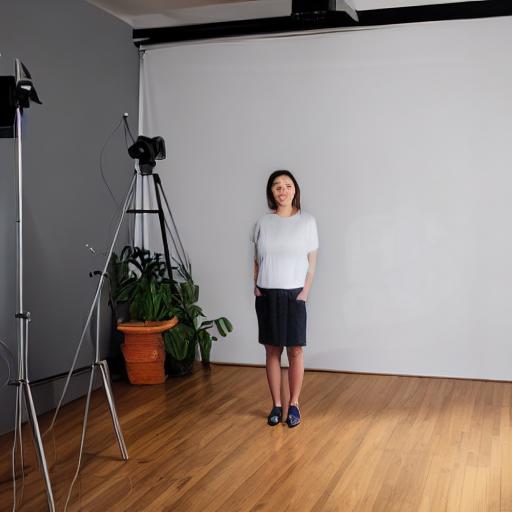}
        \end{subfigure}
        \begin{subfigure}[b]{\subwidth}
            \centering
            \includegraphics[width=\textwidth]{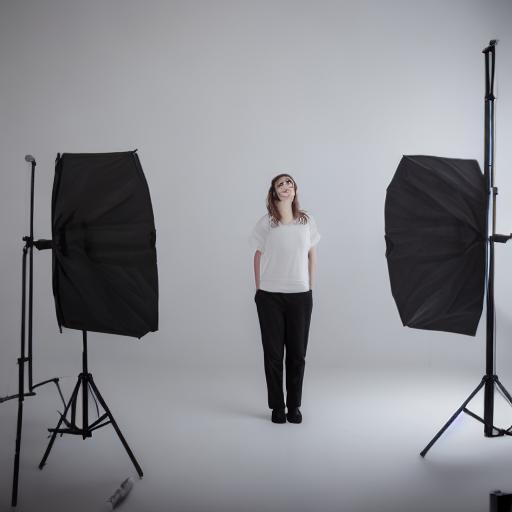}
        \end{subfigure}
        \begin{subfigure}[b]{\subwidth}
            \centering
            \includegraphics[width=\textwidth]{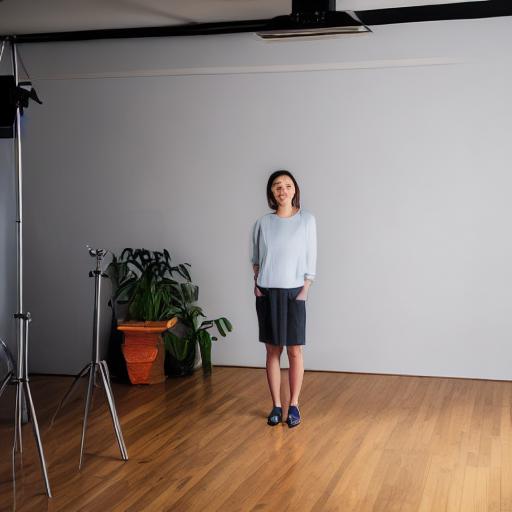}
        \end{subfigure} \\

        \begin{subfigure}[b]{\subwidth}
            \centering
            \includegraphics[width=\textwidth]{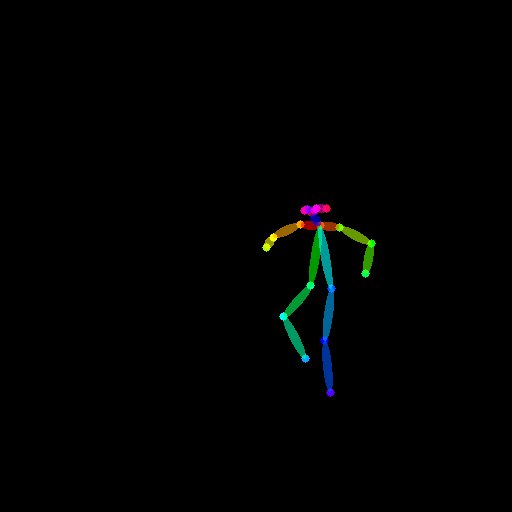}
            \caption{Pose}
        \end{subfigure}
        \begin{subfigure}[b]{\subwidth}
            \centering
            \includegraphics[trim={0cm 2.5cm 0cm 3.5cm},width=\textwidth, clip, height=\textwidth]{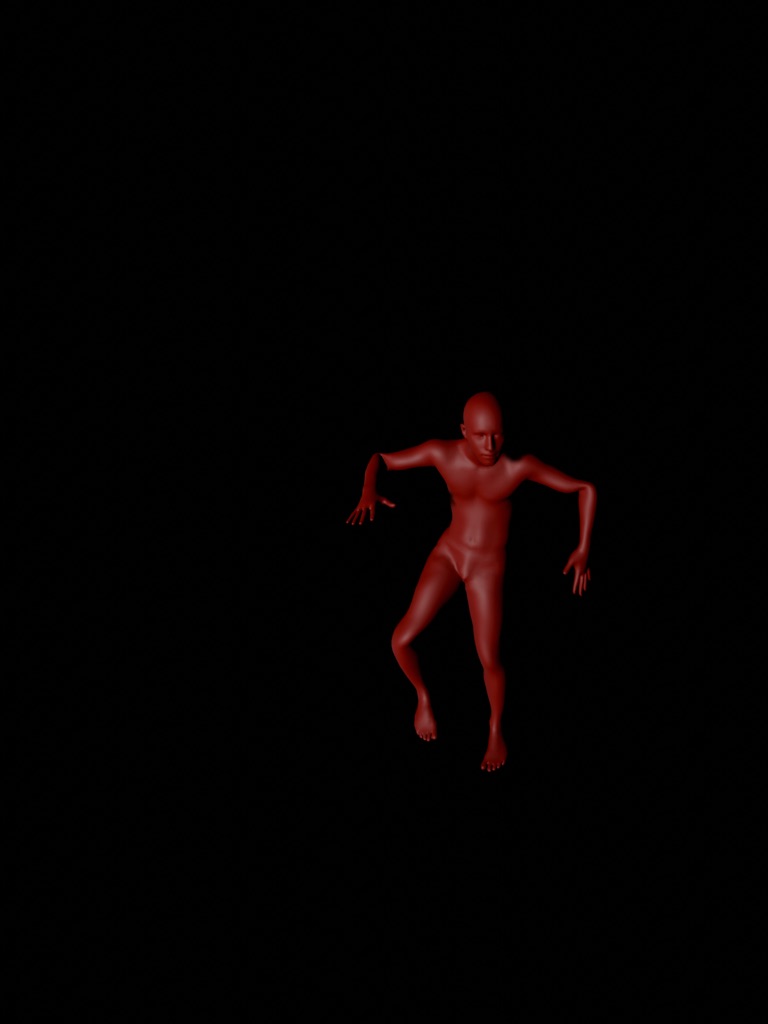}
            \caption{SMPL}
        \end{subfigure}
        \begin{subfigure}[b]{\subwidth}
            \centering
            \includegraphics[width=\textwidth]{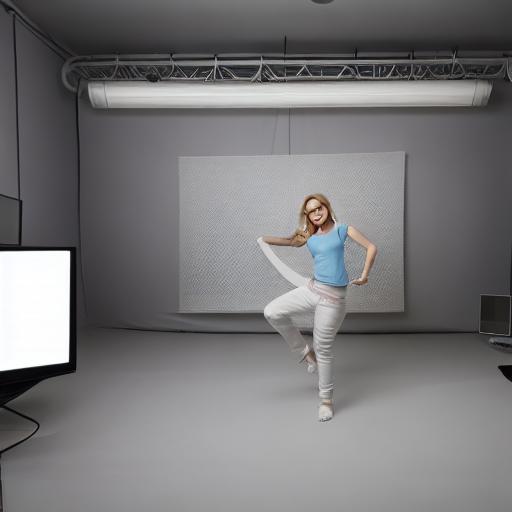}
            \caption{ft-all+CN}
        \end{subfigure}
        \begin{subfigure}[b]{\subwidth}
            \centering
            \includegraphics[width=\textwidth]{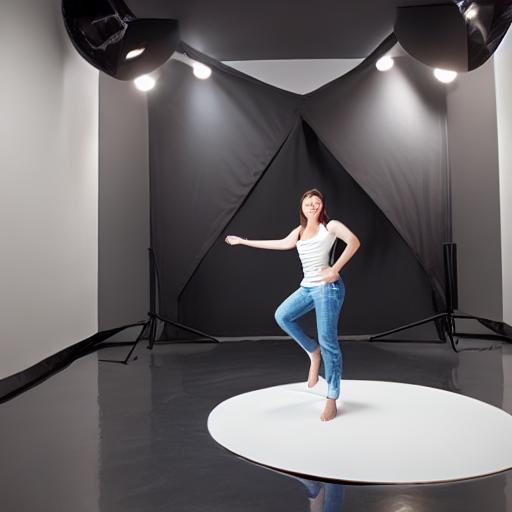}
            \caption{ft-attn+CN}
        \end{subfigure}
        \begin{subfigure}[b]{\subwidth}
            \centering
            \includegraphics[width=\textwidth]{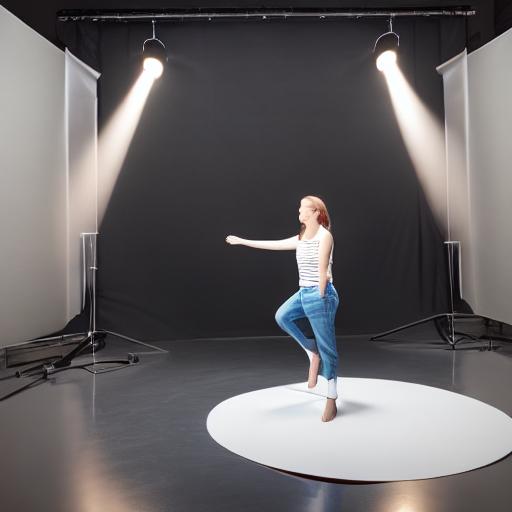}
            \caption{ControlNet}
        \end{subfigure}
        \begin{subfigure}[b]{\subwidth}
            \centering
            \includegraphics[width=\textwidth]{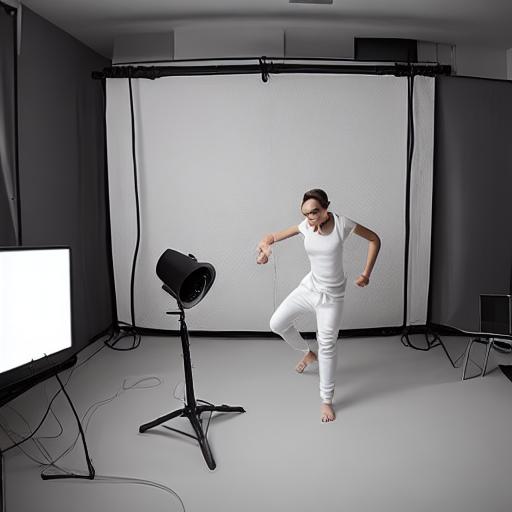}
            \caption{ft-all+T2I}
        \end{subfigure}
        \begin{subfigure}[b]{\subwidth}
            \centering
            \includegraphics[width=\textwidth]{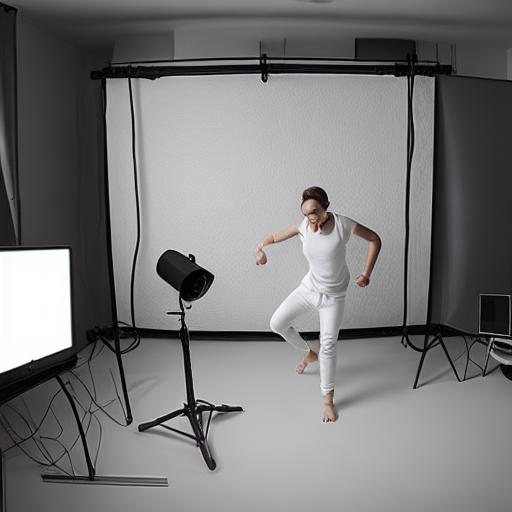}
            \caption{ft-attn+T2I}
        \end{subfigure}
        \begin{subfigure}[b]{\subwidth}
            \centering
            \includegraphics[width=\textwidth]{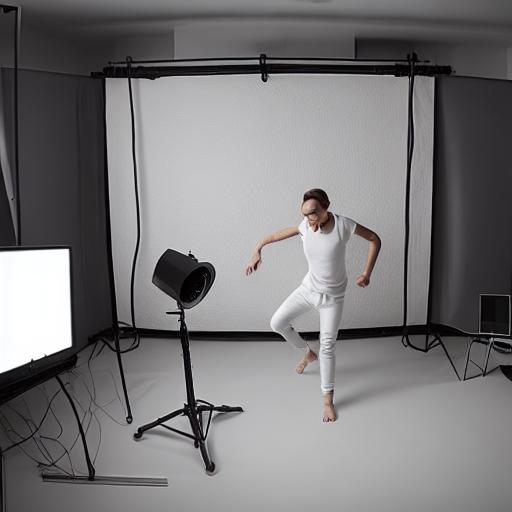}
            \caption{T2I-Adapter}
        \end{subfigure}
    \caption{Examples from the pose and shape accuracy evaluation (Table 2, main paper). Our SMPL-finetuned (``ft-'') models receive the pose and SMPL inputs, while the standard ControlNet and T2I-Adapter only receive the pose input. The prompt for generating the entire dataset was ``A person in a clean studio environment''. }
    \label{fig:aist_eval_comparison}

    \end{figure*}

%% file: suppl_fig_tablemacros.tex
\newcommand{\twobytwotable}[3]{%
    \setlength{\tabcolsep}{0em}
    \renewcommand{\arraystretch}{0}
    \begin{tabular}{l@{} l@{} >{\centering\arraybackslash}m{#2} >{\centering\arraybackslash}m{#2}}
      $w_1$ & $\rightarrow$ &&\\
      & & \textbf{0} & \textbf{7.5} \\[0.5em]
                
    \rotatebox{90}{\hspace{#3}$\leftarrow w_2$} &\rotatebox{90}{\textbf{0}} &
    \adjincludegraphics[Clip={0.0\width} {0.5\height} {0.5\width} {0.0\height},width=#2]{#1} &
    \adjincludegraphics[Clip={0.5\width} {0.5\height} {0.0\width} {0.0\height},width=#2]{#1} \\
    
    &\rotatebox{90}{\textbf{7.5}} &        
    \adjincludegraphics[Clip={0.0\width} {0.0\height} {0.5\width} {0.5\height},width=#2]{#1} &
    \adjincludegraphics[Clip={0.5\width} {0.0\height} {0.0\width} {0.5\height},width=#2]{#1} \\
    \end{tabular}
}

\newcommand{\threexthreetable}[3]{
    \setlength{\tabcolsep}{0em}
    \renewcommand{\arraystretch}{0}
    \begin{tabular}{l@{} l@{} >{\centering\arraybackslash}m{#2} >{\centering\arraybackslash}m{#2} >{\centering\arraybackslash}m{#2}}
      & $w_1\rightarrow$ &&&\\
      & & \textbf{0} & \textbf{1} & \textbf{7.5} \\
            
    \rotatebox{90}{\hspace{#3}$\leftarrow w_2$} &\rotatebox{90}{\textbf{0}} &
    \adjincludegraphics[Clip={0.0\width} {0.6667\height} {0.6667\width} {0.0\height},width=#2]{#1} &
   \adjincludegraphics[Clip={0.0\width} {0.3333\height} {0.6667\width} {0.3333\height},width=#2]{#1} &
    \adjincludegraphics[Clip={0.0\width} {0.0\height} {0.6667\width} {0.6667\height},width=#2]{#1} \\
    
    &\rotatebox{90}{\textbf{1}} &        
    \adjincludegraphics[Clip={0.3333\width} {0.6667\height} {0.3333\width} {0.0\height},width=#2]{#1} &
    \adjincludegraphics[Clip={0.3333\width} {0.3333\height} {0.3333\width} {0.3333\height},width=#2]{#1} &
    \adjincludegraphics[Clip={0.3333\width} {0.0\height} {0.3333\width} {0.6667\height},width=#2]{#1}   \\
    
    &\rotatebox{90}{\textbf{7.5}} &
    \adjincludegraphics[Clip={0.6667\width} {0.6667\height} {0.0\width} {0.0\height},width=#2]{#1} &
    \adjincludegraphics[Clip={0.6667\width} {0.3333\height} {0.0\width} {0.3333\height},width=#2]{#1} &
    \adjincludegraphics[Clip={0.6667\width} {0.0\height} {0.0\width} {0.6667\height},width=#2]{#1} \\
    \end{tabular}
}

\newcommand{\fivexonetable}[2]{
    \setlength{\tabcolsep}{0.08em}
    \renewcommand{\arraystretch}{0}
\begin{tabular}{>{\centering\arraybackslash}m{#2} >{\centering\arraybackslash}m{#2} >{\centering\arraybackslash}m{#2} >{\centering\arraybackslash}m{#2} >{\centering\arraybackslash}m{#2}}
    \textbf{0} & \textbf{1} & \textbf{5} & \textbf{7.5} & \textbf{12} \\[0.5em]
        
    \adjincludegraphics[Clip={0.0\width} {0.0\height} {0.8\width} {0.0\height},width=#2]{#1} &
    \adjincludegraphics[Clip={0.2\width} {0.0\height} {0.6\width} {0.0\height},width=#2]{#1} &
    \adjincludegraphics[Clip={0.4\width} {0.0\height} {0.4\width} {0.0\height},width=#2]{#1} &
    \adjincludegraphics[Clip={0.6\width} {0.0\height} {0.2\width} {0.0\height},width=#2]{#1} &
    \adjincludegraphics[Clip={0.8\width} {0.0\height} {0.0\width} {0.0\height},width=#2]{#1} \\

    \end{tabular}
}

\newcommand{\fivextwotable}[3]{
    \setlength{\tabcolsep}{0.08em}
    \renewcommand{\arraystretch}{0}
\begin{tabular}{>{\centering\arraybackslash}m{#3} >{\centering\arraybackslash}m{#3} >{\centering\arraybackslash}m{#3} >{\centering\arraybackslash}m{#3} >{\centering\arraybackslash}m{#3}}
    \textbf{0} & \textbf{1} & \textbf{5} & \textbf{7.5} & \textbf{12} \\[0.5em]
        
    \adjincludegraphics[Clip={0.0\width} {0.0\height} {0.8\width} {0.0\height},width=#3]{#1} &
    \adjincludegraphics[Clip={0.2\width} {0.0\height} {0.6\width} {0.0\height},width=#3]{#1} &
    \adjincludegraphics[Clip={0.4\width} {0.0\height} {0.4\width} {0.0\height},width=#3]{#1} &
    \adjincludegraphics[Clip={0.6\width} {0.0\height} {0.2\width} {0.0\height},width=#3]{#1} &
    \adjincludegraphics[Clip={0.8\width} {0.0\height} {0.0\width} {0.0\height},width=#3]{#1} \\

        \adjincludegraphics[Clip={0.0\width} {0.0\height} {0.8\width} {0.0\height},width=#3]{#2} &
    \adjincludegraphics[Clip={0.2\width} {0.0\height} {0.6\width} {0.0\height},width=#3]{#2} &
    \adjincludegraphics[Clip={0.4\width} {0.0\height} {0.4\width} {0.0\height},width=#3]{#2} &
    \adjincludegraphics[Clip={0.6\width} {0.0\height} {0.2\width} {0.0\height},width=#3]{#2} &
    \adjincludegraphics[Clip={0.8\width} {0.0\height} {0.0\width} {0.0\height},width=#3]{#2} 
    \end{tabular}
}

\newcommand{\sixxsixtable}[2]{
    \setlength{\tabcolsep}{0em}
    \renewcommand{\arraystretch}{0}
    \begin{tabular}{c c >{\centering\arraybackslash}m{#2} >{\centering\arraybackslash}m{#2} >{\centering\arraybackslash}m{#2} >{\centering\arraybackslash}m{#2} >{\centering\arraybackslash}m{#2} >{\centering\arraybackslash}m{#2}}
            & $w_1\rightarrow$ &&&&&&\\
            \multicolumn{2}{c}{} & \textbf{0} & \textbf{2} & \textbf{5} & \textbf{9} & \textbf{15} & \textbf{25} \\

        \rotatebox{90}{\hspace{1cm}$\leftarrow w_2$} &\rotatebox{90}{\textbf{0}} &
        \adjincludegraphics[width=#2]{graphics/#1_grid/#1_0_0.jpg} &
        \adjincludegraphics[width=#2]{graphics/#1_grid/#1_0_1.jpg} &
        \adjincludegraphics[width=#2]{graphics/#1_grid/#1_0_2.jpg} &
        \adjincludegraphics[width=#2]{graphics/#1_grid/#1_0_3.jpg} &
        \adjincludegraphics[width=#2]{graphics/#1_grid/#1_0_4.jpg} &
        \adjincludegraphics[width=#2]{graphics/#1_grid/#1_0_5.jpg} \\
        & \rotatebox{90}{\textbf{2}} &
        \adjincludegraphics[width=#2]{graphics/#1_grid/#1_1_0.jpg} &
        \adjincludegraphics[width=#2]{graphics/#1_grid/#1_1_1.jpg} &
        \adjincludegraphics[width=#2]{graphics/#1_grid/#1_1_2.jpg} &
        \adjincludegraphics[width=#2]{graphics/#1_grid/#1_1_3.jpg} &
        \adjincludegraphics[width=#2]{graphics/#1_grid/#1_1_4.jpg} &
        \adjincludegraphics[width=#2]{graphics/#1_grid/#1_1_5.jpg} \\
        & \rotatebox{90}{\textbf{5}} &
        \adjincludegraphics[width=#2]{graphics/#1_grid/#1_2_0.jpg} &
        \adjincludegraphics[width=#2]{graphics/#1_grid/#1_2_1.jpg} &
        \adjincludegraphics[width=#2]{graphics/#1_grid/#1_2_2.jpg} &
        \adjincludegraphics[width=#2]{graphics/#1_grid/#1_2_3.jpg} &
        \adjincludegraphics[width=#2]{graphics/#1_grid/#1_2_4.jpg} &
        \adjincludegraphics[width=#2]{graphics/#1_grid/#1_2_5.jpg} \\
        & \rotatebox{90}{\textbf{9}} &
        \adjincludegraphics[width=#2]{graphics/#1_grid/#1_3_0.jpg} &
        \adjincludegraphics[width=#2]{graphics/#1_grid/#1_3_1.jpg} &
        \adjincludegraphics[width=#2]{graphics/#1_grid/#1_3_2.jpg} &
        \adjincludegraphics[width=#2]{graphics/#1_grid/#1_3_3.jpg} &
        \adjincludegraphics[width=#2]{graphics/#1_grid/#1_3_4.jpg} &
        \adjincludegraphics[width=#2]{graphics/#1_grid/#1_3_5.jpg} \\
        & \rotatebox{90}{\textbf{15}} &
        \adjincludegraphics[width=#2]{graphics/#1_grid/#1_4_0.jpg} &
        \adjincludegraphics[width=#2]{graphics/#1_grid/#1_4_1.jpg} &
        \adjincludegraphics[width=#2]{graphics/#1_grid/#1_4_2.jpg} &
        \adjincludegraphics[width=#2]{graphics/#1_grid/#1_4_3.jpg} &
        \adjincludegraphics[width=#2]{graphics/#1_grid/#1_4_4.jpg} &
        \adjincludegraphics[width=#2]{graphics/#1_grid/#1_4_5.jpg} \\
        & \rotatebox{90}{\textbf{25}} &
        \adjincludegraphics[width=#2]{graphics/#1_grid/#1_5_0.jpg} &
        \adjincludegraphics[width=#2]{graphics/#1_grid/#1_5_1.jpg} &
        \adjincludegraphics[width=#2]{graphics/#1_grid/#1_5_2.jpg} &
        \adjincludegraphics[width=#2]{graphics/#1_grid/#1_5_3.jpg} &
        \adjincludegraphics[width=#2]{graphics/#1_grid/#1_5_4.jpg} &
        \adjincludegraphics[width=#2]{graphics/#1_grid/#1_5_5.jpg} \\
\end{tabular}
}